\documentclass{article}
\usepackage{microtype}
\usepackage{graphicx}
\usepackage{booktabs}
\usepackage{hyperref}

\usepackage[accepted]{icml2023}
\usepackage{amsmath}
\usepackage{amssymb}
\usepackage{mathtools}
\usepackage{amsthm}
\usepackage{xcolor}
\usepackage{xspace}
\usepackage[capitalize,noabbrev]{cleveref}
\usepackage{multirow}
\usepackage{float}
\usepackage{subfig}
\usepackage{listings}
\usepackage{pythonhighlight}

\theoremstyle{plain}

\theoremstyle{definition}

\theoremstyle{remark}

\usepackage[textsize=tiny]{todonotes}
\definecolor{amber}{rgb}{1.0, 0.49, 0.0}

\newcommand{\set}[1]{\{#1\}}

\newcommand{\norm}[1]{\left\lVert#1\right\rVert}

\newcommand{\method}{AFR\xspace}
\newcommand{\jtt}{\textsc{jtt}\xspace}
\newcommand{\cnc}{C\textsc{n}C\xspace}

\icmltitlerunning{Simple and Fast Group Robustness by Automatic Feature Reweighting}

\begin{document}

\twocolumn[
\icmltitle{Simple and Fast Group Robustness by Automatic Feature Reweighting}
\icmlsetsymbol{equal}{*}

\begin{icmlauthorlist}
\icmlauthor{Shikai Qiu}{nyu,equal}
\icmlauthor{Andres Potapczynski}{nyu,equal}
\icmlauthor{Pavel Izmailov}{nyu}
\icmlauthor{Andrew Gordon Wilson}{nyu}
\end{icmlauthorlist}
\icmlaffiliation{nyu}{New York University}
\icmlcorrespondingauthor{Shikai Qiu}{sq2129@nyu.edu}
\icmlcorrespondingauthor{Andres Potapczynski}{ap6604@nyu.edu}
\icmlcorrespondingauthor{Pavel Izmailov}{pi390@nyu.edu}
\icmlcorrespondingauthor{Andrew Gordon Wilson}{andrewgw@cims.nyu.edu}

\icmlkeywords{Machine Learning, ICML}

\vskip 0.3in
]
\printAffiliationsAndNotice{\icmlEqualContribution}
\begin{abstract}
A major challenge to out-of-distribution generalization is reliance on spurious features --- patterns that are predictive of the class label in the training data distribution, but not causally related to the target.
Standard methods for reducing the reliance on spurious features typically assume that we know what the spurious feature is, which is rarely true in the real world. Methods that attempt to alleviate this limitation are complex, hard to tune, and lead to a significant computational overhead compared to standard training.
In this paper, we propose Automatic Feature Reweighting (\method), an extremely simple and fast method for updating the model to reduce the reliance on spurious features.
AFR retrains the last layer of a standard ERM-trained base model with a weighted loss that emphasizes the examples where the ERM model predicts poorly, automatically upweighting the minority group without group labels.
With this simple procedure, we improve upon the best reported results among competing methods trained without spurious attributes on several vision and natural language classification benchmarks, using only a fraction of their compute.
\end{abstract}

\section{Introduction}
\label{Intro}

Most realistic datasets contain features that can be used to achieve strong performance in-distribution, but that are not inherently relevant to the predictive task.
For example, on Waterbirds \citep{sagawa2020GDRO}, an image classification dataset where the goal is to distinguish landbirds (e.g. woodpecker) from waterbirds (e.g. seagull), a model can reach 95\% in-distribution accuracy by only looking at the background and completely ignoring the actual bird.

Features such as background (ocean), which are correlated with but not causally related to the target (waterbird) are known as \textit{spurious features} or \textit{spurious attributes}.
By relying on spurious features, a model performs well on the training data distribution without learning the \textit{core features} that are intrinsically descriptive of the targets. Such models inevitably fail to generalize to new data where the spurious correlation breaks, such as images of waterbirds appearing on land backgrounds.

The best existing approaches for addressing spurious features require access to group labels on the training data \citep{sagawa2020GDRO, idrissi2021simple}, which simultaneously specify both the class label and the spurious attribute for each datapoint.
This requirement presents a major limitation: while many real-world problems contain spurious correlations, we do not know what they are a priori! Moreover, even in rare cases where we could potentially identify the spurious feature, it can be prohibitively expensive to manually add group labels to a large dataset.
Consequently, there has been great interest in reducing the reliance on spurious features without explicit access to the spurious attributes \citep[e.g.,][]{liu2021jtt, nam2020lff, zhang2022cnc, lee2022divdis}.
However, compared to standard training, the resulting methods are generally more complex, computationally heavy, and difficult to tune, making them difficult to adopt in practice.

\begin{figure*}[!ht]
\centering
    \begin{tabular}{ccc}
    \includegraphics[width=.24\linewidth]{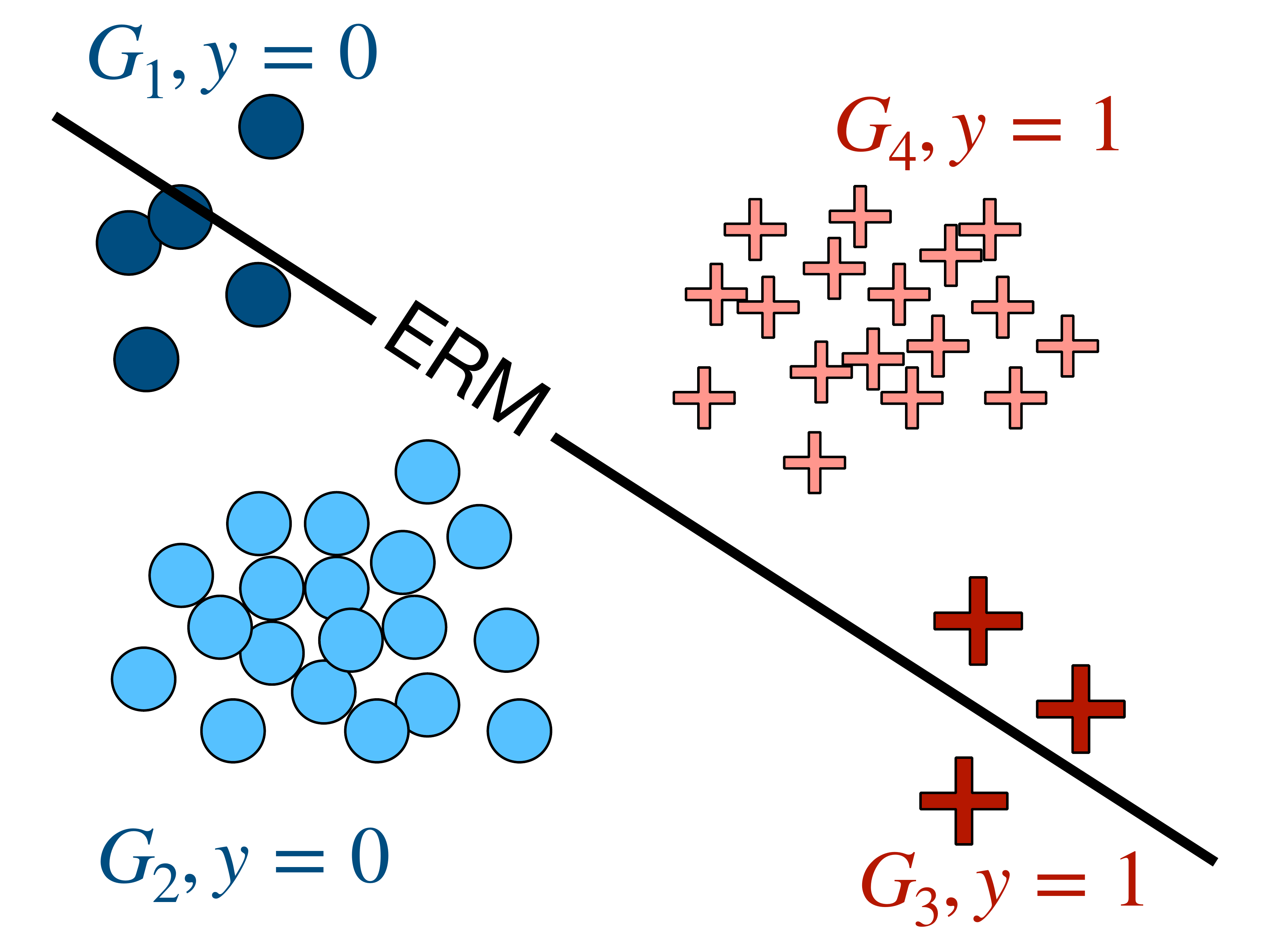} &
    \includegraphics[width=.24\linewidth]{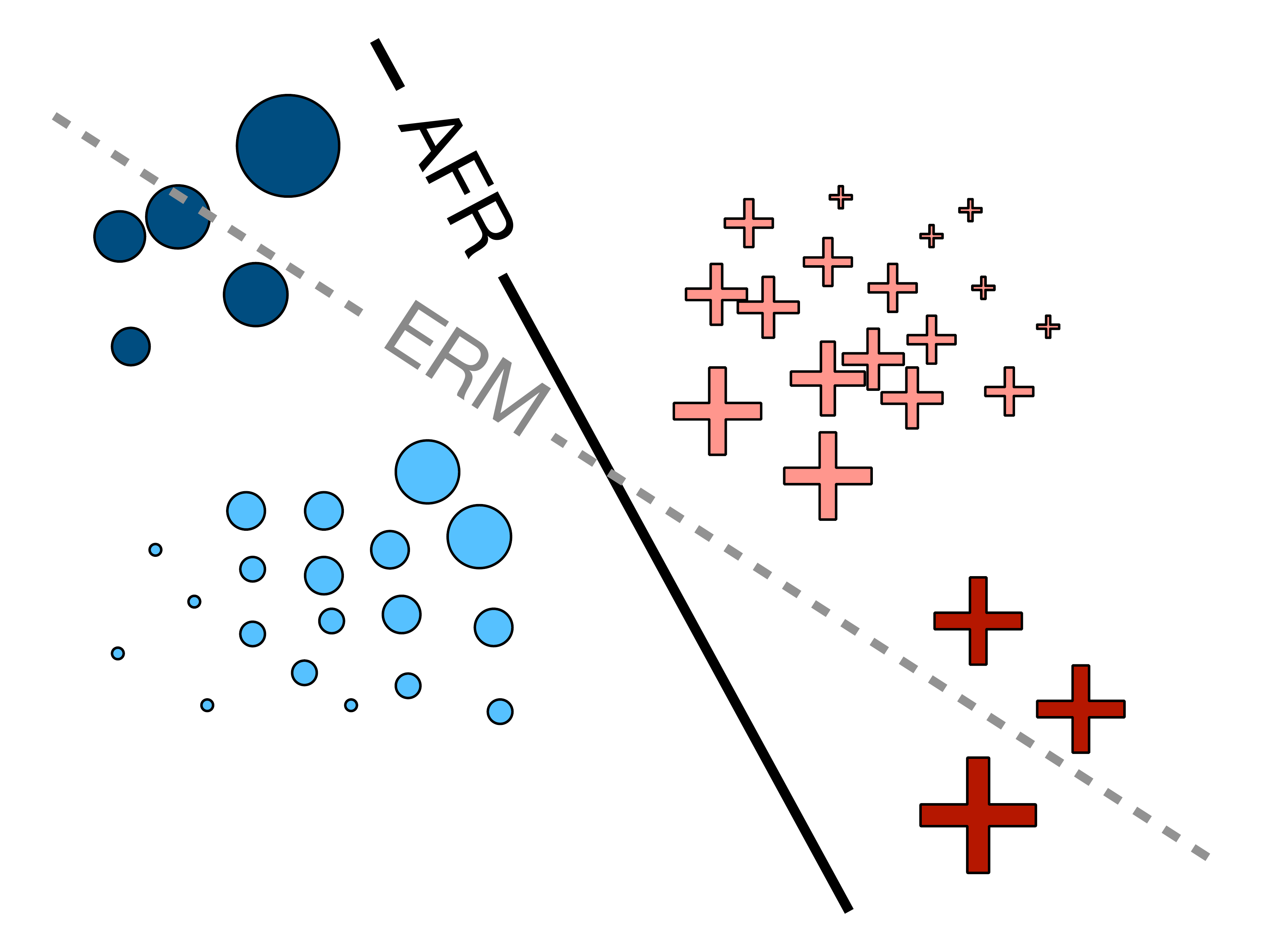} &
    \includegraphics[width=.43\linewidth]{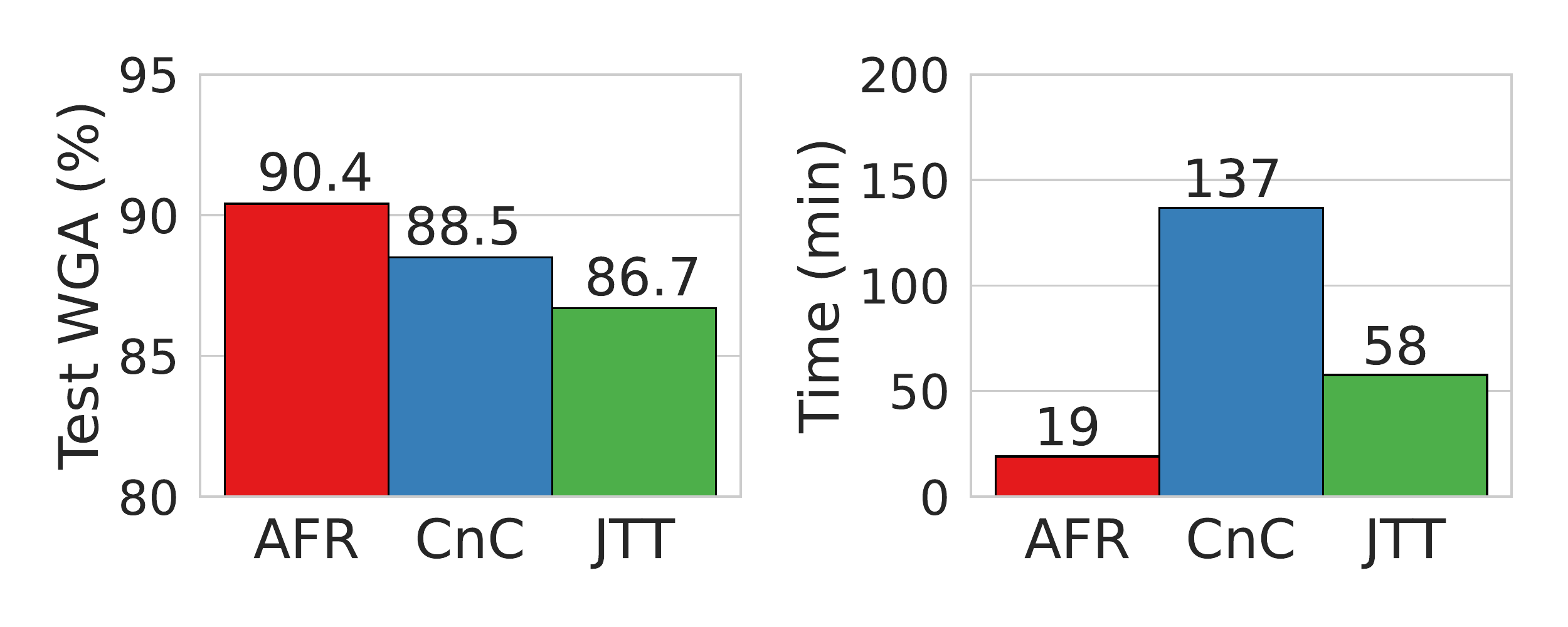}
      \\
    \hspace{-0.2cm}{\small (a) ERM classifier} &
    \hspace{-0.3cm}{\small (b) \method classifier} &
    \hspace{-0.4cm}{\small (c) Performance and training time on Waterbirds}
    \end{tabular}
   \caption{
   \textbf{Automatic Feature Reweighting.}
   \textbf{(a)}
   An illustration of the ERM classifier trained on a binary classification problem with a spurious correlation. The dataset consists of four groups shown in different colours, and two classes shown with circles and pluses.
   The ERM classifier (decision boundary shown with the black line) performs well on the majority groups $G_2, G_4$ (shown in lighter colors), but underperforms on the minority groups $G_1, G_3$ (lighter colors).
   \textbf{(b)}
   AFR upweights the datapoints where the base ERM model achieves high loss.
   The size of a marker reflects the weight of the corresponding datapoint.
   This weighting automatically prioritizes the minority data, leading to a classifier that achieves much better performance on the minority groups.
   \textbf{(c)}
   Test worst-group accuracy and training time
   comparison on Waterbirds for state-of-the-art methods that do not use group information during training.
   \method outperforms the baselines, while only requiring a small fraction of compute time.
   }
    \label{fig:intro}
\end{figure*}

In this paper, we propose \textit{Automatic Feature Reweighting} (\method), a simple and fast method for reducing the reliance on the spurious attributes with a minimal computational overhead compared to standard training. As in \citet{kirichenko2022dfr}, we freeze the feature extractor of the base model pretrained on a dataset with a spurious correlation, and focus on retraining the last layer of the model. However, unlike \citet{kirichenko2022dfr}, which uses a group-balanced dataset for re-training,
 we use an weighted loss on a held-out dataset \emph{drawn
simply from the training distribution} for retraining the last layer, where the weights prioritize datapoints on which the base model performs poorly.
These weights automatically result in an approximately group-balanced dataset for re-training without either group labels or intervention in base model training. We illustrate AFR in Figure~\ref{fig:intro}.

We show that \method achieves strong results on vision and language benchmarks such as Waterbirds, CelebA, MultiNLI, CivilComments and Chest X-Ray
while requiring a negligible training time overhead compared to ERM.
In particular, we improve upon the best reported results on Waterbirds and MultiNLI among methods which do not require labels for the spurious attributes during training.
Furthermore, we show that AFR can significantly outperform a state-of-the-art method that explicitly uses group labels during training, when only a small number of group labels are available.
Through extensive ablations, we find \method is robust to its hyperparameters such that it often significantly improves group robustness without careful hyperparameter tuning.

Code for AFR is available at \\ \url{https://github.com/AndPotap/afr}.

\section{Preliminaries} \label{sec:preliminaries}

We frame the group robustness problem and outline several baseline methods.

\subsection{Problem Setting}

We consider the group robustness  setting
\citep{sagawa2020GDRO}.
Specifically, we assume that the data distribution consists of several \textit{groups} $g \in \mathcal{G}$, which are typically defined by a combination of the class label $y \in \mathcal{Y}$ and a spurious attribute $s \in \mathcal{S}$.
For example, in CelebA where we classify $y \in \{\text{Blond}, \text{Not Blond}\}$, the spurious attribute is the gender of the person shown in the image
$s \in \{\text{Male}, \text{Female}\}$.
The groups then correspond to all pairs of values of the class label and spurious attribute $\mathcal{G} = \mathcal{Y} \times \mathcal{S}$.

The attribute $s$ is spurious if it is correlated with $y$, but not causally related with it.
In CelebA, about $94\%$ of the datapoints with $y = \text{Blond}$ have the spurious attribute $s = \text{Female}$.
As a result, models trained on this problem rely on the gender feature to predict the hair color, and underperform on the minority group $g=(\text{Blond}, \text{Male})$.
Throughout the paper, we will be evaluating the \textit{worst group accuracy} (WGA), i.e. the minimum of predictive accuracies of our model across all groups.

\textbf{Access to spurious attributes.}\quad
Most existing methods for robustness to spurious correlations assume access to spurious attributes $s$ on the training data \citep{sagawa2020GDRO}, or
on a subset of the available data used to train the parameters of the model \citep{ kirichenko2022dfr, nam2022spread, sohoni2021barack}.
In contrast, we consider the more challenging setting, where the group attributes are not available to us on any of the datapoints used to train the parameters of the model.
We underscore that, as all other methods following this setting \citep{liu2021jtt, zhang2022cnc}, we tune hyperparameters on a validation set with explicit spurious attributes.
However, as seen in Section \ref{sec:efficient}, our method does not require a large amount of validation data.

\subsection{Group Robustness Baselines}
\label{sec:baselines}

\noindent \textbf{Empirical Risk Minimization (ERM).} \quad
In ERM, we learn the parameters $\theta$ of the model by minimizing the loss $\ell$ averaged over the training data:
\begin{equation}
    \begin{split}
    \label{eq:erm}
      \mathcal{L}^{\mathrm{ERM}}\left(\theta\right)
      =
      \frac{1}{N} \sum_{i=1}^{N} \ell\left(x_{i}, y_{i} ; \theta\right).
    \end{split}
\end{equation}
We use cross-entropy
$\ell^{\mathrm{xe}}\left(x_i, y_i; \theta\right) = -\log p_{y_i}(x_i; \theta)
$, where $p_{y}(x; \theta)$ is the probability of the class $y$ predicted by the model with parameters $\theta$ on an input $x$.

\noindent \textbf{Group Distributionally Robust Optimization (GDRO).} \quad
Intuitively, GDRO \citep{sagawa2020GDRO}
minimizes the worst group loss given by
\begin{equation*}
    \begin{split}
      \mathcal{L}^{\mathrm{GDRO}}\left(\theta\right)
      =
      \max_{g \in \mathcal{G}}
      \frac{1}{N_{g}} \sum_{i: g_{i} = g}^{}
      \ell\left(x_{i}, y_{i} ; \theta\right)
    \end{split}
\end{equation*}
where $N_{g}$ is the number of observations with group $g_{i} = g$.
To compute the GDRO objective, we require group labels $g_n$ for all of the training data and these group labels help GDRO to considerably improve worst group performance. Therefore, the worst group accuracy achieved by GDRO is considered an upper bound on a given dataset.
\citet{sagawa2020GDRO} provides full details of the method.

\noindent \textbf{Deep Feature Reweighting (DFR).} \quad
\citet{kirichenko2022dfr} showed that it is sufficient to retrain just the
last layer of a classifier
on a group-balanced held-out \textit{reweighting} dataset to obtain results similar to GDRO.
Specifically, they represent the model $m_{\theta} = c_\phi \circ e_{\psi}$, where $e_\psi$ is the feature extractor with parameters $\psi$ and $c_{\phi}$ is the classification head (last layer) with parameters $\phi$, and $\theta = (\phi, \psi)$.
Starting with a pretrained model, we freeze the feature extractor parameters $\psi$, and retrain only the classifier $c_\phi$ using a held-out dataset (not used to train the feature extractor) where the number of datapoints from each group is the same.
They use the ERM objective in Eq.~\eqref{eq:erm} to retrain the classifier.
While DFR does not require group annotations on all of the training data like GDRO, it requires a group-balanced reweighting dataset.

\noindent \textbf{Just Train Twice (\textsc{jtt}).} \quad
In \textsc{jtt} \citep{liu2021jtt}, we first train a classifier $m_{\theta}$ with ERM for $T$ steps finding weights $\hat \theta$,
and then construct the set $\mathcal{E} = \set{i: \arg\max_{y} m_{\hat{\theta}} \left(x_{i}\right)[y] \neq y_i}$
of misclassified training indices.
We then retrain $m_{\theta}$ by minimizing the loss
\begin{equation*}
    \begin{split}
      \mathcal{L}^{\mathrm{JTT}}\left(\theta\right)
      =
      \frac{\lambda}{\left|\mathcal{E}\right|} \sum_{i \in \mathcal{E}}^{}
      \ell^{\mathrm{xe}}\left(x_{i}, y_{i};\theta\right)
      +
      \frac{1}{\left|\mathcal{E}^{c}\right|} \sum_{i \notin \mathcal{E}}^{}
      \ell^{\mathrm{xe}}\left(x_{i}, y_{i};\theta\right),
    \end{split}
\end{equation*}
where the hyperparameter $\lambda \in \mathbb{R}$ upweights the relevance of the missclassified observations.
\textsc{jtt} does not explicitly require access to the group labels.
Importantly, \textsc{jtt} tunes the number $T$ of epochs for which the first model is trained to avoid both under and over-fitting the training set. If $T$ is too small, the error set will approach the entire training set. If $T$ is too large, the error set will approach the empty set.

\noindent \textbf{Correct-\textsc{n}-Contrast (C\textsc{n}C).} \quad
The method in \citet{zhang2022cnc} builds upon JTT and also trains two models, but it uses a contrastive loss to train the second model.
The first model is used to define the positive pairs (objects with the same class, but different predictions), and negative pairs (objects with different classes but the same prediction).
C\textsc{n}C achieves strong WGA without requiring group annotations, but requires tuning and comes at a large computational overhead compared to standard training:
the authors report $10\times$ training time for C\textsc{n}C compared to ERM on multiple datasets. Similar to how \jtt requires tuning the number of epochs to train the first model, \cnc requires tuning the weight decay parameter for training the first model to avoid both under and over-fitting the training set in order to extract information on the spurious attributes from its prediction.

We provide an extended discussion of additional related work in Appendix \ref{Related Work}.

\section{Automatic Feature Reweighting}
We now introduce \method, a fast and simple method to improve group robustness without group annotations during training.
As we will show, \method is based on the key insight that we can infer minority group examples from an ERM model alone, trained in the standard way without intervention.
The ability to use a standard ERM model is an important advantage over \jtt and \cnc where a model is trained specifically to learn the spurious features by applying heavy regularization.
We use the inferred information about the minority examples to automatically construct a more group-balanced dataset and retrain only the last layer of the ERM model without access to any group labels, unlike DFR where group labels are required.

\subsection{Method Description}
At a high level, \method\ proceeds in two stages. In the first stage, we train an ERM model on the training set $\mathcal{D}_\mathrm{ERM}.$ In the second stage, we retrain only the last layer of the model on a reweighting set $\mathcal{D}_\mathrm{RW}$, with examples receiving high loss under the ERM checkpoint upweighted. Here, $\mathcal{D}_\mathrm{ERM}$ and $\mathcal{D}_\mathrm{RW}$ are two distinct datasets drawn from the same distribution and neither of them needs to be group-balanced.
We find that splitting the training set in a $80\%$--$20\%$ proportion to construct $\mathcal{D}_\mathrm{ERM}$ and $\mathcal{D}_\mathrm{RW}$ works well in practice, but we show that performance is not particularly sensitive to the split in Appendix~\ref{sec:ablate_ratio}.

We summarize \method in Algorithm \ref{alg:cwxe}.
We now describe the two stages in further detail.

\begin{algorithm}[tb]
  \caption{Automatic Feature Reweighting}
   \label{alg:cwxe}
   \begin{algorithmic}
     \STATE {\bfseries Input:} Training set partitioned into $\mathcal{D}_\mathrm{ERM}$ (80\%),
     $\mathcal{D}_\mathrm{RW}$ (20\%), $\gamma \geq 0$ and a classifier that decomposes as
     $m_{\theta}= c_\phi \circ e_\psi$
     where $\theta = \left(\phi, \psi\right)$.

     \textbf{Stage 1:} Train checkpoint $\hat{\theta}
     = (\hat{\phi}, \hat{\psi})$
     until convergence on $\mathcal{D}_\mathrm{ERM}$
     using the loss $\mathcal{L}^{\mathrm{ERM}}$.

     \textbf{Stage 2:} Re-train last layer $c_\phi$
     on $\mathcal{D}_\mathrm{RW}$
     using the loss $\mathcal{L}^{\mathrm{AFR}}$ in Eq.~\eqref{eq:cwxe},
     leaving the feature extractor $e_{\hat{\psi}}$ fixed.

   \end{algorithmic}
   \label{alg:ours}
\end{algorithm}

\textbf{Stage 1.}\quad
We train a model checkpoint on $\mathcal{D}_\mathrm{ERM}$ \emph{until convergence} via standard ERM without any modifications to the standard training procedure.
Typically in this stage, we would achieve near 100\% accuracy on the training data and the resulting model would have poor test performance on minority groups.
In contrast to \jtt and \cnc (see Section \ref{sec:baselines}),
we do not under-trained our first stage model to avoid completely fitting the training data, i.e., achieving near-zero training cross-entropy loss.

\textbf{Stage 2.}\quad
Denote the model parameters as $\theta = (\phi, \psi),$ where $\phi$ represents the last layer parameters and $\psi$ represents all the other parameters
and denote by $\hat{\theta} = (\hat{\phi}, \hat{\psi})$ their values learned at the end of stage 1.
In the second stage, we retrain the last layer parameters $\phi$ of the model on $\mathcal{D}_\mathrm{RW}$ using the following objective:
\begin{equation}\label{eq:cwxe}
    \begin{split}
      \mathcal{L}^{\mathrm{AFR}}\left(\phi\right)
      =
      \sum_{i=1}^{M}
      \mu_i \ell^{\mathrm{xe}}\left(x_{i}, y_{i}; \phi, \hat{\psi} \right)
      +
      \lambda \norm{\phi - \hat{\phi}}_{2}^{2}, \\
    \end{split}
\end{equation}
where $M = |\mathcal{D}_\mathrm{RW}|$,
and $\{\mu_i\}_{i=1}^{M}$ are per-example weights for the cross-entropy loss $\ell^{\mathrm{xe}}$.
These weights are designed to be large for datapoints where the original model $\hat \theta$ predicts poorly, thus automatically upweighting the minority groups.
We consider a simple functional form for the weights given by a softmax over the per-example ``incorrectness" $1-\hat{p}_i,$ where $\hat{p}_i$ is the probability for the correct class $y_i$ assigned by the stage 1 checkpoint $\hat\theta$, with a tunable inverse temperature $\gamma \geq 0$:
$\mu_i \propto \exp(\gamma \, (1 - \hat{p}_i)) \propto \exp(-\gamma \, \hat{p}_i)$.
To address datasets with class imbalance, we further modulate the per-example weight with a class-dependent constant and define
\begin{equation}
    \label{eq:weights}
    \begin{split}
      \mu_i = \frac{\beta_{y_i} \exp(-\gamma \, \hat{p}_i)}{\sum_{j=1}^{M} \beta_{y_j} \exp(-\gamma \, \hat{p}_j)},
    \end{split}
\end{equation}
where $\beta_{y}$ is one divided by the number of examples belonging to class $y$ in the reweighting set. Note that the weights $\mu_i$ in Eq.~\eqref{eq:weights} are only computed once and fixed during the second stage.

The hyperparameter $\gamma$ specifies how much to upweight examples with poor predictions, while $\lambda \geq 0$ is a regularization hyperparameter to prevent the last layer from focusing only on minority examples at the expense of degrading performance on majority group examples to an unacceptable level. The regularization also reduces overfitting to limited reweighting data. In Section~\ref{sec:optimal}, we investigate how optimal our weighting function is for group robustness.

\begin{figure}[!t]
\centering
    \includegraphics[width=0.7\linewidth]{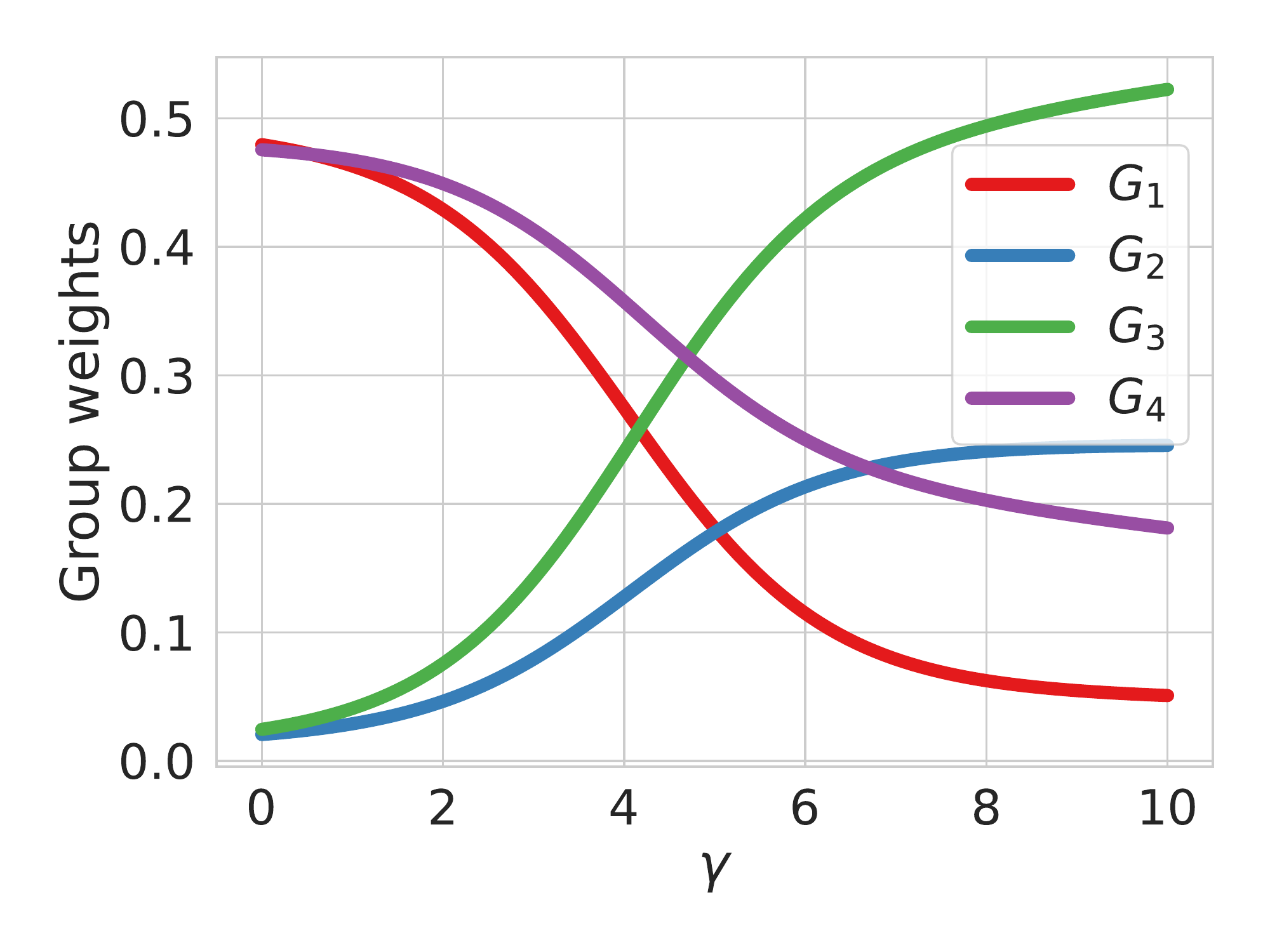}
   \caption{
   \textbf{\method weights on Waterbirds.}
   Group aggregated weights as a function of $\gamma$ on $\mathcal{D}_\mathrm{RW}$ for the Waterbirds dataset, with majority groups $G_1$ and $G_4$, and minority groups $G_1$ and $G_2$. The ERM model performs poorly on minority groups $G_2$ and $G_3,$
   enabling automatic group re-balancing by upweighting poorly predicted examples.
   The reweighted group distribution is more balanced than the original group distribution,
   for a broad range of $\gamma$ values.
   }
    \label{fig:wb_group_weights}
    \vspace{-4mm}
\end{figure}

\textbf{Effect of $\gamma$.}\quad
The hyperparameter $\gamma$ determines how strongly we upweight the datapoints where the first stage model $\hat \theta$ provides poor predictions.
In particular, by setting $\gamma = 0$, we recover standard ERM (with class reweighting), as all the weights in Eq.~\eqref{eq:weights} become simply
$\mu_i \propto \beta_{y_i}$.
In contrast, as we increase $\gamma$, \method's loss concentrates more on the poorly classified points.

To illustrate how we automatically upweight minority examples and reduce group imbalance with the weights $\{\mu_{i}\}_{i=1}^{M},$ in Figure~\ref{fig:wb_group_weights} we plot the group aggregated weights defined as the sum $\sum_{i: g_i = g} \mu_i$ of all per-example weights within each group $g$, as a function of $\gamma$ on $\mathcal{D}_\mathrm{RW}$ for the Waterbirds dataset.
As we increase $\gamma,$ the minority groups ($G_2$ and $G_3$) receive increasingly higher weights. The re-weighted group distribution is more balanced than the original group distribution, regardless of the value of $\gamma$, as long as it is positive.
At $\gamma = 4.4,$ the group aggregated weights are $23\%, 15\% , 31\%$ and $31\%,$ which is close to group balanced weights of $25\%$ each.

In the following sections, we expand on key design decisions in AFR and important conceptual differences between AFR and existing work that underlies AFR's strong performance, simplicity, and efficiency.

\subsection{Inferring Minority Examples without Intervention}

One key insight that distinguishes \method from \jtt and \cnc is that we do not need to intervene on the ERM model (e.g. by  over-regularizing) to infer spurious attributes or to identify minority group examples.
While an ERM model trained via the standard procedure can fit all datapoints in its training set $\mathcal{D}_\mathrm{ERM}$, its performance on minority group examples in a held-out dataset will be much worse compared to the majority group examples.
Indeed, this performance discrepancy is precisely why spurious correlations are problematic in the first space. In AFR, we take advantage of this performance discrepancy to automatically isolate and upweight the minority group examples on the reweighting set $\mathcal{D}_\mathrm{RW}$  during the second stage to retrain only the last layer.

In contrast, both \jtt and \cnc rely on heavy regularization to encourage the model to under-fit the data and only learn spurious features in the first stage.
These methods are thus more expensive and harder to tune, as the entire model has to be retrained from scratch in order to find, for example, the optimal regularization strength.
Moreover, an entirely new model needs to be trained in the second stage to learn the core features relevant to the actual classification problem, resulting in substantial computational overhead over training a single ERM model.

A potential downside of splitting the training data into $\mathcal{D}_\mathrm{ERM}$ and $\mathcal{D}_\mathrm{RW}$ is that we only use a subset of the available data for training the feature extractor. As such, we may learn a slightly weaker feature extractor compared to a standard ERM model trained with the entire training set. However, by reweighting the last layer with the remaining data $\mathcal{D}_\mathrm{RW},$ \method\ is able to achieve much higher worst group accuracies than ERM, showing that the benefit of the procedure drastically outweighs its cost when the goal is to achieve group robustness. In Appendix~\ref{sec:ablate_ratio} we also show \method\ is not very sensitive to the choice of this splitting ratio.

\subsection{Last Layer Retraining without Group Labels}
Retraining only the last layer on the reweighting set is an important design decision in \method\ that has several important benefits. First, as a standard ERM model learns the core features sufficiently well despite spurious correlations, retraining only the last layer is sufficient to significantly improve group robustness \citep{kirichenko2022dfr, lee2022surgical}.
Second, as the last layer tends to have several orders of magnitude fewer parameters than the entire model, retraining only the last layer is more data efficient than retraining the entire model.
This data efficiency enables us to use a small reweighting set and leave most of the training data to train the feature extractor during the first stage. Finally, retraining the last layer is extremely fast. By caching the embeddings from the fixed feature extractor, we can complete the last layer retraining in less than a minute on standard hardware.

In contrast to DFR, our reweighting set does not need to be group-balanced and can simply be drawn from the training distribution.
The weighting in Eq.~\eqref{eq:weights} automatically constructs a more balanced dataset. As such, AFR can tackle problems where no group-annotated data exists, whereas DFR cannot.

\subsection{\method's Advantages over Alternatives}

To the best of our knowledge, \method is the only group robustness method to achieve state-of-the-art results across a wide variety of benchmarks without requiring group labels through training, and with negligible computational overhead over standard ERM, as we will show in Section~\ref{sec: experiments}. We summarize AFR's advantages over prior group robustness methods as follows:

\begin{itemize}
\item \textbf{More broadly applicable:} Unlike methods such as DFR and GDRO, AFR does not require group labels during training and is therefore applicable to problems where DFR and GDRO are not due to unavailability of group annotated data.

\item \textbf{Simpler to use:} To run \method, we only need a standard ERM checkpoint, potentially from an existing pretrained model.
We then only retrain its last layer using a weighted cross-entropy loss and $\ell_2$ regularization to the initial weights.
Unlike \jtt and \cnc, AFR does not require any intervention for the first stage of training, or implementing special training objectives and batch samplers for the second stage.

\item \textbf{Faster to train:} \method only retrains the last layer parameters, often taking less than a minute by caching the embeddings $\{e_{\hat{\psi}}(x)\}_{x\in \mathcal{D}_\mathrm{RW}},$ while \jtt and \cnc train a second model entirely from scratch, which can take hours. We show a comparison of training times in Figure~\ref{fig:time}.

\item \textbf{Easier to tune:} Since \method has no hyperparameter for training the first checkpoint, unlike \jtt and \cnc, hyperparameter tuning is much easier and cheaper for \method.
Indeed, as a single first stage checkpoint can be reused, the time for sweeping over $K$ hyperparameter settings is roughly $O(1)$ for \method but $O(K)$ for \jtt and \cnc since training time for the last layer is negligible compared to retraining the entire model.
\end{itemize}

These advantages reflect important conceptual differences between AFR and alternatives as outlined in previous sections, despite a few high level similarities to methods like \jtt and DFR, such as two-stage training.

\section{Experiments}
\label{sec: experiments}

We evaluate \method on a range of benchmarks and provide detailed ablations on design decisions and hyperparameters.

\subsection{Datasets, Models and Hyperparameters}
\label{sec:exp_setup}
We now describe the datasets, models and hyperparameters that we use in the
experiments.

\textbf{Datasets.} \quad
We consider several image and text classification problems.
For more details, see Appendix \ref{sec:app_details}.

\setlength{\tabcolsep}{2.4pt}
\begin{table*}[t]
\begin{center}
  \footnotesize{
\begin{tabular}{c ccc ccc ccc ccc ccc}
\hline\noalign{\smallskip}
\multirow{2}{*}{\textbf{Method}}
  & \multicolumn{2}{c}{\textbf{Waterbirds}}
  && \multicolumn{2}{c}{\textbf{CelebA}}
  && \multicolumn{2}{c}{\textbf{MultiNLI}}
  && \multicolumn{2}{c}{\textbf{CivilComments}}
  && \multicolumn{2}{c}{\textbf{Chest X-Ray}}
\\
\\[-3mm]
  & Worst(\%) & Mean(\%)
  && Worst(\%) & Mean(\%)
  && Worst(\%) & Mean(\%)
  && Worst(\%) & Mean(\%)
  && Worst(\%) & Mean(\%)
\\[1mm]\hline\\[-3mm]
ERM &
    $72.6$ & $97.3$
    &&
    $47.2$ & $95.6$
    &&
    $67.9$ & $82.4$
    &&
    $57.4$ & $92.6$
    &&
    $4.4$ & $95.3$
\\\hline\\[-3mm]
\textsc{jtt} &
    $86.7$ & $93.3$
    &&
    $81.1$ & $88.0$
    &&
    $72.6$ & $78.6$
    &&
    $\mathbf{69.3}$ & $\mathbf{91.1}$
    &&
    $52.3$ & $64.2$
    \\

C\textsc{n}C &
    $88.5_{\pm0.3}$ & $90.9_{\pm0.1}$
    &&
    $\mathbf{88.8_{\pm0.9}}$ & $89.9_{\pm0.5}$
    &&
    $-$ & $-$
    &&
    $68.9_{\pm2.1}$ & $81.7_{\pm0.5}$
    &&
    $-$ & $-$
    \\
\method (Ours) &
    $\mathbf{90.4_{\pm1.1}}$ & $\mathbf{94.2_{\pm1.2}}$
    &&
    $82.0_{\pm0.5}$ & $\mathbf{91.3_{\pm0.3}}$
    &&
    $\mathbf{73.4_{\pm0.6}}$ & $\mathbf{81.4_{\pm0.2}}$
    &&
    $68.7_{\pm 0.6}$ & $89.8_{\pm 0.6}$
    &&
    $\mathbf{56.0_{\pm3.4}}$ & $\mathbf{70.1_{\pm6.0}}$
\\\hline\\[-3mm]
  Group-DRO$^{\dagger}$ &
    $91.4$ & $93.5$
    &&
    $88.9$ & $92.9$
    &&
    $77.7$ & $81.4$
    &&
    $69.9$ & $88.9$
    &&
     $-$ & $-$
    \\
DFR$^{\dagger}$ &
    $92.9_{\pm0.2}$ & $94.2_{\pm0.4}$
    &&
    $88.3_{\pm1.1}$ & $91.3_{\pm0.3}$
    &&
    $74.7_{\pm0.7}$ & $82.1_{\pm0.2}$
    &&
    $70.1_{\pm0.8}$ & $87.2_{\pm0.3}$
    &&
    $59.8_{\pm 1.8}$ & $64.2_{\pm 3.1}$
\\[1mm]\hline
\end{tabular}
}
\end{center}
\caption{
\textbf{Results on spurious correlation benchmarks.}
We report test worst-group accuracy and test mean accuracy.
Additionally, $^{\dagger}$ denotes oracle methods that make explicit use of group annotations.
For \method, we report the mean $\pm$ std over $3$ independent runs.
We report \cnc numbers from \citet{zhang2022cnc}, ERM and \jtt numbers from \citet{liu2021jtt} and DFR numbers from \citet{kirichenko2022dfr}.
For Chest X-Ray, we report the \jtt number from \citet{yang2022chromavae} and the ERM number from our own run.
For mean accuracy, we follow \citet{liu2015deep} and \citet{sagawa2020GDRO} and weight the group accuracies according to their prevalence in the training data.
\method provides competitive results with substantially lower runtime (shown in Figure~\ref{fig:time}) than alternatives that do not use group information during training.
}
\label{tab:benchmark_results}
\end{table*}
\setlength{\tabcolsep}{2.4pt}

\begin{itemize}
  \item \textbf{Waterbirds} \citep{sagawa2020GDRO} is an image classification dataset where the goal is to classify birds into
    landbirds (woodpecker, catbird, etc) and waterbirds (albatross, seagull, etc).
    The background is a spurious feature, as most waterbirds are shown on water backgrounds and most landbirds are shown on land.

  \item \textbf{CelebA} \citep{liu2015deep} is an image classification dataset, where we focus on classifying whether an image shows a person with blond hair or not.
    Gender is the spurious feature, as $94\%$ of images showing people with blond hair in CelebA are of women, and
    blond men constitute a minority group.

   \item \textbf{MultiNLI} \citep{williams2017broad} is a text classification problem, where
  we discern whether the second sentence in a given pair of sentences is entailed by, contradicts or is neutral to the first sentence.
The spurious attribute is the presence of negation words such as ``never" which correlates with ``contradiction".

   \item \textbf{CivilComments} \citep{borkan2019nuanced} is a text classification benchmark, where we classify
    an online comment as ``toxic" or ``not toxic''.
    We use the version from the WILDS benchmark \citep{koh2021wilds}, where the presence of text related to gender (male, female),
    sexual orientation (LGBTQ), race (black, white) and religion (Christian, Muslim or other) is spuriously correlated with the comment being labeled as ``toxic".

  \item \textbf{CXR}
 \citep{yang2022chromavae} is an image classification dataset where we
 consider the task of predicting pneumonia based on chest X-rays from two sources: CheXpert \citep{irving2019chexpert} and NIH \citep{wang2017chestx}.
 Here the spurious feature is the source of the image (machine-specific artifacts).
\end{itemize}

\textbf{Models.} \quad
We follow standard model choices consistent with the baselines on each of the datasets \citep{liu2021jtt, kirichenko2022dfr, zhang2022cnc}:
for Waterbirds and CelebA we use ResNet-50 \citep{he2016deep} pretrained on ImageNet1k \citep{russakovsky2015imagenet},
for MultiNLI and CivilComments we use BERT \citep{devlin2018bert} pre-trained on Book Corpus and English Wikipedia data,
and for Chest X-Ray we use DenseNet-121 (pretrained on ImageNet1K).
We provide further details in Appendix \ref{sec:app_details}.

\textbf{Hyperparameters.} \quad
Following all other group robustness methods, we use validation WGA to tune \method's $\gamma$ and $\lambda$ and perform early stopping (see Appendix \ref{sec:app_details} for details).

\subsection{AFR Achieves Strong Group Robustness} \label{sec:results}
We report the results for \method\ and the baselines in Table~\ref{tab:benchmark_results}, showing that \method\ outperforms or that it is competitive with the best reported results among methods trained without access to the spurious attributes on all datasets except CelebA.
\method\ improves upon the best reported results on Waterbirds and MultiNLI. Compared to \jtt, \method achieves better performance on all datasets except for CivilComments where the results are slightly worse. Compared to \cnc, \method achieves better results on Waterbirds, similar results on CivilComments, and underperforms on CelebA.
In Appendix \ref{sec:cifar} we present additional experiments where we show that \method can be safely applied to data without known spurious features
without losing performance.

\subsection{AFR is Much Faster Than Alternatives}
In Figure \ref{fig:time}, we visualize the training time required by \method and baselines on each of the datasets (see Appendix~\ref{sec:times_details} for more details). Unsurprisingly, \method\ incurs a negligible overhead (6\% on average) compared to ERM by only retraining the last layer. Here we included the time to pre-compute and cache the embeddings, which in practice can be ignored by amortizing across a hyperparameter sweep.
In contrast, \jtt is on average three times as expensive (205\% overhead compared to ERM).
It might appear counter-intuitive that \jtt takes $3\times$ the time of ERM, as we are ``just training twice''!
However, both stages of \jtt, especially the last one, are more expensive than ERM.
First, the
initial checkpoint takes longer to run as it converges slower due to high regularization (training to convergence is needed in order to tune the first stage epoch number).
Second, the second phase has substantially more data which increases the epoch runtime, as all misclassified examples are repeated multiple times in the dataset.
Finally, \cnc is the least scalable method
with an average overhead of 550\% compared to ERM.
\cnc shares similar steps to \jtt
but the contrastive objective is expensive to optimize.

\begin{figure}[!t]
\centering
    \includegraphics[width=0.85\linewidth]{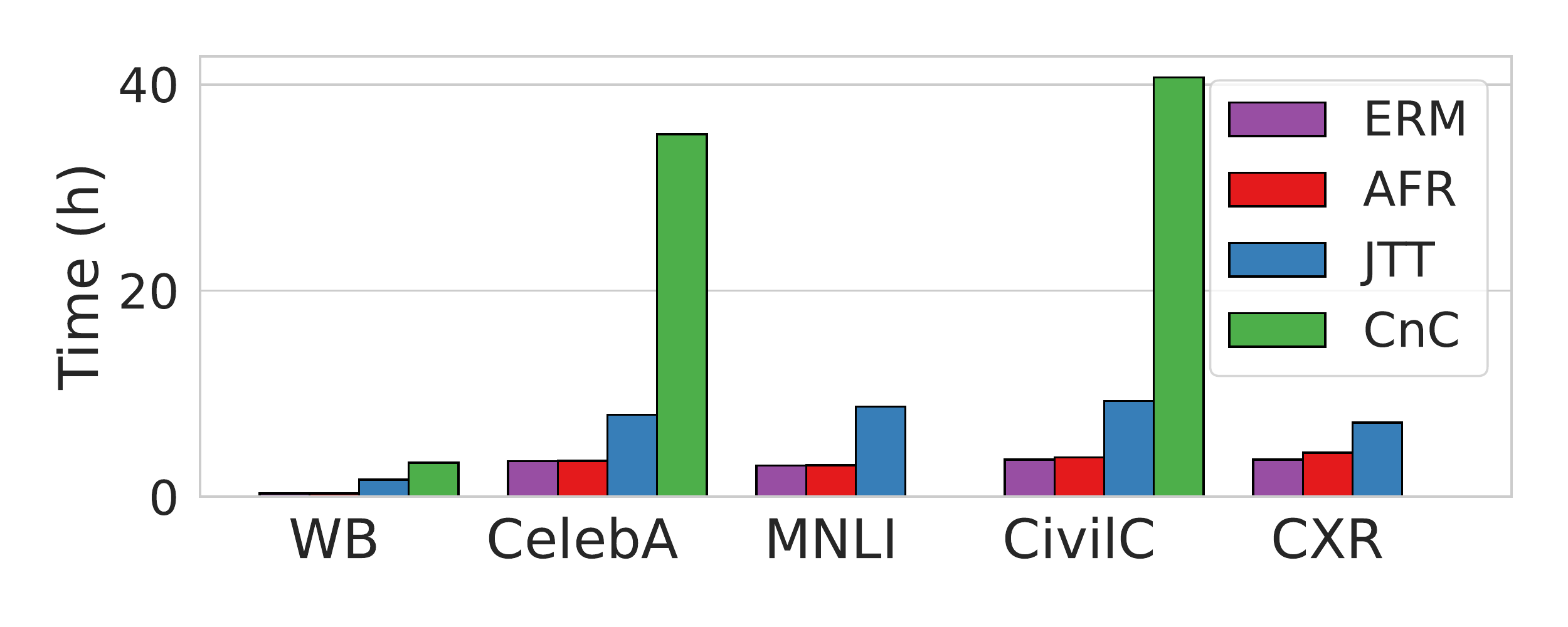}
    \caption{
    \textbf{Training time.}\quad
    Training time (in hours) across different spurious correlation benchmarks.
    \method only incurs a negligible overhead compared to standard ERM
    and is substantially less expensive than \jtt and \cnc.
    }
    \label{fig:time}
    \vspace{-4mm}
\end{figure}

\subsection{AFR is Group Label Efficient}
\label{sec:efficient}

\begin{figure*}[!t]
    \centering
    \subfloat[Waterbirds]{
    \includegraphics[width=0.5\columnwidth]{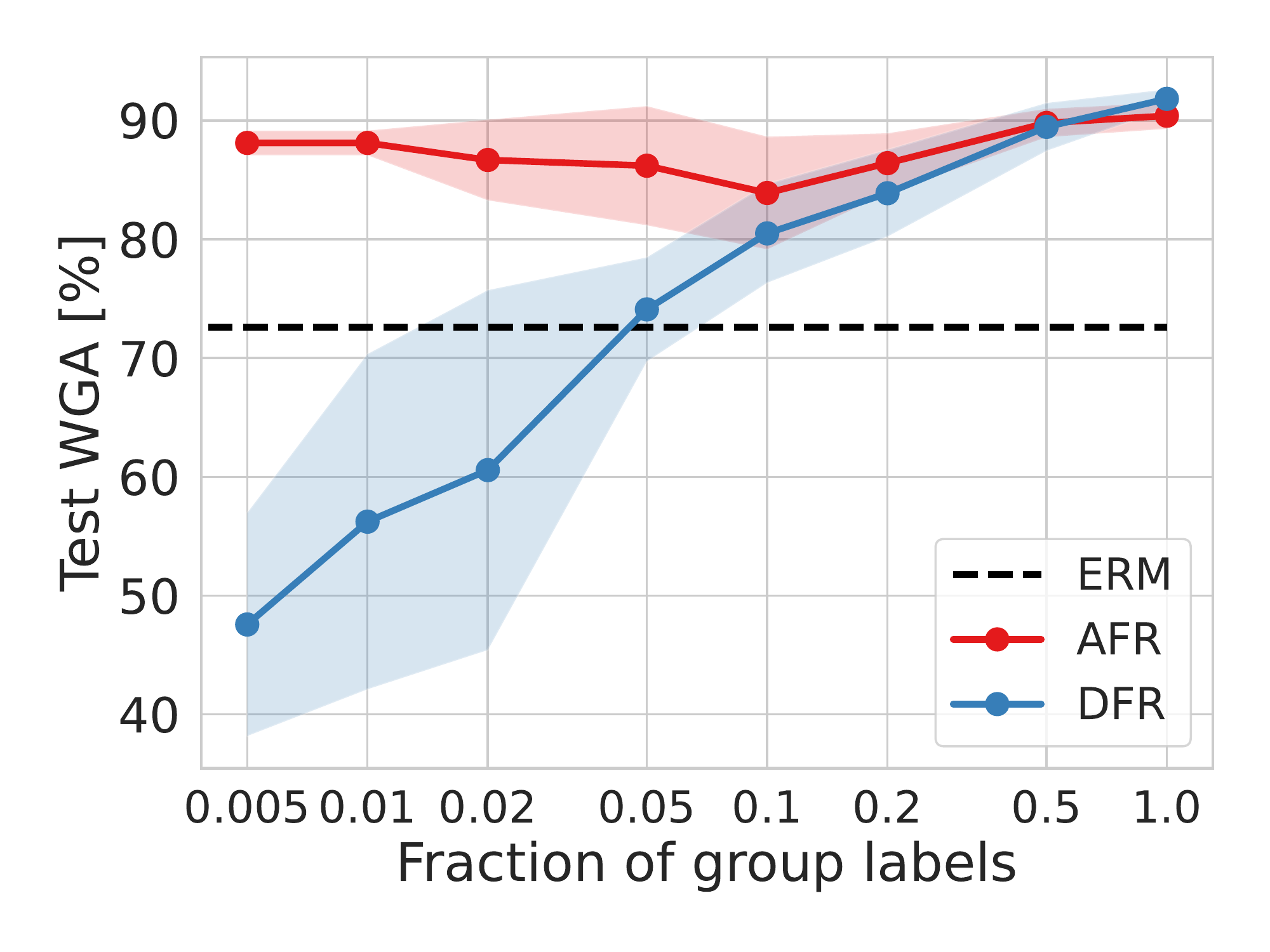}
    }
    \subfloat[CelebA]{
    \includegraphics[width=0.5\columnwidth]{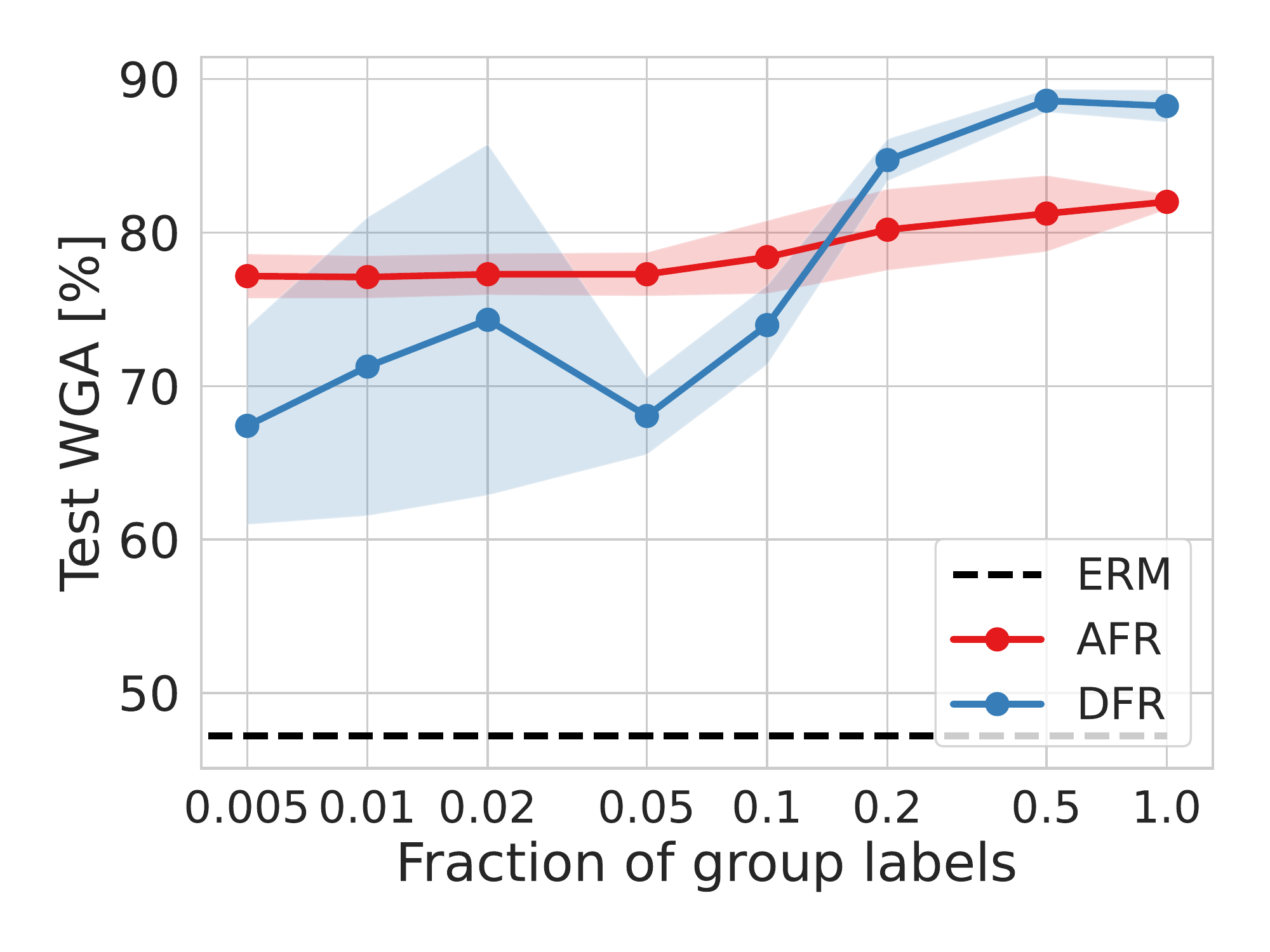}
    }
    \subfloat[MultiNLI]{
    \includegraphics[width=0.5\columnwidth]{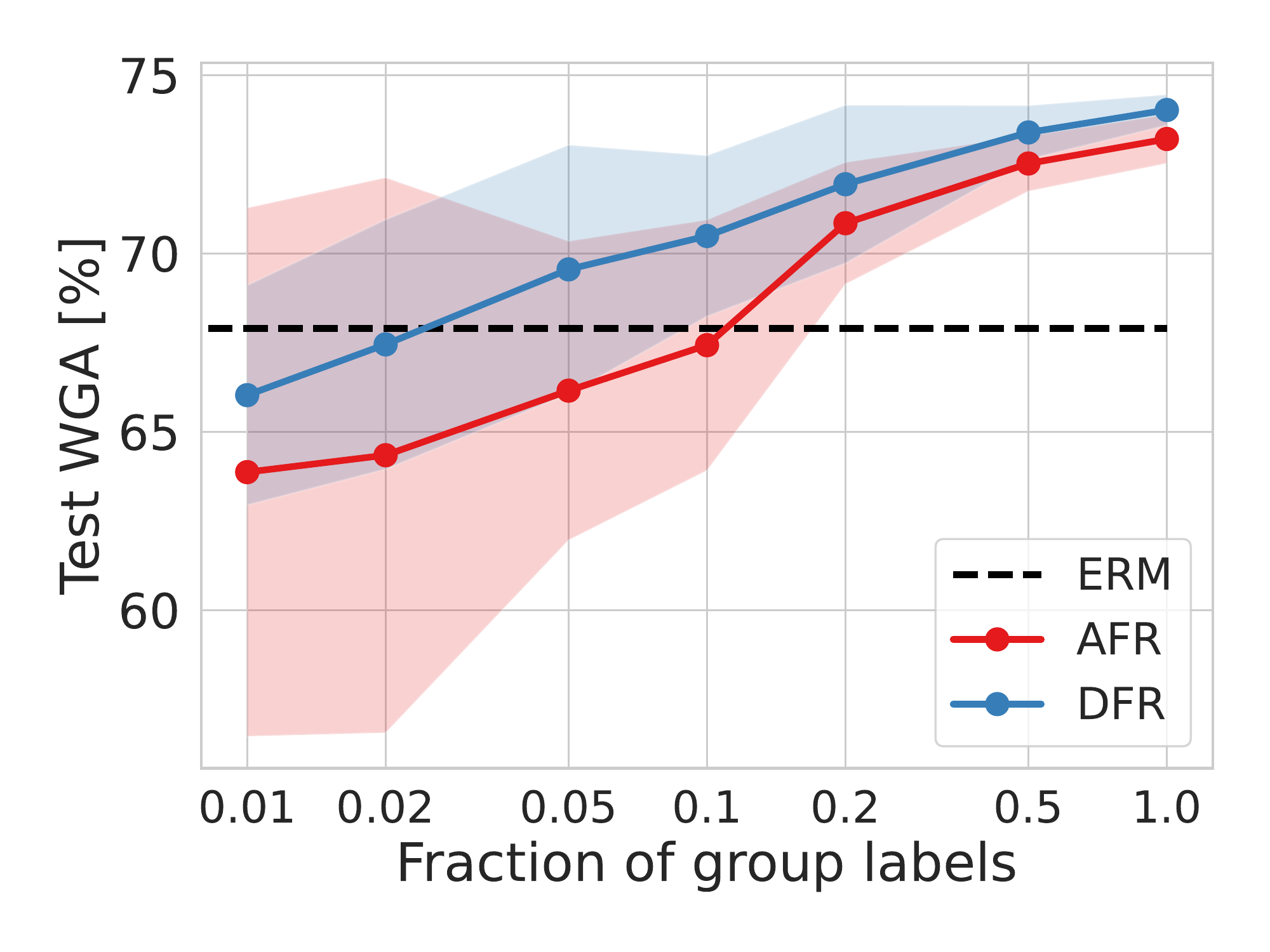}
    }
    \subfloat[CivilComments]{
    \includegraphics[width=0.5\columnwidth]{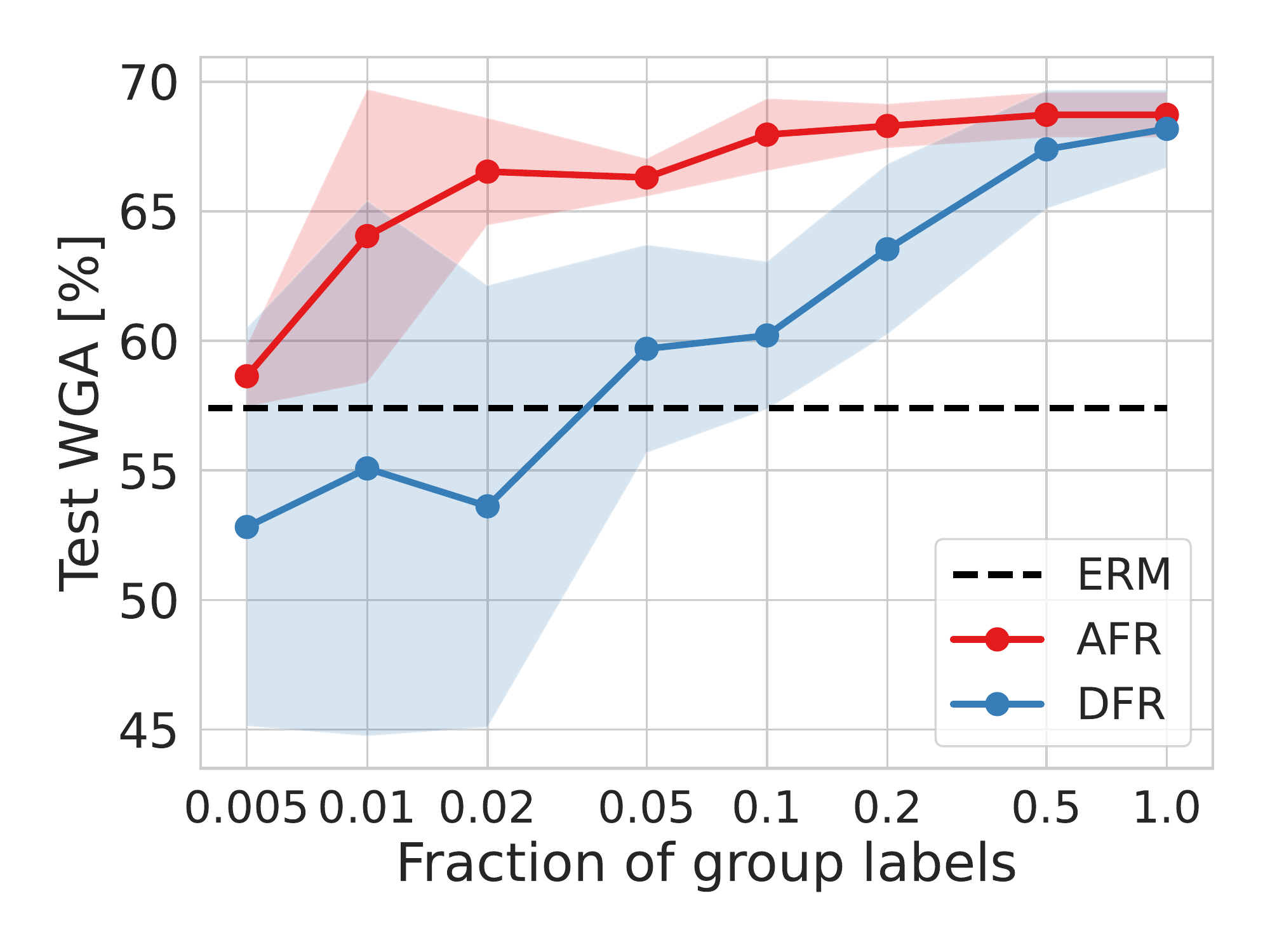}
    }
    \caption{
   \textbf{Group label efficiency of AFR vs DFR.} AFR significantly outperforms DFR on 3 out of 4 datasets when only a small fraction of group-annotated validation data is used by AFR and DFR for hyperparameter tuning and training respectively. We show means and standard deviations computed over 9 runs, with 3 seeds for base model training and 3 seeds for  validation set down-sampling.}
   \label{fig:group-label-efficiency}
   \vspace{-4mm}
\end{figure*}

Efficiency in terms of group labels is valuable for group robustness methods, because correctly identifying the spurious attributes and collecting corresponding group labels can be a time-consuming and expensive process.
Unlike methods such as DFR and Group DRO, AFR can be applied when group labels are not available on the training set.
On the other hand, while AFR is generally not sensitive to the choice of hyperparameters, as we show in Section~\ref{sec:ablations} and Appendix~\ref{sec:app_ablations}, optimal performance still requires hyperparameter tuning on a group-annotated validation set, and the same applies to \jtt and \cnc.
However, since AFR only uses group labels for hyperparameter tuning and not for training, it is reasonable to expect AFR to be more efficient in terms of group labels than methods like DFR.
In Figure~\ref{fig:group-label-efficiency}, we show that AFR can indeed achieve significantly higher test WGA than DFR when only a small fraction of group-annotated validation data are accessible to both methods.
On Waterbirds, for example, AFR achieves near-maximal test WGA improvement with only 0.5\% of the group-annotated validation data, consisting of merely 5 examples, whereas DFR achieves lower test WGA than ERM by overfitting to these examples.
On the other hand, since DFR directly uses group labels to train the model, rather than only to tune the hyperparameters, it can eventually outperform AFR as more group labels become available, as shown in the case of CelebA.

\begin{figure}
    \centering
    \subfloat[Test WGA]{
    \includegraphics[width=0.5\columnwidth]{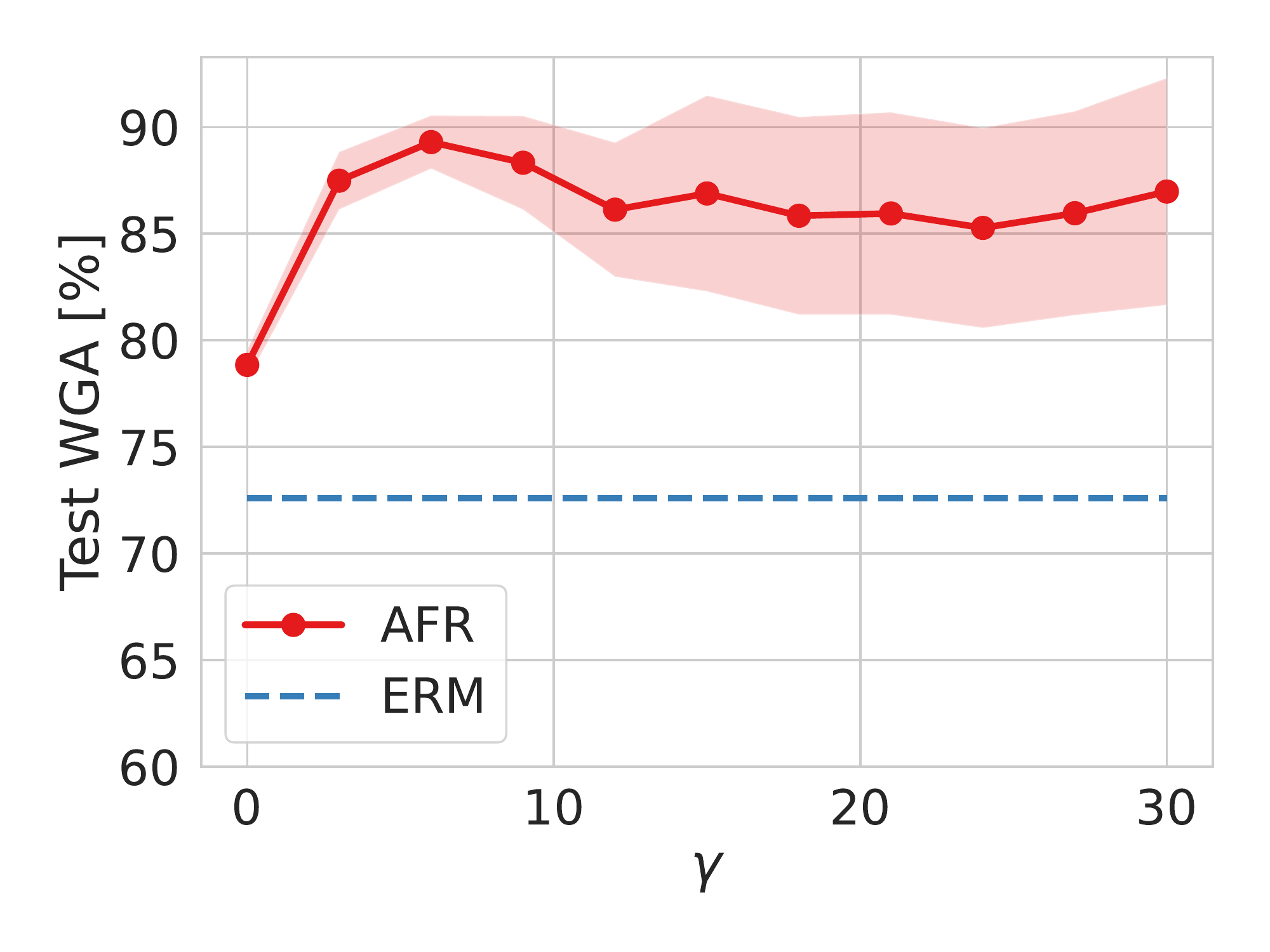}
    \label{fig:wb_wga_vs_gamma}
    }
    \subfloat[Effective sample size]{\includegraphics[width=0.5\columnwidth]{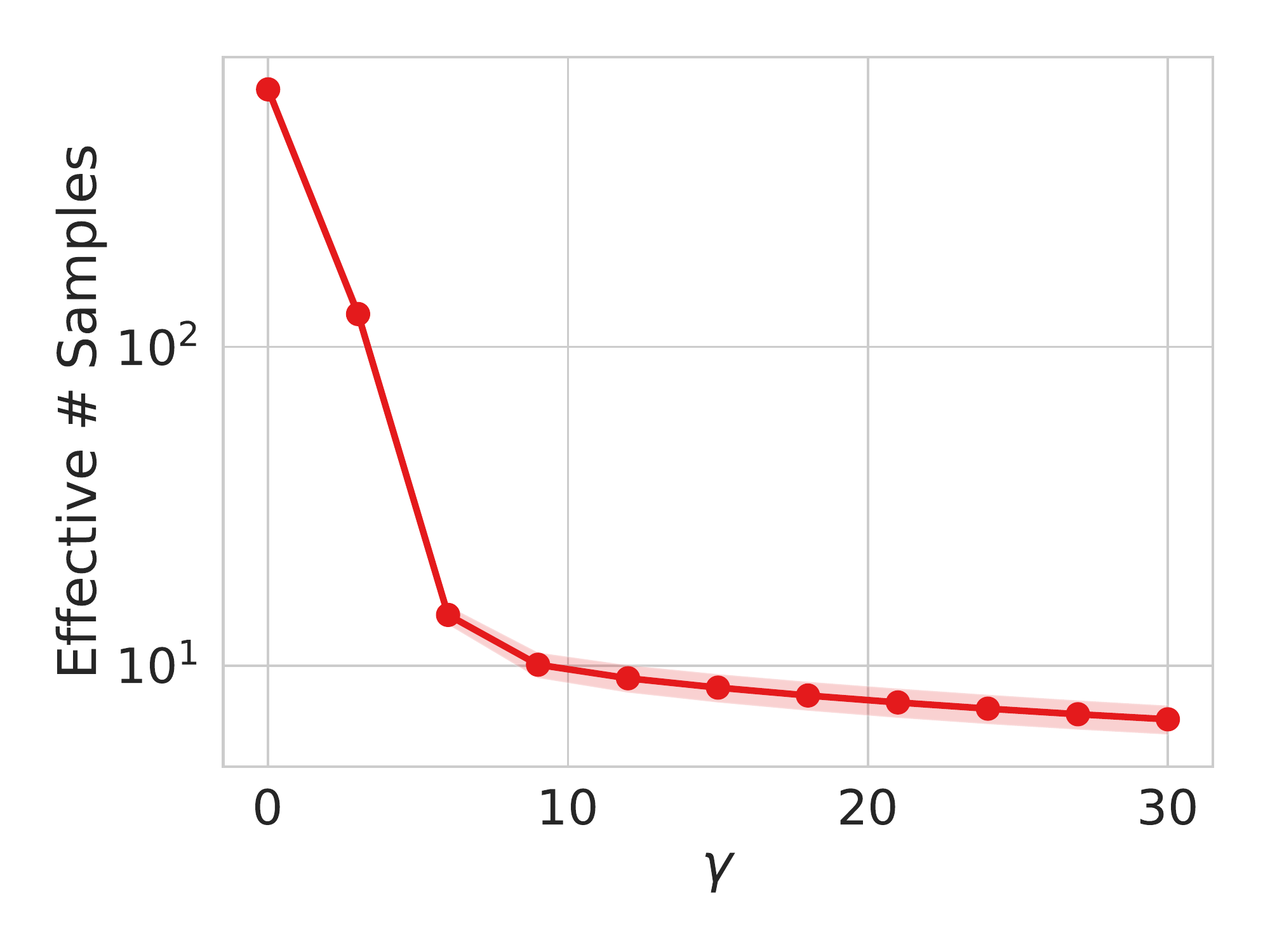}
    \label{fig:wb_effsize_vs_gamma}
    }
    \caption{
    \textbf{Robustness to $\gamma$.}
    \textbf{(a)} \method\ significantly improves test WGA on Waterbirds for any $\gamma \in [0,30].$
    We show mean and standard deviation (error bars) across 3 runs.
    \textbf{(b)} The effective sample size decreases and then stabilizes as $\gamma$ grows, explaining stability in AFR's performance as $\gamma$ increases.
    }
    \label{fig:gamma}
    \vspace{-4mm}
\end{figure}

As a result, we conclude that AFR is often more efficient in terms of group labels than DFR when the number of available group labels is limited, though not necessarily when group labels are abundant.

\subsection{AFR is Robust to Hyperparameters}
\label{sec:ablations}
We study the robustness of \method to the choice of the hyperparameter
$\gamma$ for Waterbirds.
Furthermore, in Appendix~\ref{sec:app_ablations}, we present additional results ablating early stopping, $\ell_2$ regularization to initial checkpoint, dataset splitting ratio, and the functional form of the weights, demonstrating AFR is also robust to these hyperparameters and design decisions.

As argued earlier, we do not expect the performance to be highly sensitive to $\gamma,$ since as long as $\gamma$ is positive, the reweighted group distribution will be more balanced than the original group distribution.
We confirm this intuition in Figure~\ref{fig:gamma}(a), which shows the mean and standard deviation of the test worst group accuracy achieved by our method over 3 runs on Waterbirds across a wide range of $\gamma,$ keeping $\lambda = 0$ for simplicity.
AFR achieves near-optimal test WGA by the time $\gamma=6.$ As we continue to increase $\gamma$, the test WGA degrades slightly and then stabilizes around a value still much higher than ERM.
To understand this behavior, note that for a sufficiently large $\gamma,$ the set of examples that receive non-negligible weights $\mu_i$ converges approximately to a fixed set, namely, the set of most poorly predicted examples under the first stage checkpoint.
This property is evident in Figure~\ref{fig:gamma}(b), where we show the effective sample size $N_\mathrm{eff}$ approximately stabilizes after initially plummeting for $\gamma < 6$.
Here, $N_\mathrm{eff} = 1 / \sum_{i=1}^{M} \mu_{i}^{2}$ represents the effective number of observations for weighted samples \citep{kish65}.
As a result, the performance of AFR doesn't vary significantly beyond $\gamma=6$, revealing only a gradual decrease in mean and an increase in variance as the effective sample size slowly diminishes.
Overall, AFR maintains a significant improvement over ERM for all $\gamma \in [0,30].$
In Appendix~\ref{sec:other_gamma}, we show AFR is even more robust to $\gamma$ on CelebA where its performance is almost constant over $\gamma.$

Setting $\gamma = 0$ already yields a significant increase in test WGA over standard ERM.
This improvement is not surprising since the per-example weights $\mu_i$ in \method account for class imbalance and the smallest group in Waterbirds indeed belongs to the less represented class.

\begin{figure*}
    \centering
    \subfloat[Group aggreagated weights]{
    \includegraphics[width=0.5\columnwidth]{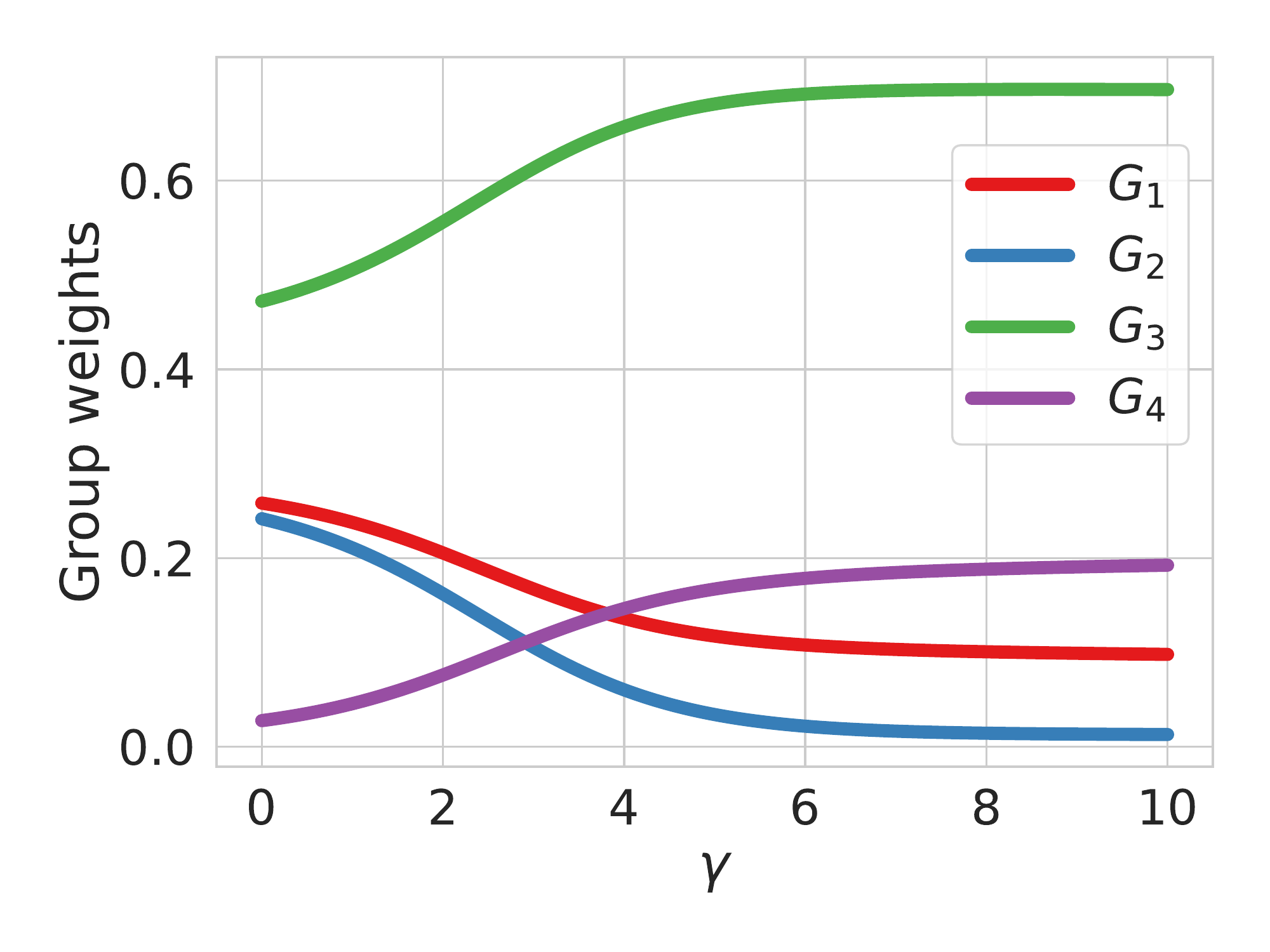}
    \label{fig:celeba_group_weights}
    }
    \subfloat[$\gamma=0$]{\includegraphics[width=0.5\columnwidth]{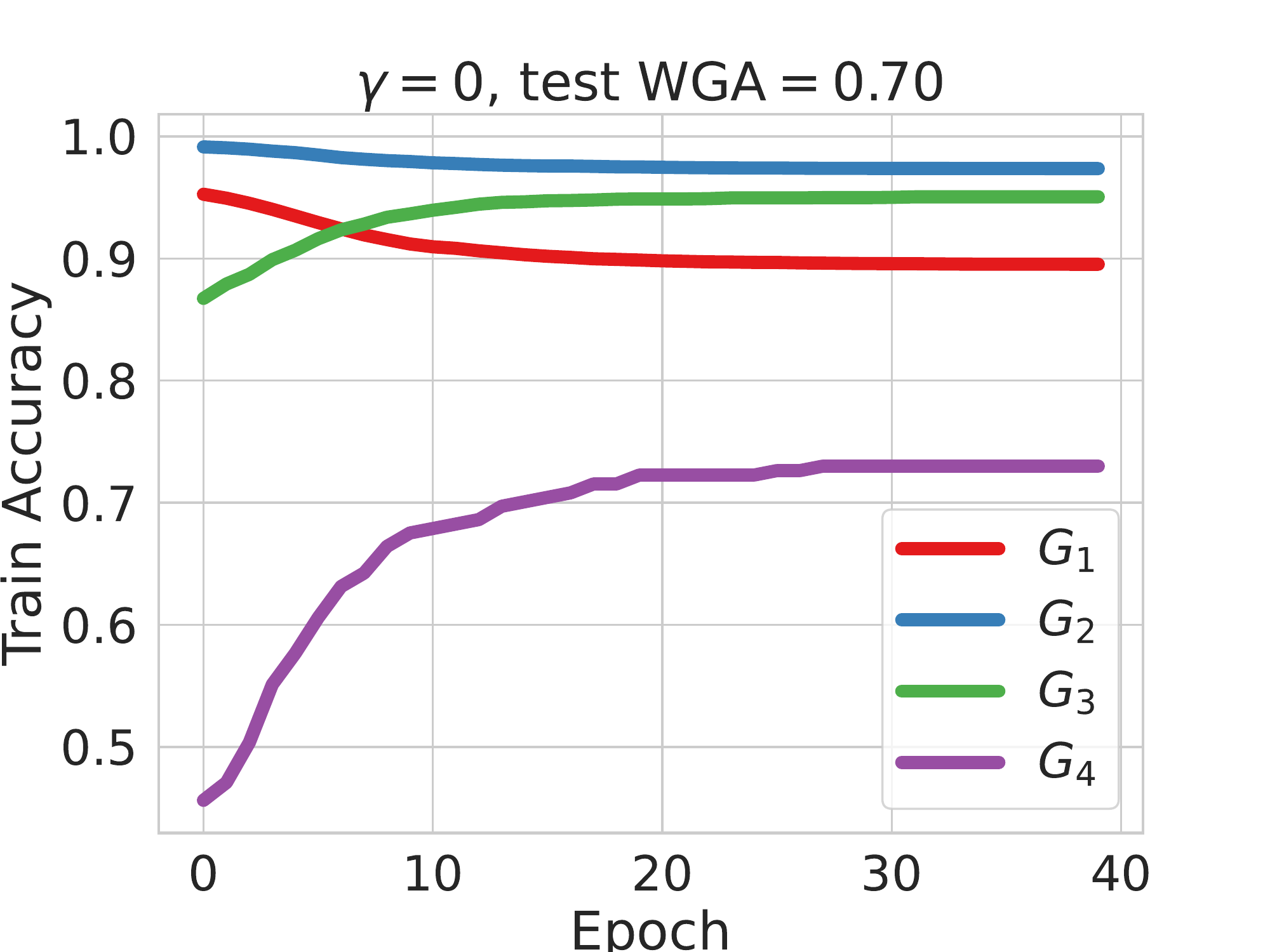}
    \label{fig:celeba_train_group_acc_gamma=0}
    }
    \subfloat[$\gamma=4$]{
    \includegraphics[width=0.5\columnwidth]{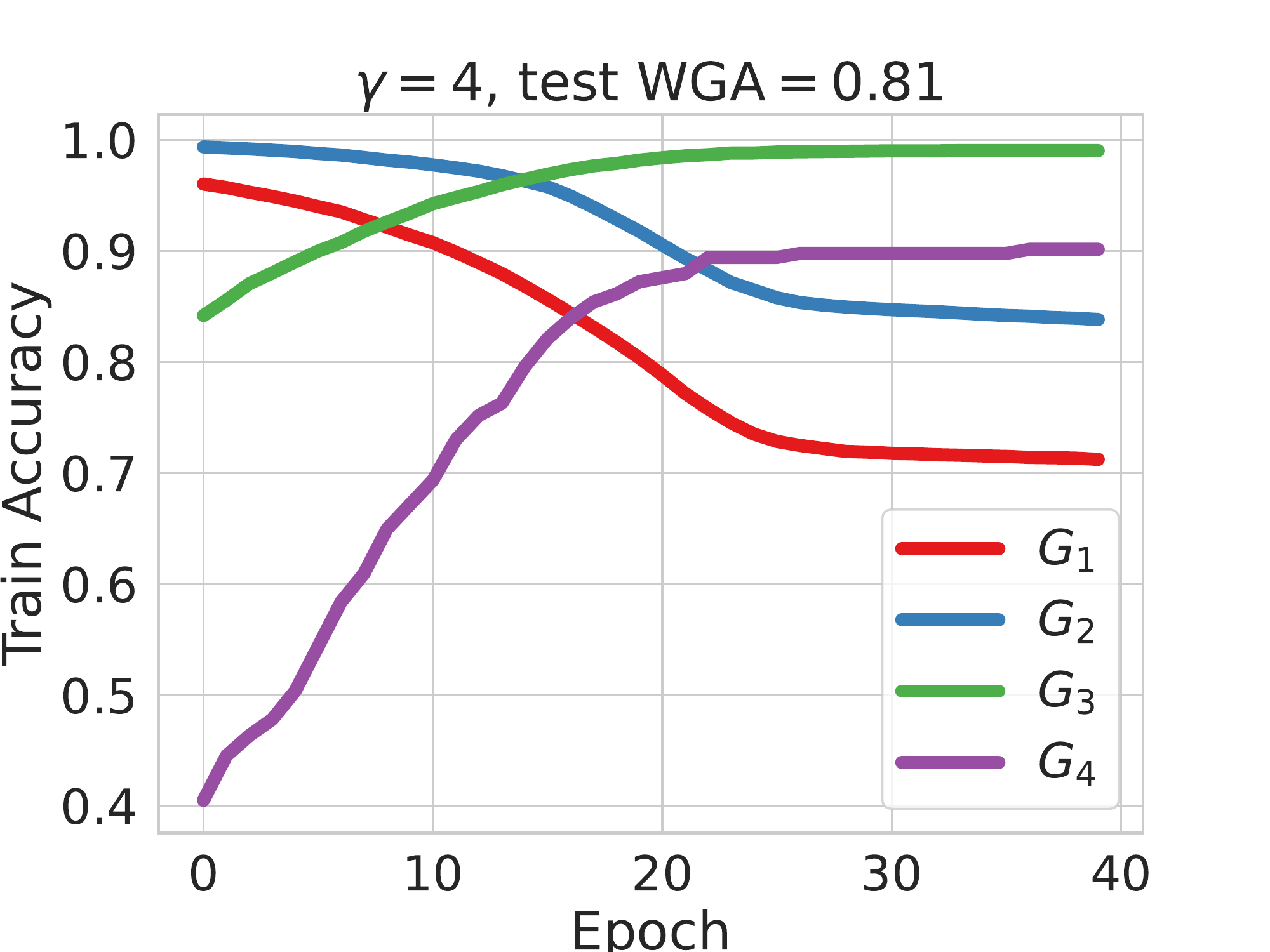}
    \label{fig:celeba_train_group_acc_gamma=4}
    }
    \subfloat[Group-balanced]{\includegraphics[width=0.5\columnwidth]{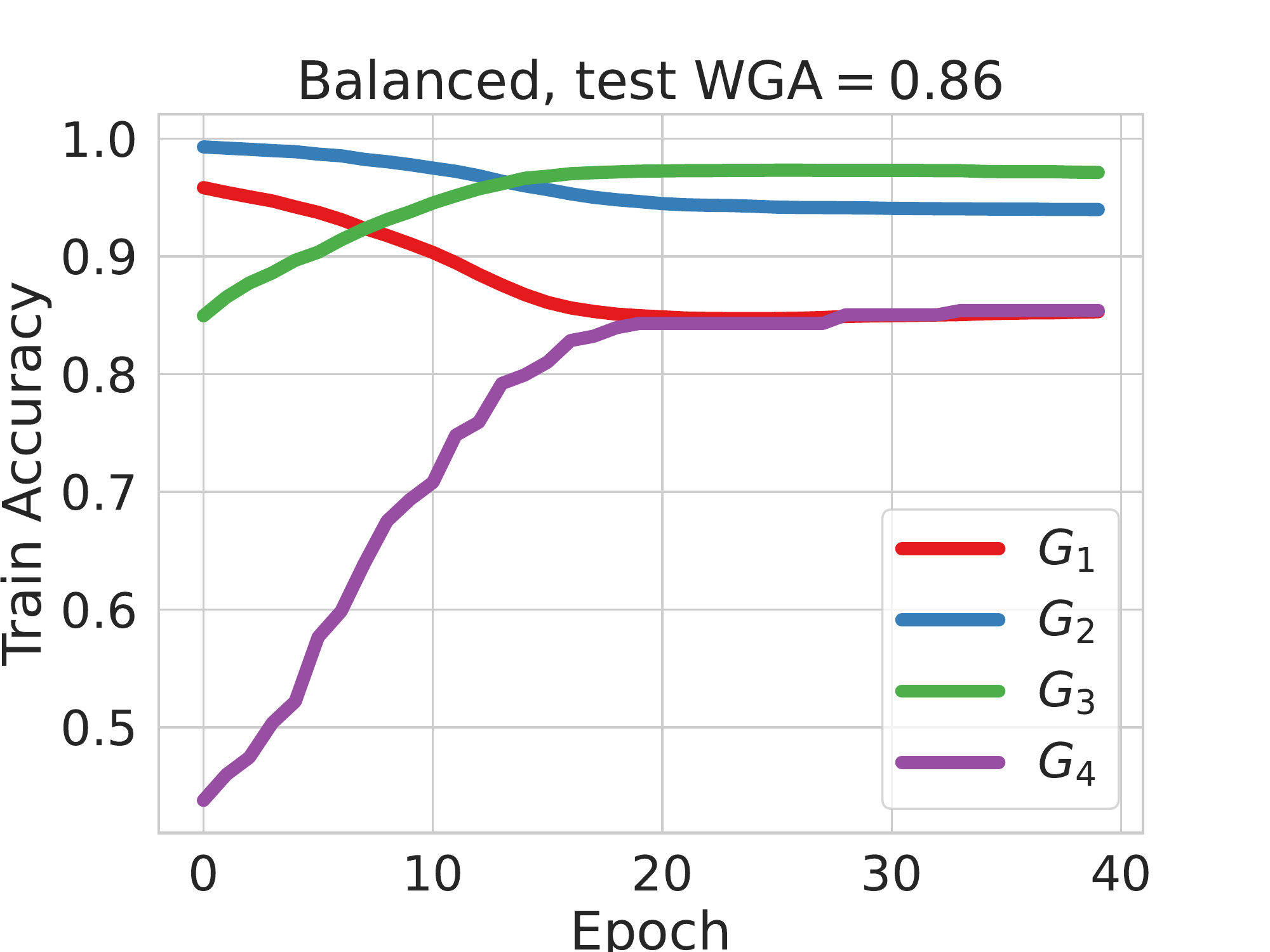}
    \label{fig:celeba_train_group_acc_gamma=balanced}
    }
    \caption{
    \textbf{\method on CelebA.}
    \textbf{(a)} \method\ upweights the minority groups but does not achieve group-balance.
    \textbf{(b)} $\gamma = 0:$ \method\ weight on the worst group $G_4$ is not high enough.
    \textbf{(c)} $\gamma = 4:$ Accuracy increases for $G_3$ and $G_4$ but decreases for $G_1$ and $G_2$ due to group-imbalance in \method\ weights.
    \textbf{(d)} Group-balanced weights improves training dynamics and performance.}
    \label{fig:celeba_analysis}
    \vspace{-4mm}
\end{figure*}

\section{Are AFR Weights Optimal?}
\label{sec:optimal}
In this section, we explore how optimal the AFR weights are for improving group robustness.
We find that it is not always possible for AFR to produce group-balanced weights and we could achieve better performance if we had access to such weights.
However, we also find that AFR's weights are in fact \textit{near-optimal} for maximally re-balancing the group distribution, given constraints on the available information and practical considerations. These findings suggest both that AFR should be very effective among approaches based on dataset reweighting, and that these approaches, including AFR and \jtt, share intrinsic limitations,
and alternative approaches, e.g. \cnc, may provide better results on certain datasets, such as CelebA.

\subsection{AFR Weights Are Imbalanced on CelebA}
\citet{kirichenko2022dfr} showed that retraining the last layer of an ERM checkpoint with a group-balanced held-out set is sufficient for state-of-the-art worst group accuracy. By upweighting minority group examples, \method\ aims to automatically construct a group-balanced held-out set without access to group labels.
However, achieving completely balanced group weights with \method is not always possible.

To illustrate this point, we analyze how \method\ weights distribute over different groups, and study their training dynamics, as a function of $\gamma$ on CelebA, where \method does not match state of the art results.
In Figure~\ref{fig:celeba_analysis}, we show the group aggregated weights produced by \method\ on CelebA and the training accuracy per group during the second stage.
While \method\ correctly upweights the two groups $G_4$ and $G_3$ with the lowest accuracy and downweights the remaining groups $G_1$ and $G_2,$ at no value of $\gamma$ are the group aggregated weights close to balanced, unlike on Waterbirds (see Figure \ref{fig:wb_group_weights}).
Either the worst group $G_4$ is assigned too small a weight that it is not sufficiently optimized for by the model as in Figure~\ref{fig:celeba_analysis}(b), or the second worst group $G_3$ is assigned too large a weight that the accuracy on one of the other groups ($G_1$) drops significantly as seen in Figure~\ref{fig:celeba_analysis}(c).
The reason \method\ assigns a much larger weight to $G_3$ than to $G_4$ is because $G_3$ is $14$ times larger in population than $G_4,$ despite being the second worst group.

To demonstrate that the imbalance of the weights is the cause for AFR's sub-optimal performance, we show in Figure~\ref{fig:celeba_analysis}(d)
that last layer retraining on the reweighting set can achieve as high as $86\%$ train and test WGA if we had access to group-balanced weights where $\mu_i \propto 1 / |G_i|$ with $|G_i|$ denoting the size of group $i.$

\subsection{AFR Weights are Near-Optimal Given Constraints}
AFR weights are not optimal in the unrestricted sense, since group-balanced weights perform better on CelebA.
But are AFR weights optimal without using additional information?

In Appendix~\ref{sec:impossible}, we present strong evidence showing that, rather than AFR's specific choice of $\mu_i \propto \beta_{y_i} \exp(-\gamma \, \hat{p}_i)$ being ineffective, there in fact exists no function of only the first-stage predictions (which contains $\hat{p}_i$) and the true class label $y_i$ that can produce completely balanced weights on CelebA.
We do this by showing a neural network trained to maximally balance group weights is unable to do so given the first-stage predictions and class labels.
We provide training details in Appendix~\ref{sec:impossible}.
This result indicates that, without access to additional information, any approach based on reweighting the dataset, such as AFR and \jtt, is unlikely to lead to optimal performance on CelebA, potentially explaining the fact that both AFR and \jtt are outperformed by \cnc on this dataset by a wide margin.

\begin{figure}[!h]
    \centering
    \subfloat[Waterbirds]{
    \includegraphics[width=0.46\columnwidth]{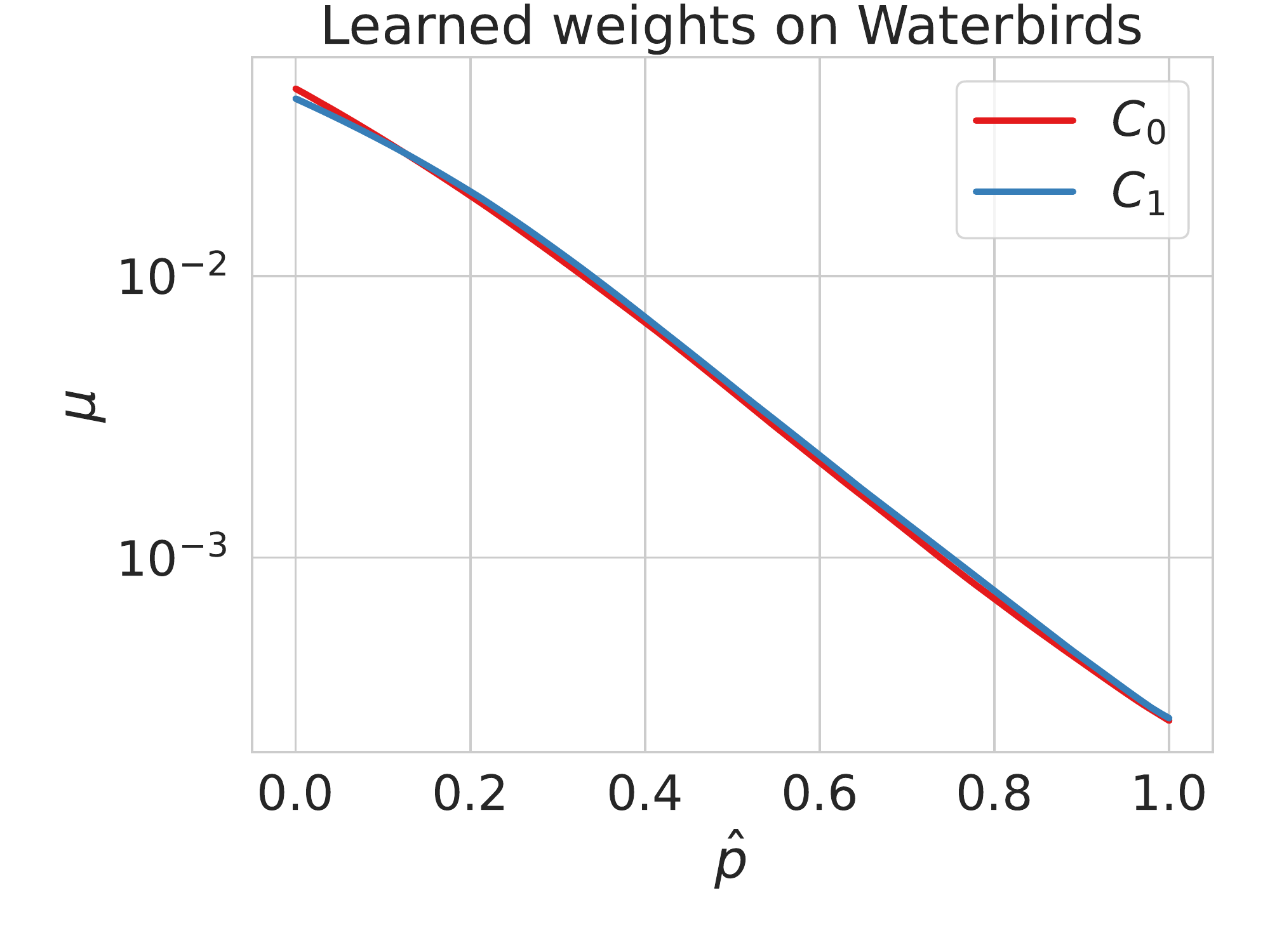}
    }
    \subfloat[CelebA]{
    \includegraphics[width=0.46\columnwidth]{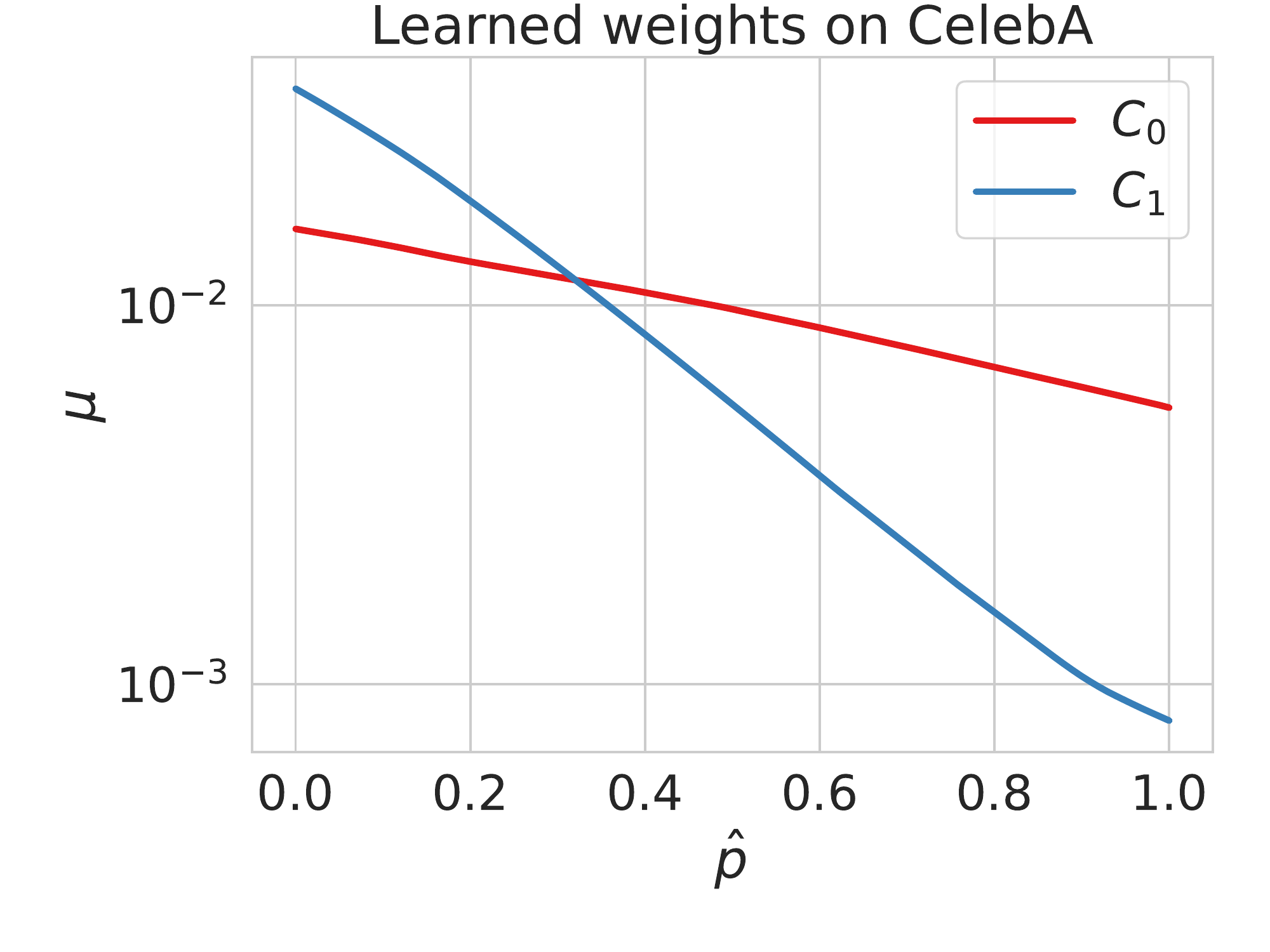}
    }
    \caption{
    The optimal weighting function approximated by a neural network resembles the one used by AFR: a decaying exponential in the predicted probability $\hat{p}$ for the correct class under the first ERM checkpoint, multiplied by class-dependent constants to address class imbalance.
    }
    \label{fig:learned_weights}
    \vspace{-4mm}
\end{figure}

In Figure~\ref{fig:learned_weights}, we plot the learned weighting function against $\hat{p}$ for the each class on Waterbirds and CelebA.
Using a log-scale for the $y$-axis, all learned functions follow an approximately straight line, showing that the optimal weighting function is indeed well-modelled by a decaying exponential in $\hat{p}$.
Furthermore, the class with fewer examples ($C_1$) is offset by a positive constant in log-scale from the other class ($C_0$) in CelebA, showing that it is useful to allow a class-dependent multiplicative constant before the exponential for addressing class imbalance.

Since neural networks can learn functions much more complex than what can be represented with a single parameter $\gamma$, unsurprisingly there are still some differences between the learned functional form and ours: in log-space the weights aren’t perfectly linear as a function of $\hat{p}$ and on CelebA the slope, which corresponds to  $-\gamma$ in AFR, is class-dependent, which can be accounted for at the cost of more hyperparameters. Given the constraints that 1) we can only afford a handful of hyperparameters so that cross-validation with a small group-annotated validation set is effective and efficient, and 2) no additional information, such as group labels, is used to define the weights, AFR's specification of the weights, $\mu_i \propto \beta_{y_i} \exp(-\gamma \, \hat{p}_i),$ is in fact near-optimal in maximally re-balancing the group distribution.

\vspace{-5mm}

\section{Discussion}

We have seen that we can profoundly simplify approaches to addressing spurious correlations while achieving state-of-the-art performance.
AFR is driven by the key insight that, without either group labels or special intervention, a standard ERM model can be used to infer minority examples and then be minimally modified for improved group robustness.
As a result, \method is considerably simpler and faster than alternatives such as \jtt and \cnc, and more broadly applicable than methods like Group DRO and DFR which use group annotations during training.
\method reduces the barrier to deploying group robustness methods by reducing computational overhead, the need to recognize and label the spurious attributes a priori,
and the need to use non-standard training procedures.
Moreover, we have shown that \method is robust to hyperparameters and can outperform methods like DFR that use group labels for training the model, when only a small number of group labels are available.

On the other hand, \method is not without limitations. Similar to methods like \jtt and \cnc, \method can still benefit from some group-labeled data for hyperparameter tuning. As the number of groups increases, more group labels will be needed to estimate worst group performance and select the best hyperparameters, which may pose challenges for its use in datasets with numerous classes and spurious attributes. Indeed, \method cannot always re-balance the groups, though it achieves near-optimal balance without access to group label information. While using group-balanced weights is not necessarily best for maximizing worst group accuracy, we have seen improving the balance in \method's weights can further improve worst group performance.
Addressing these limitations offers an exciting opportunity for future research.

\section*{Acknowledgements}
We thank Greg Benton and Yucen Lily Li for helpful discussions.
This work is supported by NSF CAREER IIS-2145492,
NSF I-DISRE 193471, NIH R01DA048764-01A1, NSF IIS-1910266, NSF 1922658 NRT-HDR,
Meta Core Data Science, Google AI Research, BigHat Biosciences, Capital One, and an
Amazon Research Award.

\clearpage
\bibliography{references}
\bibliographystyle{icml2023}

\newpage
\appendix
\onecolumn

\section*{Appendix Outline}

This Appendix is organized as follows:
\begin{itemize}
    \item In Section \ref{sec:code}, we provide code for computing the weights in \method.
    \item In Section \ref{sec:app_details}, we describe the details of the experimets.
    \item In Section \ref{sec:app_ablations}, we provide additional ablations.
    \item In Section \ref{Related Work}, we discuss additional related work.
\end{itemize}

\section{Automatic Feature Reweighting Implementation Details}
\label{sec:code}

In this section, we provide the code for computing the weights for the weighted loss in \method.

\begin{python}
def compute_afr_weights(erm_logits, class_label, gamma):
  # erm_logits: (n_samples, n_classes)
  # class_label: (n_samples,)
  # gamma: float
  with torch.no_grad():
    p = erm_logits.softmax(-1)
    y_onehot = torch.zeros_like(erm_logits).scatter_(-1, class_label.unsqueeze(-1), 1)
    p_true = (p * y_onehot).sum(-1)
    weights = (-gamma * p_true).exp()
    n_classes = torch.unique(class_label).numel()
    # class balancing
    class_count = []
    for y in range(n_classes):
      class_count.append((class_label == y).sum())
    for y in range(1, n_classes):
      weights[class_label == y] *= class_count[0] / class_count[y]
    weights /= weights.sum()
  return weights
\end{python}

\section{Experimental details}
\label{sec:app_details}

\subsection{Last layer retraining implementation}
To retrain the last layer, we pre-compute and store the embeddings produced by the stage 1 ERM model. Unless stated otherwise, we use full-batch gradient descent without momentum to retrain the last layer parameters, initializing them to their values at the end of stage 1. We clip the $\ell_2$ norm of the gradient vector to 1. The epoch achieving highest validation worst group accuracy is used to report our result, except for the experiment in Section~\ref{sec:efficient} where we do not perform early stopping on Waterbirds and CelebA when not using the entire validation set, which we found to improve performance by avoiding stopping too early.

\subsection{Timing experiment details}
\label{sec:times_details}
Timings for AFR and \jtt in Figure \ref{fig:time} are obtained by running on a single
RTX8000 (48 GB) NVIDIA GPU.
For JTT we have the following
(epochs, batch size) tuples per dataset which incorporate the two stages of training.
Waterbirds (360, 64), CelebA (51, 128), MultiNLI (7, 32),
CivilComments (7, 16), and CXR (150, 128).
For \cnc, we present timings reported by \citet{zhang2022cnc}.

\subsection{Dataset details}
We now detail the models and hyperparameters used on each of the tasks.

\begin{itemize}
    \item \textbf{Waterbirds:} ResNet-50 (torchvision.models.resnet50(pretrained=True))
    \begin{itemize}
     \item \textbf{\method} 1st stage: epochs = 50,
     optimizer=sgd, scheduler=cosine, batch size = 32,
     learning rate = 3e-3, weight decay = 1e-4.
     \item \textbf{\method} 2nd stage: epochs = 500, $\gamma$ from 33 points linearly spaced between $[4, 20]$, learning rate in = 1e-2, $\lambda \in \{0, 0.1, 0.2, 0.3, 0.4\}.$
     \item \textbf{\jtt} same hyperparameters as in \citet{liu2021jtt}.
     \item \textbf{\cnc} same hyperparameters as in \citet{zhang2022cnc}.
    \end{itemize}

    \item \textbf{CelebA} ResNet-50 (torchvision.models.resnet50(pretrained=True))
    \begin{itemize}
     \item \textbf{\method} 1st stage: epochs = 20,
     optimizer=sgd, scheduler=cosine, batch size = 32,
     learning rate = 3e-3, weight decay = 1e-4.
     \item \textbf{\method} 2nd stage: epochs = 1000, $\gamma$ from 10 points linearly spaced between $[1, 3]$, learning rate  =2e-2, $\lambda \in \{0.001, 0.01, 0.1\}.$
     \item \textbf{\jtt} same hyperparameters as in \citet{liu2021jtt}.
     \item \textbf{\cnc} same hyperparameters as in \citet{zhang2022cnc}.
    \end{itemize}

    \item \textbf{MultiNLI} BERT (using HuggingFace)
    \begin{itemize}
     \item \textbf{\method} 1st stage: epochs = 5,
     optimizer = SGD, scheduler = constant, batch size = 32,
     learning rate = 2e-5, weight decay = 0.
     \item \textbf{\method} 2nd stage:  epochs = 200, $\gamma$ from 10 points linearly spaced between $[1e2, 1e5]$, learning rate  = 1e-2, $\lambda$ from 26 points linearly spaced between $[0, 50].$
     \item \textbf{\jtt} same hyperparameters as in \citet{liu2021jtt}.
     \item \textbf{\cnc} same hyperparameters as in \citet{zhang2022cnc}.
    \end{itemize}

    \item \textbf{CivilComments} BERT (using HuggingFace)
    \begin{itemize}
     \item \textbf{\method} 1st stage: epochs = 3,
     optimizer = SGD, scheduler = constant, batch size = 24,
     learning rate = 2e-5, weight decay = 0.
     \item \textbf{\method} 2nd stage: $\gamma \in \{0, 0.01, 0.1, 1, 3, 10\}$,
     $\lambda=0$, learning rate $3 \cdot 10^{-3}$.
      For hyperparameter tuning and early stopping, we combine all religions into a single attribute, as we found it to produce more stable results.
     \item \textbf{\jtt} same hyperparameters as in \citet{liu2021jtt}.
     \item \textbf{\cnc} same hyperparameters as in \citet{zhang2022cnc}.
    \end{itemize}

    \item \textbf{Chest X-Ray} DenseNet121 (torchvision.models.densenet121(pretrained=True))
    \begin{itemize}
     \item \textbf{\method} 1st stage: epochs = 100,
     optimizer = AdamW, scheduler = constant, batch size = 32,
     learning rate = 1e-4, weight decay = 1e-5.
     \item \textbf{\method} 2nd stage: $\gamma \in [1, 6]$, $\lambda=0$, learning rate $\in \{10^{-4}, 50^{-4}, 10^{-3}, 50^{-3}, 10^{-2}, 50^{-2} \}$.
     \item \textbf{\jtt} same hyperparameters as in \citet{yang2022chromavae}.
    \end{itemize}

    \item \textbf{CIFAR-10} ResNet-18 (torchvision.models.resnet18(pretrained=True))
    \begin{itemize}
    \item \citep{krizhevsky2009cifar10},
    is an image classification task with no obvious spurious features.
    we want to identify
    to what of 10 groups (airplane, automobile, bird, cat, deer, dog, frog, horse, ship and truck) does an
    image belong to.
     \item \textbf{\method} 1st stage / \textbf{ERM}: epochs = 20,
     optimizer = AdamW, scheduler = cosine, batch size = 32,
     learning rate = 1e-4, weight decay = 5e-4.
    \item \textbf{\method} 2nd stage: epochs = 500, optimizer = sgd, scheduler = constant, batch size = full batch,
     $\gamma \in \{1, 2, 5, 10\}$, learning rate in
     $\{1e-4, 1e-3, 1e-2\}$,
     weight decay = 0.
     $\lambda = 0$.
     \end{itemize}

     \item \textbf{Camelyon17} DenseNet-121 (torchvision.models.densenet121(pretrained=True))
    \begin{itemize}
    \item \citep{zech2018variable, koh2021wilds}
    is an image classification task where the goal is to predict whether a a digital whole-slide image (WSIs) of
    a lymph node section contains cancer metastases.
     \item \textbf{\method} 1st stage / \textbf{ERM}: epochs = 20,
     optimizer = AdamW, scheduler = constant, batch size = 256,
     learning rate = 1e-3, weight decay = 1e-4.
     When evaluating the final worst group accuracy, we treat all groups separately, as prescribed in \citet{koh2021wilds}.
     \item \textbf{\method} 2nd stage: epochs = 500, optimizer = SGD, scheduler = constant, batch size = full batch,
     $\gamma \in (1, 100)$, learning rate in
     $\{1e-4, 5e-4, 1e-3, 5e-3, 1e-2\}$,
     weight decay = 0.
     $\lambda = 0$.
    \end{itemize}

\end{itemize}

\begin{figure*}
    \centering
    \begin{tabular}{cccc}
    \hspace{-3em}
    \includegraphics[width=0.27\textwidth]{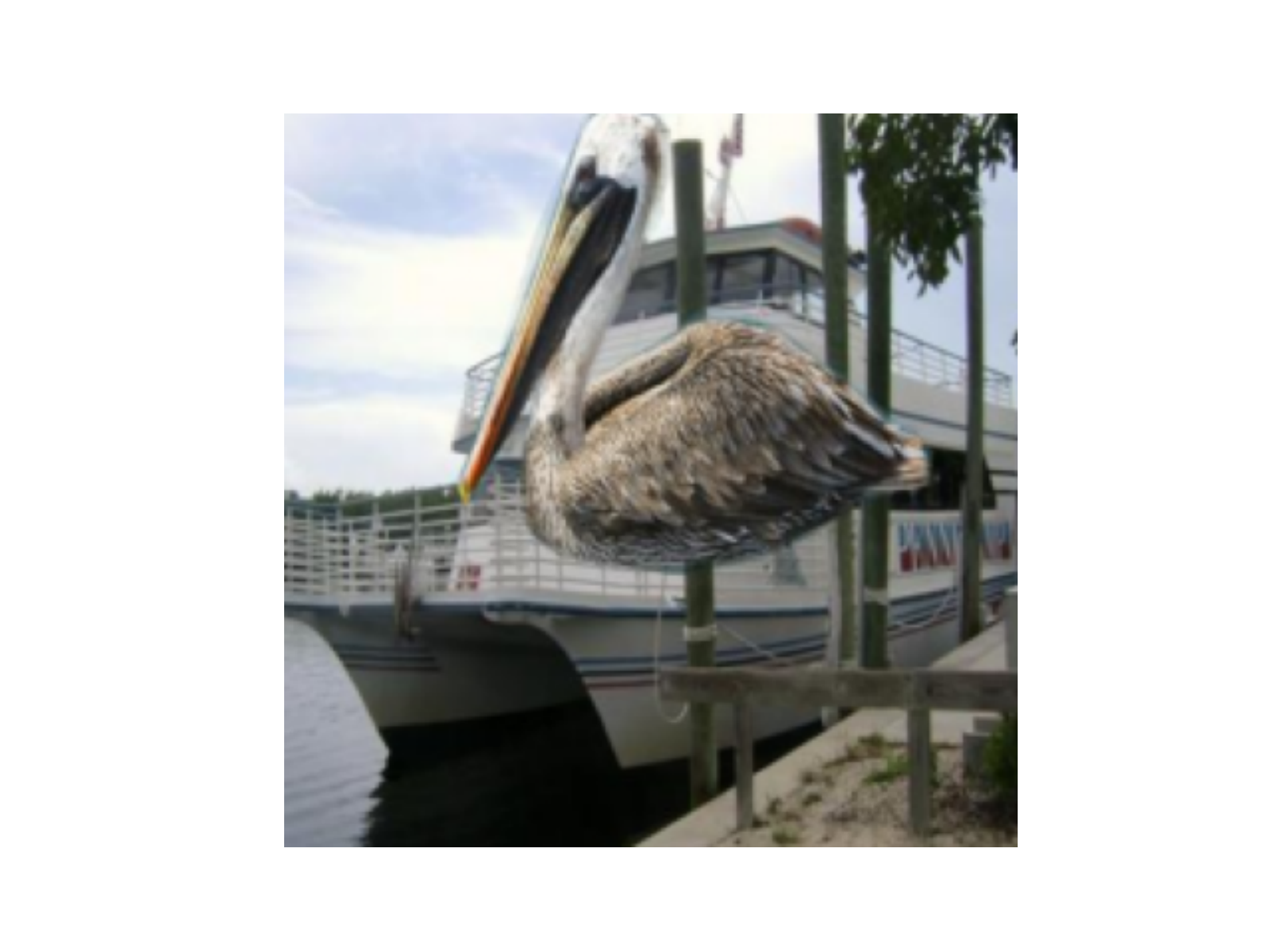}
      &
    \hspace{-3em}
    \includegraphics[width=0.27\textwidth]{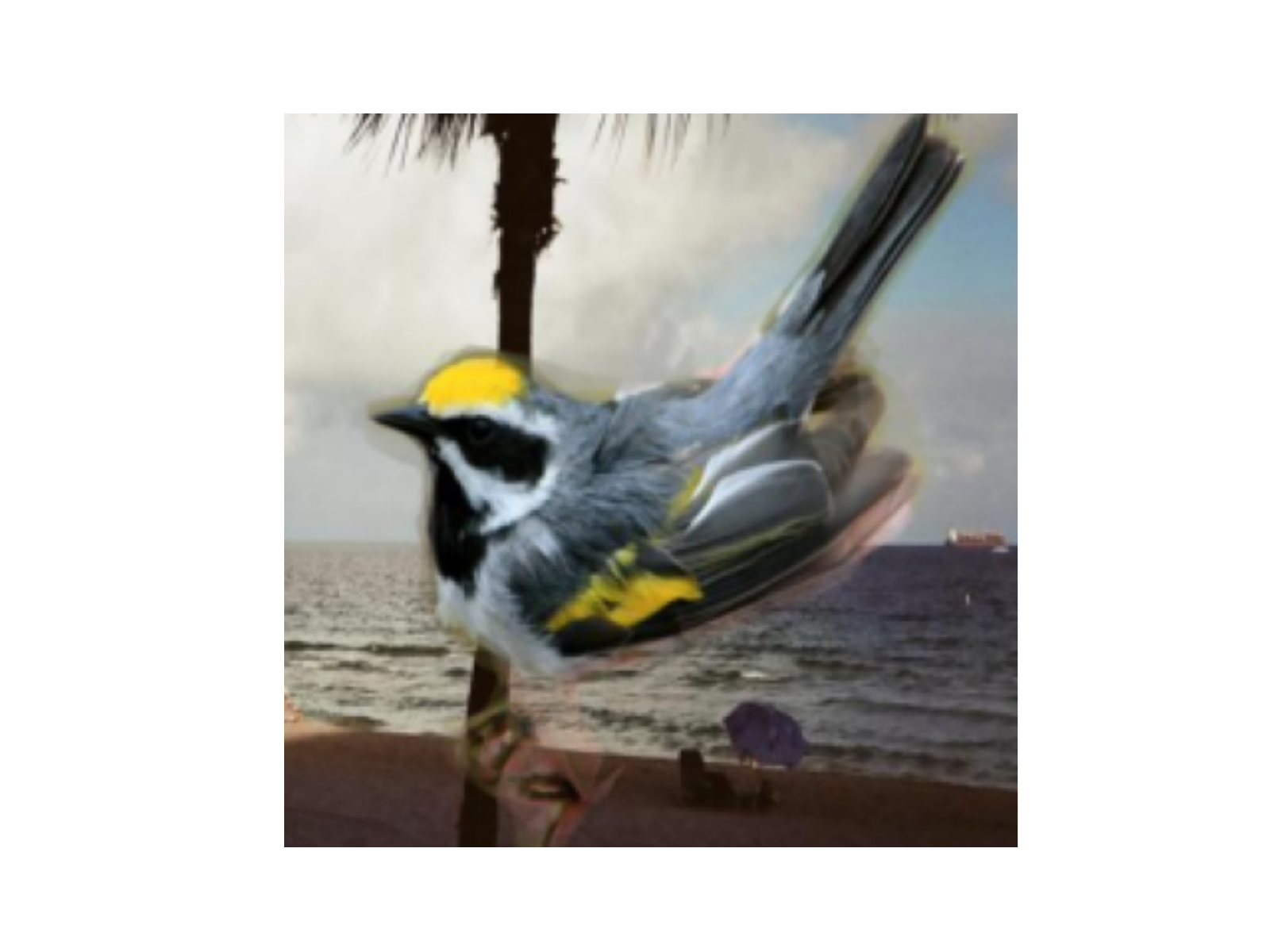}
    &
    \hspace{-3em}
    \includegraphics[width=0.27\textwidth]{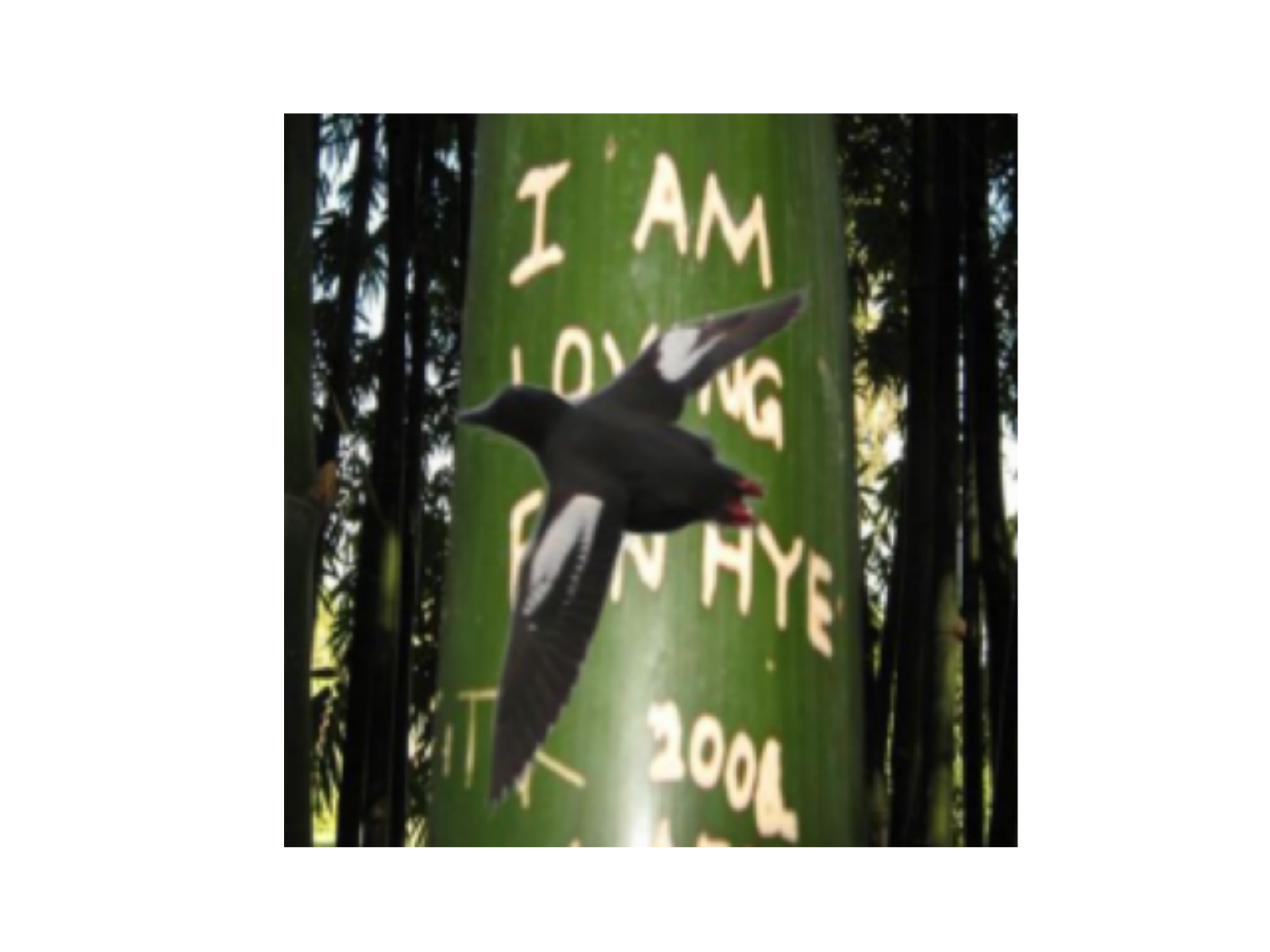}
    &
    \hspace{-3em}
    \includegraphics[width=0.27\textwidth]{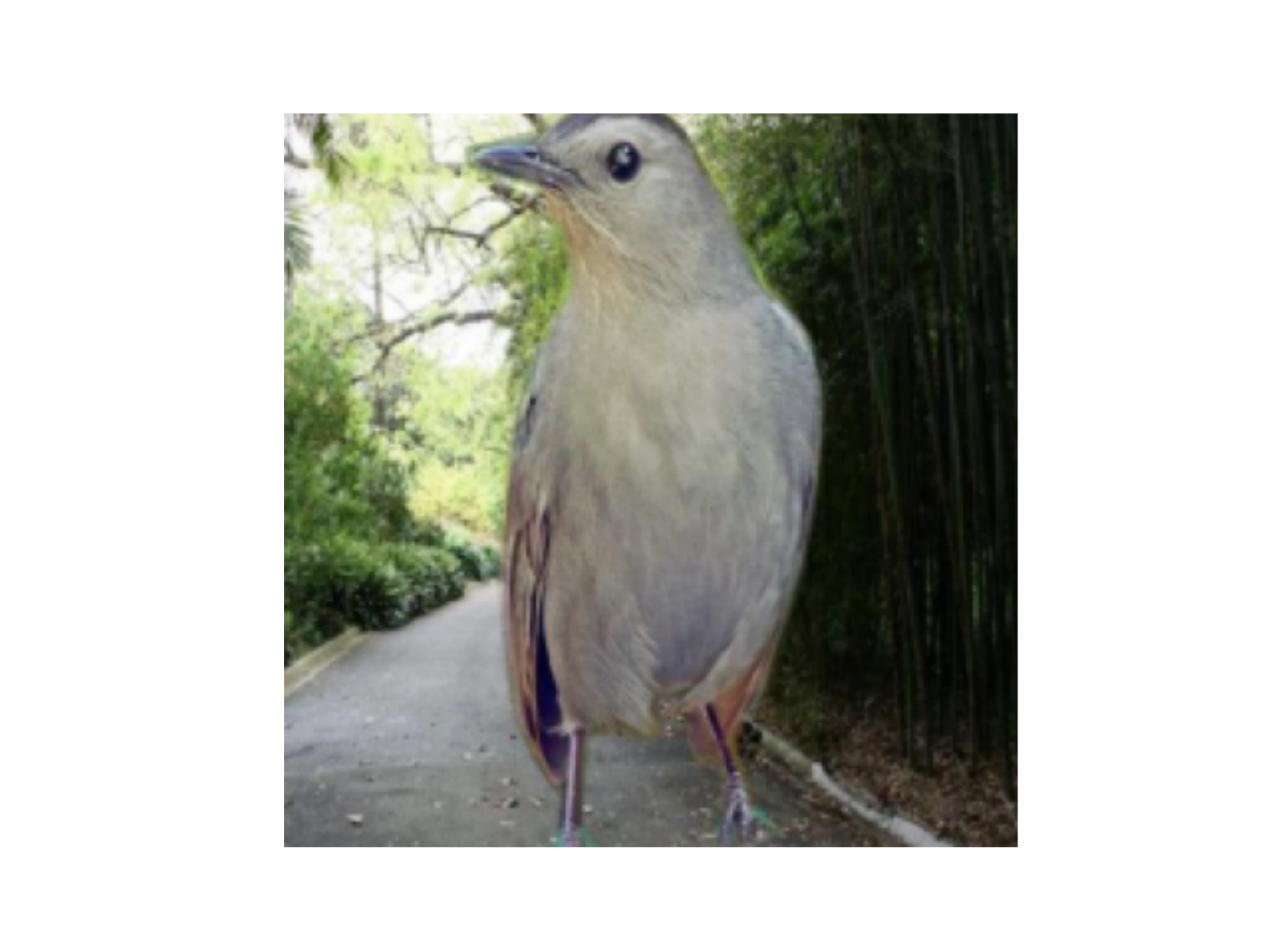}
    \vspace{-1em}
    \\
      \hspace{-3em}
      (a) Waterbird on water &
      \hspace{-3em}
      (b) Landbird on water &
      \hspace{-3em}
      (c) Waterbird on land &
      \hspace{-3em}
      (d) Lanbird on land
    \end{tabular}
    \caption{
      \textbf{Example images on Waterbirds.}
      The objective is to identify the bird type (landbird $y=0$ vs waterbird $y=1$)
      where the spurious feature is the background (water vs land).
      The groups are $\mathcal{G}_{1}=\text{lanbird on land}$,
      $\mathcal{G}_{2}=\text{lanbird on water}$,
      $\mathcal{G}_{3}=\text{waterbird on land}$, and
      $\mathcal{G}_{4}=\text{waterbird on water}$.
      In terms of number of training samples per group we have
      $\mathcal{G}_{1}=3,498 (73\%)$,
      $\mathcal{G}_{2}=184 (4\%)$,
      $\mathcal{G}_{3}=56 (1\%)$, and
      $\mathcal{G}_{4}=1,057 (22\%)$.
    }
    \label{fig:examples_wb}
\end{figure*}

\begin{figure*}
    \centering
    \begin{tabular}{cccc}
    \hspace{-3em}
    \includegraphics[width=0.27\textwidth]{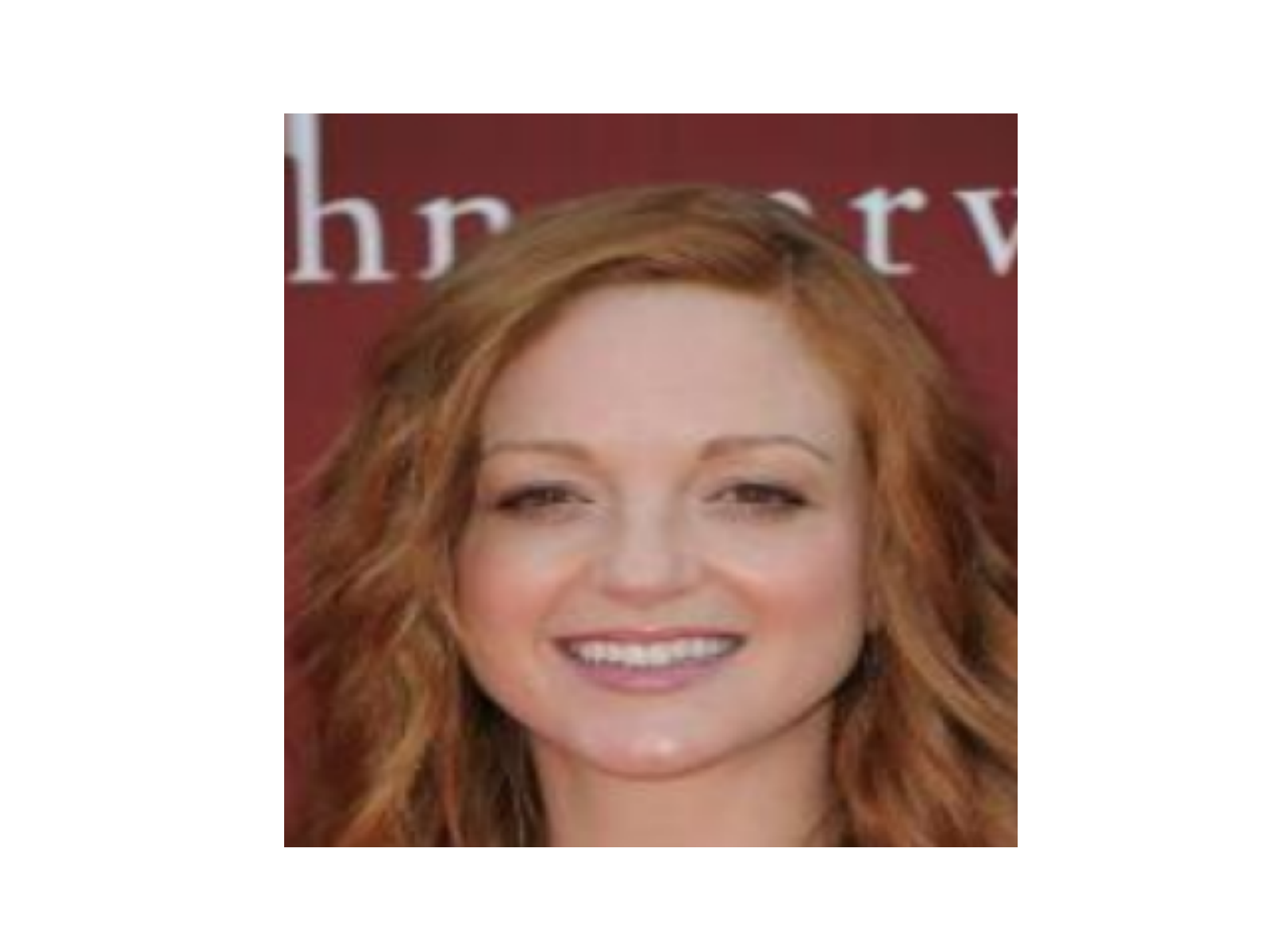}
      &
    \hspace{-3em}
    \includegraphics[width=0.27\textwidth]{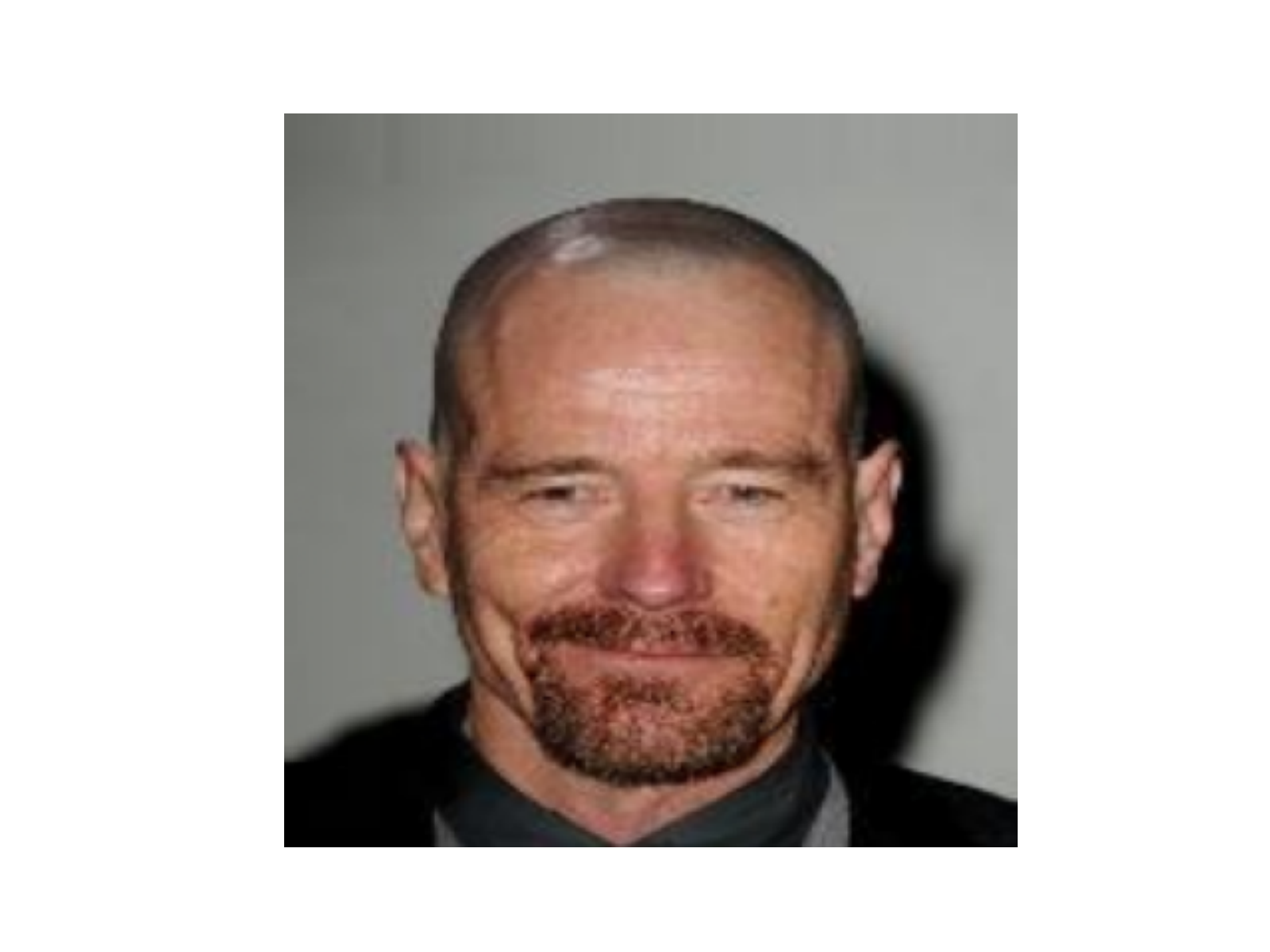}
    &
    \hspace{-3em}
    \includegraphics[width=0.27\textwidth]{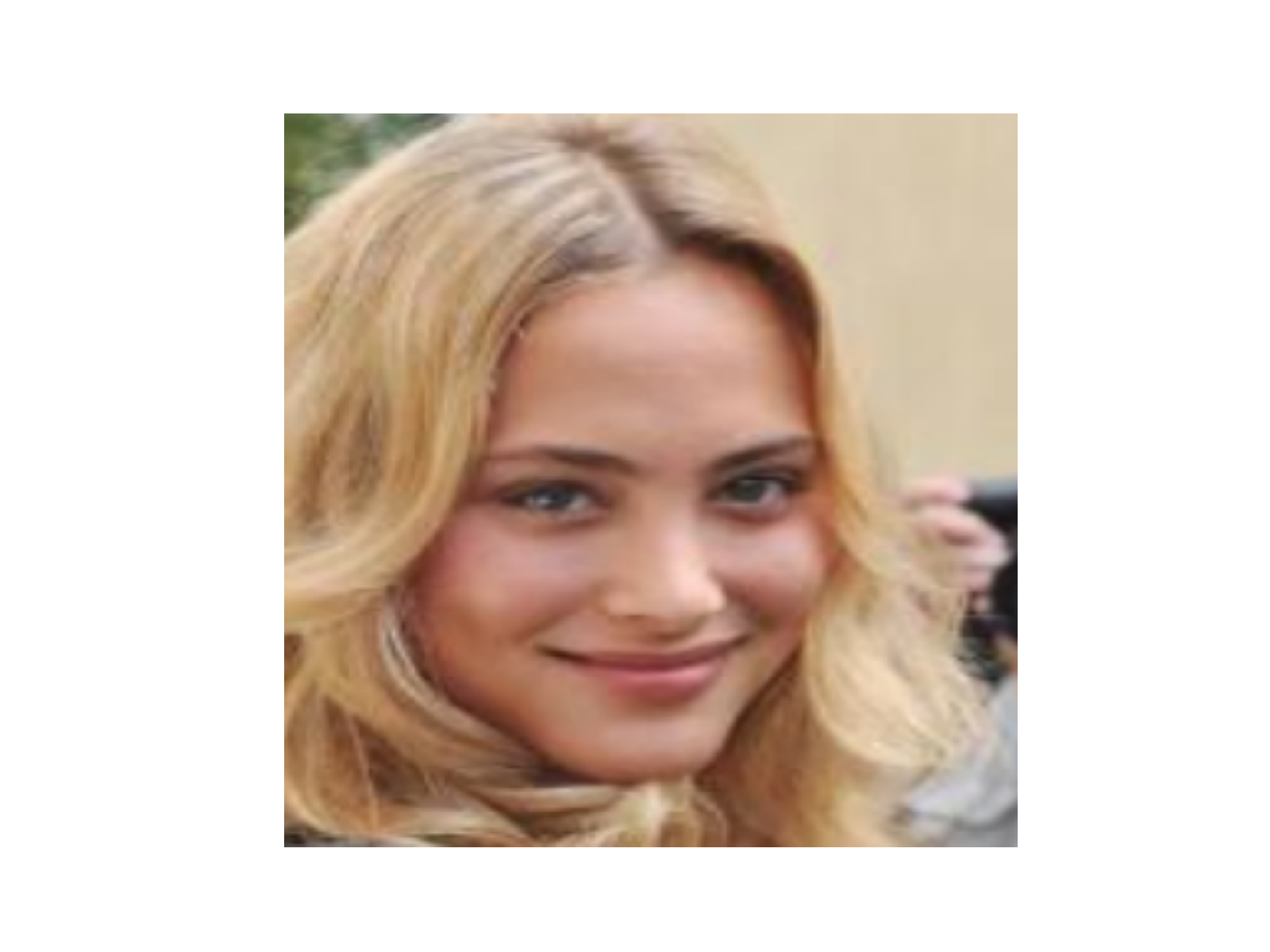}
    &
    \hspace{-3em}
    \includegraphics[width=0.27\textwidth]{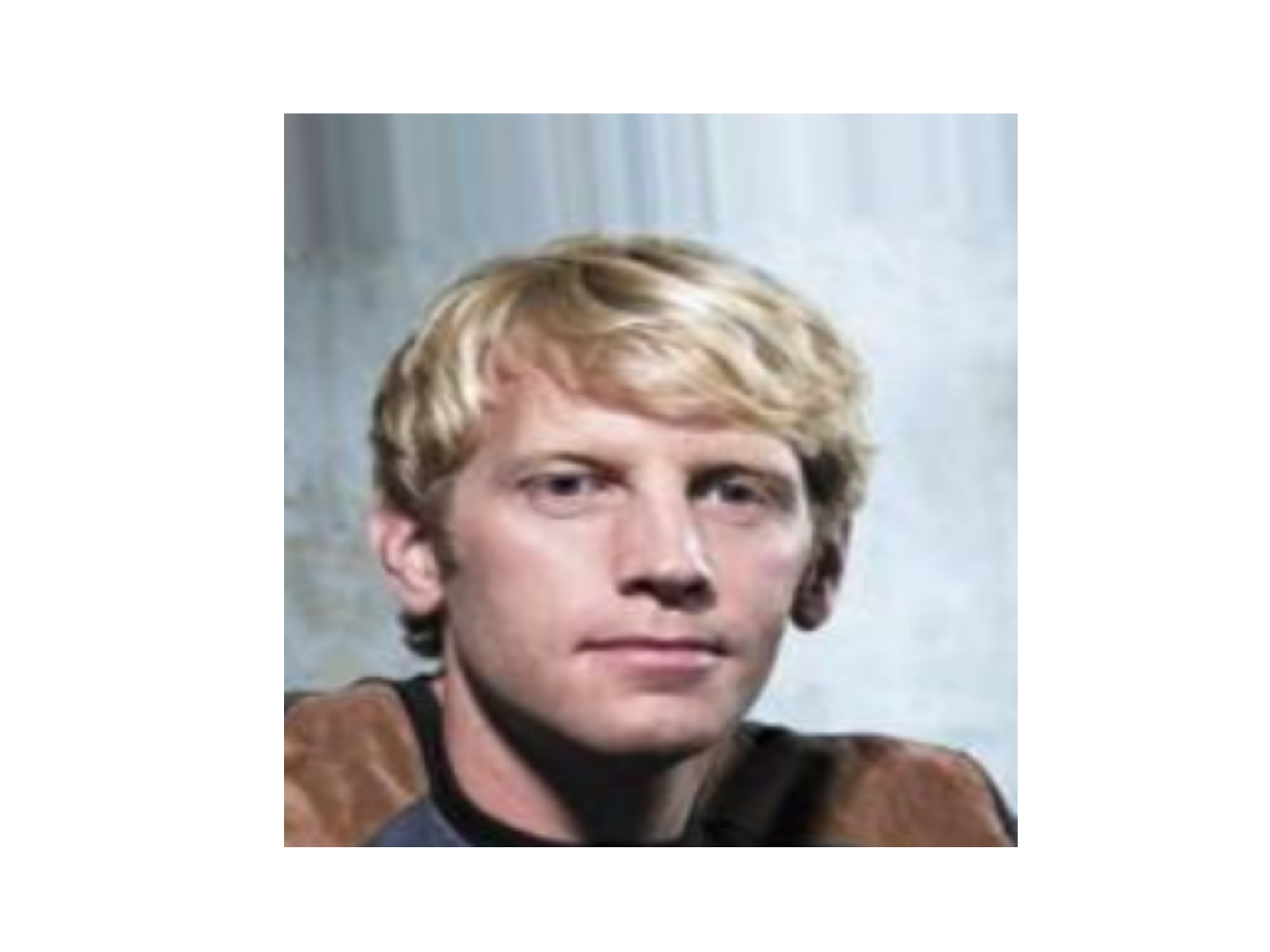}
    \vspace{-1em}
    \\
      \hspace{-3em}
      (a) Non-blond female &
      \hspace{-3em}
      (b) Non-blond male &
      \hspace{-3em}
      (c) Blond female &
      \hspace{-3em}
      (d) Blond male
    \end{tabular}
    \caption{
    \textbf{Example images on CelebA.}
      The objective is to differentiate hair color (non-blond $y=0$ vs blond $y=1$)
      where the spurious feature is the gender (female vs male).
      The groups are $\mathcal{G}_{1}=\text{non-blond female}$,
      $\mathcal{G}_{2}=\text{non-blond male}$,
      $\mathcal{G}_{3}=\text{blond female}$, and
      $\mathcal{G}_{4}=\text{blond male}$.
      In terms of number of training samples per group we have
      $\mathcal{G}_{1}=71,629 (44\%)$,
      $\mathcal{G}_{2}=66,874 (41\%)$,
      $\mathcal{G}_{3}=22,880 (14\%)$, and
      $\mathcal{G}_{4}=1,387 (1\%)$.
    }
    \label{fig:examples_celeba}
\end{figure*}

\begin{figure*}
    \centering
    \begin{tabular}{cccc}
    \hspace{-3em}
    \includegraphics[width=0.27\textwidth]{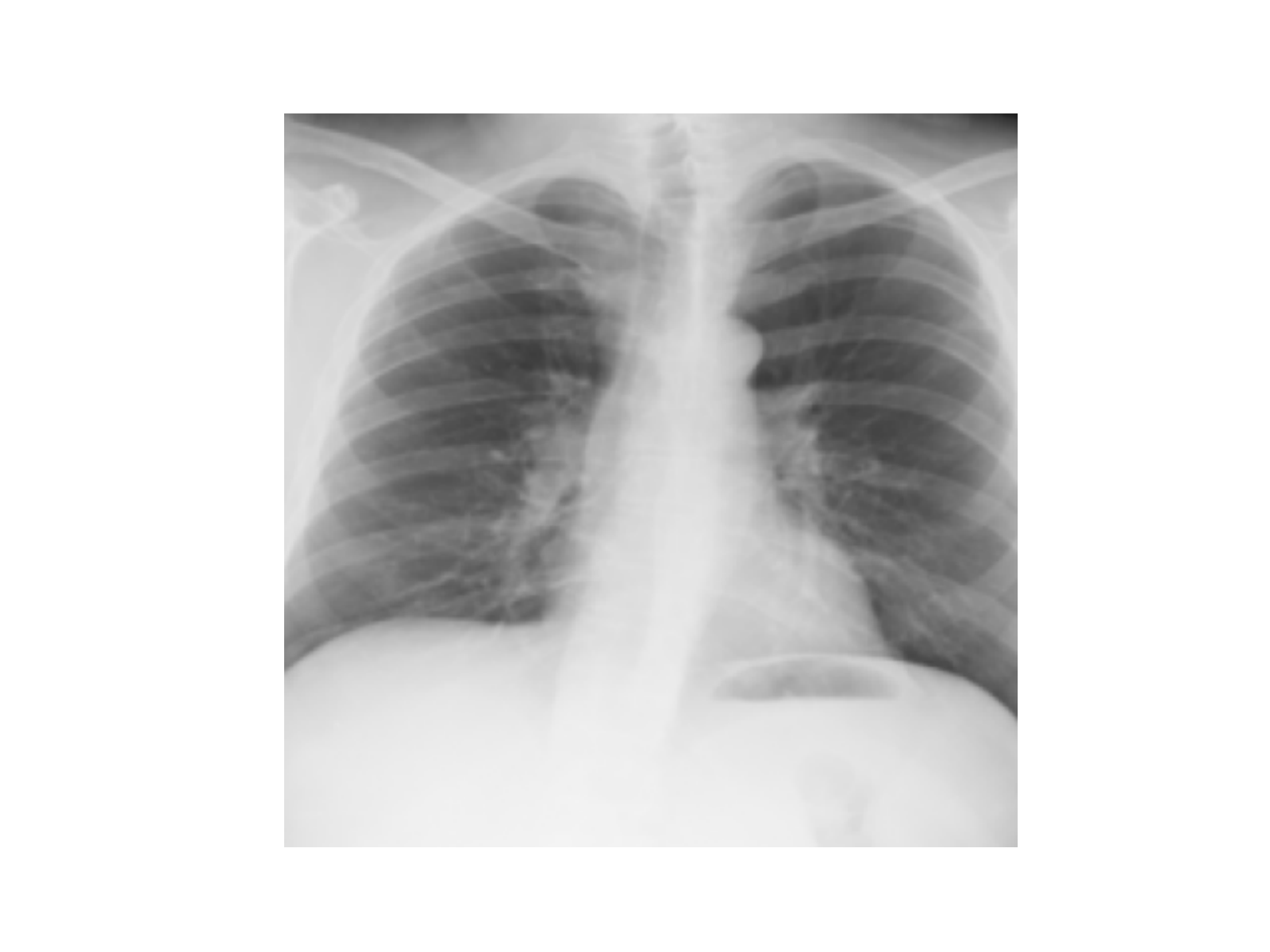}
      &
    \hspace{-3em}
    \includegraphics[width=0.27\textwidth]{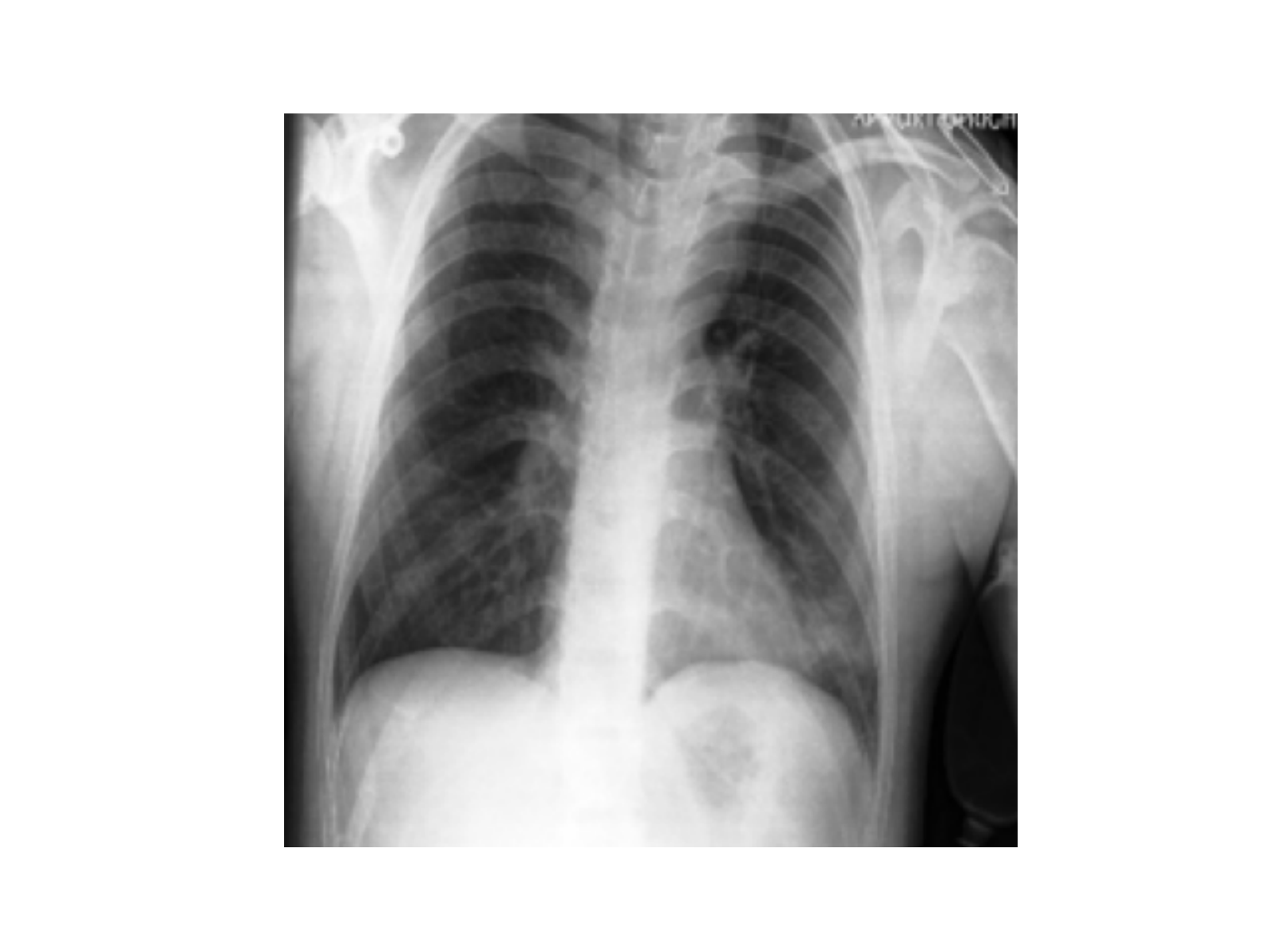}
    &
    \hspace{-3em}
    \includegraphics[width=0.27\textwidth]{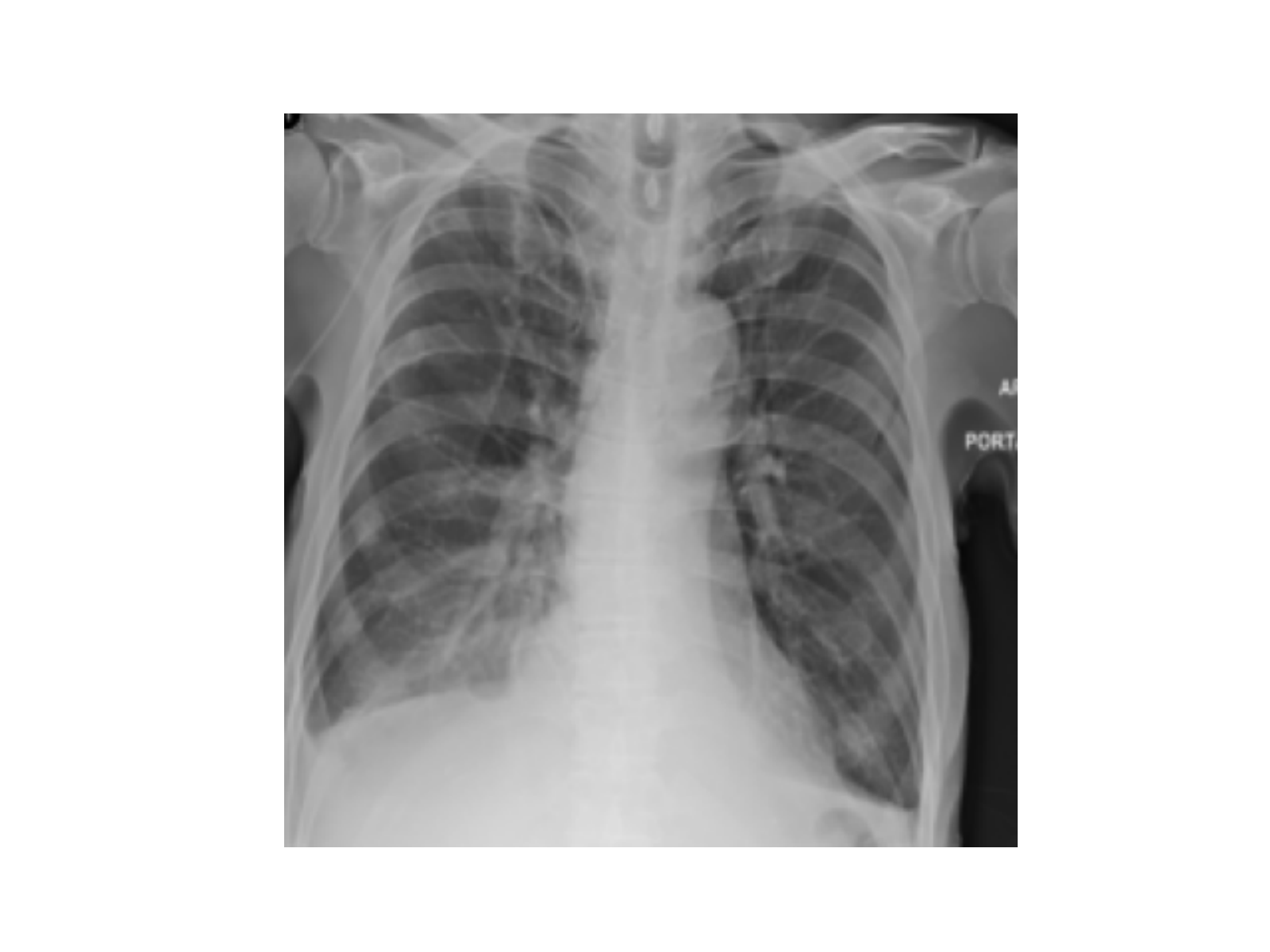}
    &
    \hspace{-3em}
    \includegraphics[width=0.27\textwidth]{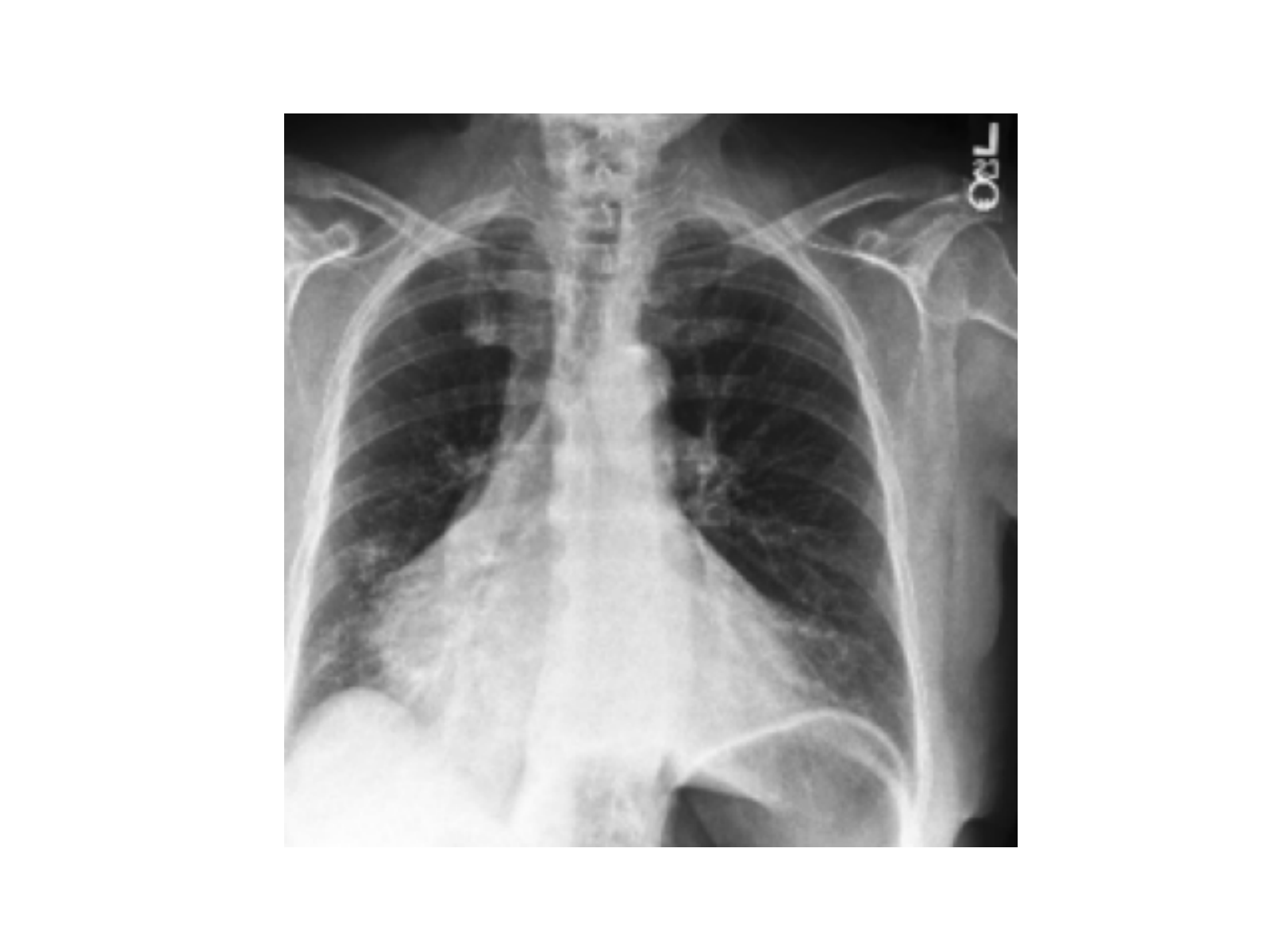}
    \vspace{-1em}
    \\
      \hspace{-3em}
      (a) NIH - no &
      \hspace{-3em}
      (b) CheXpert - no &
      \hspace{-3em}
      (c) NIH - pneumonia &
      \hspace{-3em}
      (d) CheXpert - pneumonia
    \end{tabular}
    \caption{
      \textbf{Example images on Chest-X-Ray}.
      The objective is to identify if the patient has pneumonia (healthy $y=0$ vs pneumonia $y=1$)
      where the spurious features are the hospital markings from
      the different sources (NIH and CheXpert).
      The groups are $\mathcal{G}_{1}=\text{NIH - no}$,
      $\mathcal{G}_{2}=\text{CheXpert - no}$,
      $\mathcal{G}_{3}=\text{NIH - pneumonia}$, and
      $\mathcal{G}_{4}=\text{CheXpert - pneumonia}$.
      In terms of number of training samples per group we have
      $\mathcal{G}_{1}=9,281 (47\%)$,
       $\mathcal{G}_{2}=8,766 (44\%)$,
      $\mathcal{G}_{3}=978 (5\%)$, and
      $\mathcal{G}_{4}=855 (4\%)$.
    }
    \label{fig:examples_cxr}
\end{figure*}

\begin{figure*}
    \centering
    \begin{tabular}{cccc}
    \hspace{-3em}
    \includegraphics[width=0.27\textwidth]{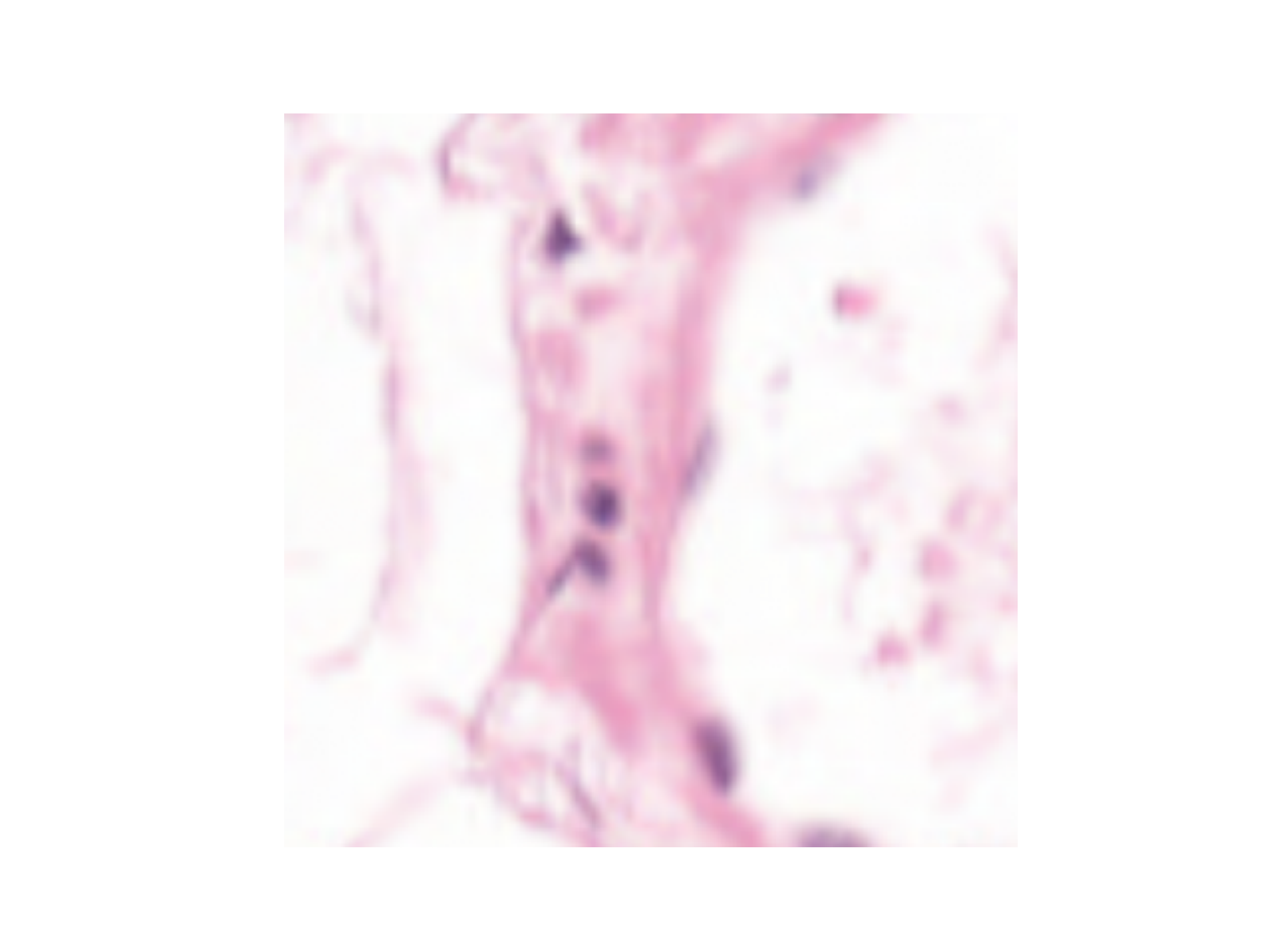}
      &
    \hspace{-3em}
    \includegraphics[width=0.27\textwidth]{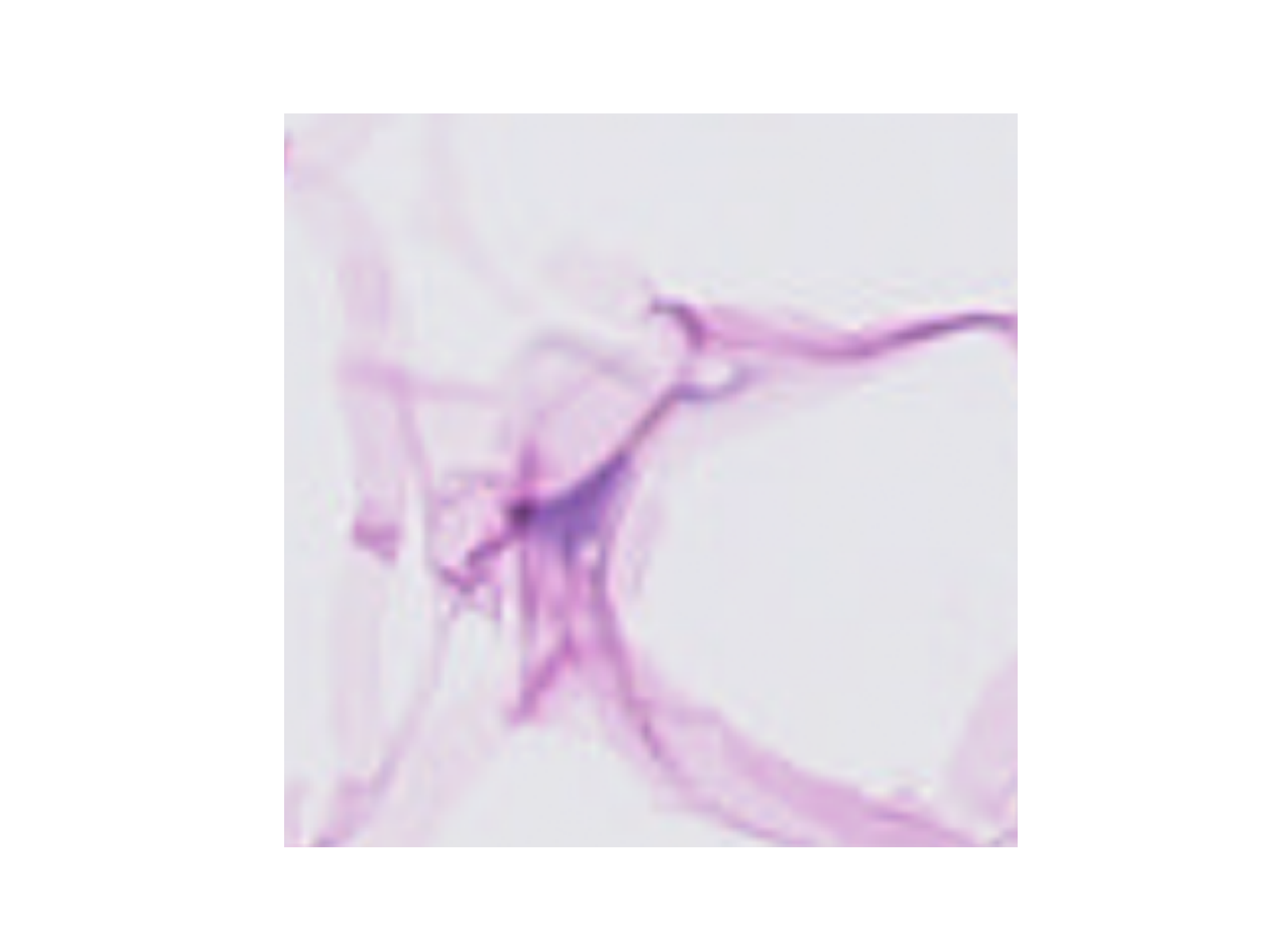}
    &
    \hspace{-3em}
    \includegraphics[width=0.27\textwidth]{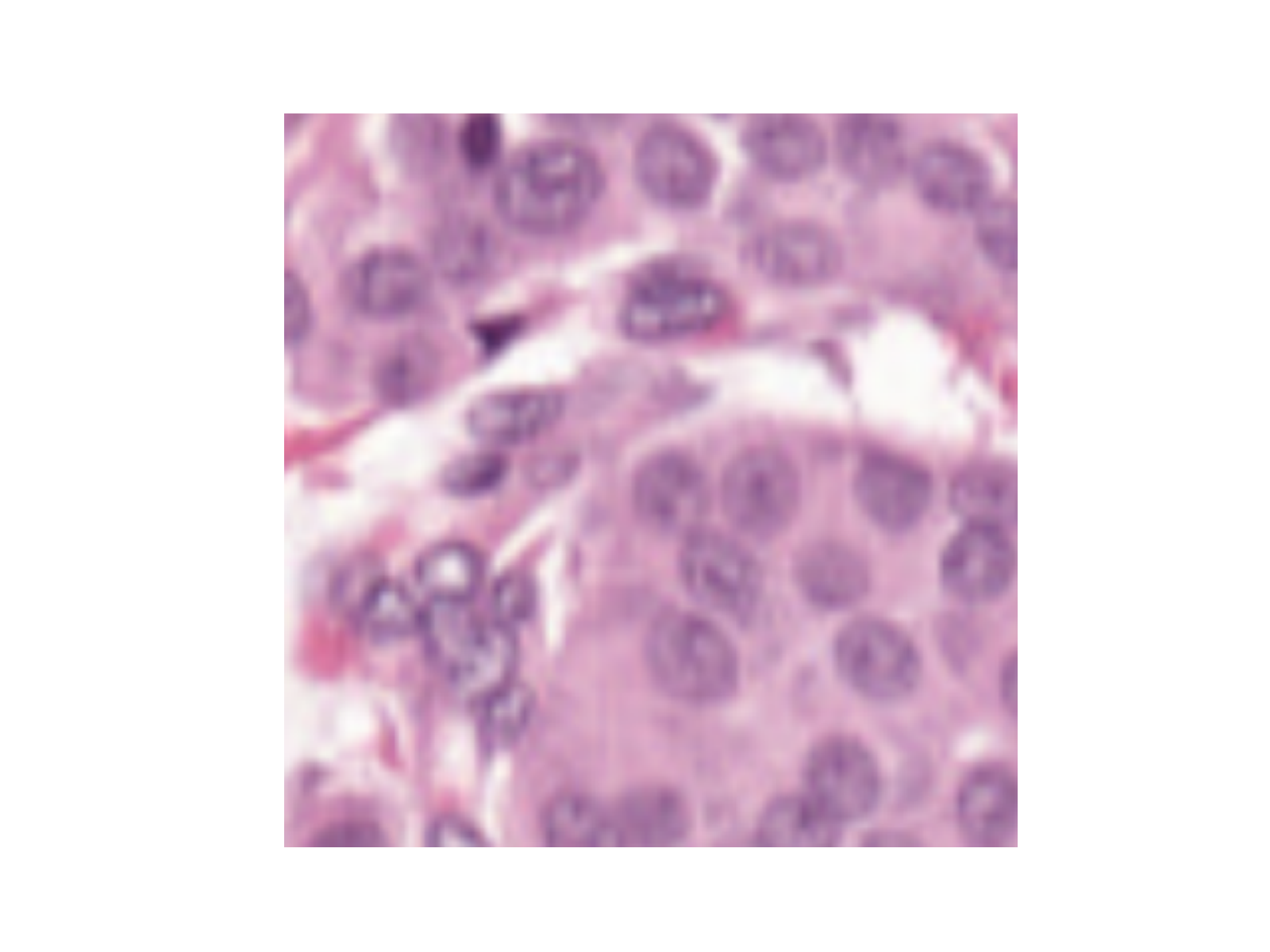}
    &
    \hspace{-3em}
    \includegraphics[width=0.27\textwidth]{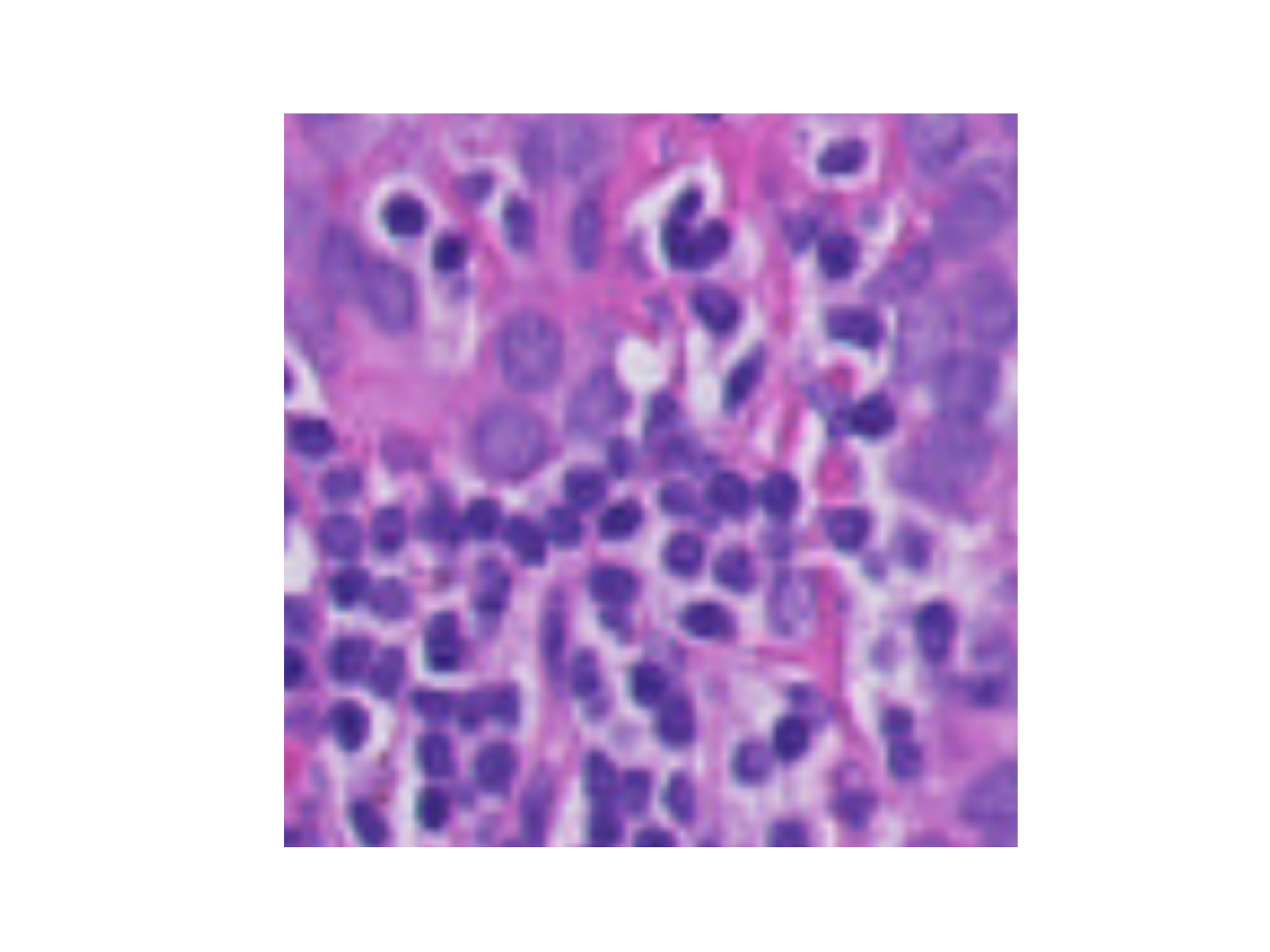}
    \vspace{-1em}
    \\
      \hspace{-3em}
      (a) H123 - No &
      \hspace{-3em}
      (b) H5 - No &
      \hspace{-3em}
      (c) H123 - Cancer &
      \hspace{-3em}
      (d) H5 - Cancer
    \end{tabular}
    \caption{
      \textbf{Example images on Camelyon-17.}
      The objective is to identify if a lymph node slide has cancer (no cancer $y=0$ vs metastases $y=1$)
      where the spurious features are the hospital slide characteristics.
      In this case, the in-distribution sources are hospitals 1, 2, and 3 (H123) and
      the out-of-distribution source is hospital 5 (H5).
      The groups are $\mathcal{G}_{1}=\text{H123 - no}$,
      $\mathcal{G}_{2}=\text{H4 - no}$,
      $\mathcal{G}_{3}=\text{H123 - Cancer}$, and
      $\mathcal{G}_{4}=\text{H4 - Cancer}$.
      In terms of number of data samples per group we have
      $\mathcal{G}_{1}=151,390(39\%)$,
      $\mathcal{G}_{2}=42,527(11\%)$,
      $\mathcal{G}_{3}=151,046(38\%)$, and
      $\mathcal{G}_{4}=42,627(12\%)$.
      In contrast to other datasets, $\mathcal{G}_{2}$ and $\mathcal{G}_{4}$ are not observed in the training data.
    }
    \label{fig:examples_camelyon}
\end{figure*}

\begin{figure*}
    \centering
    \begin{tabular}{cccccccccc}
    \hspace{-2em}
    \includegraphics[width=0.13\textwidth]{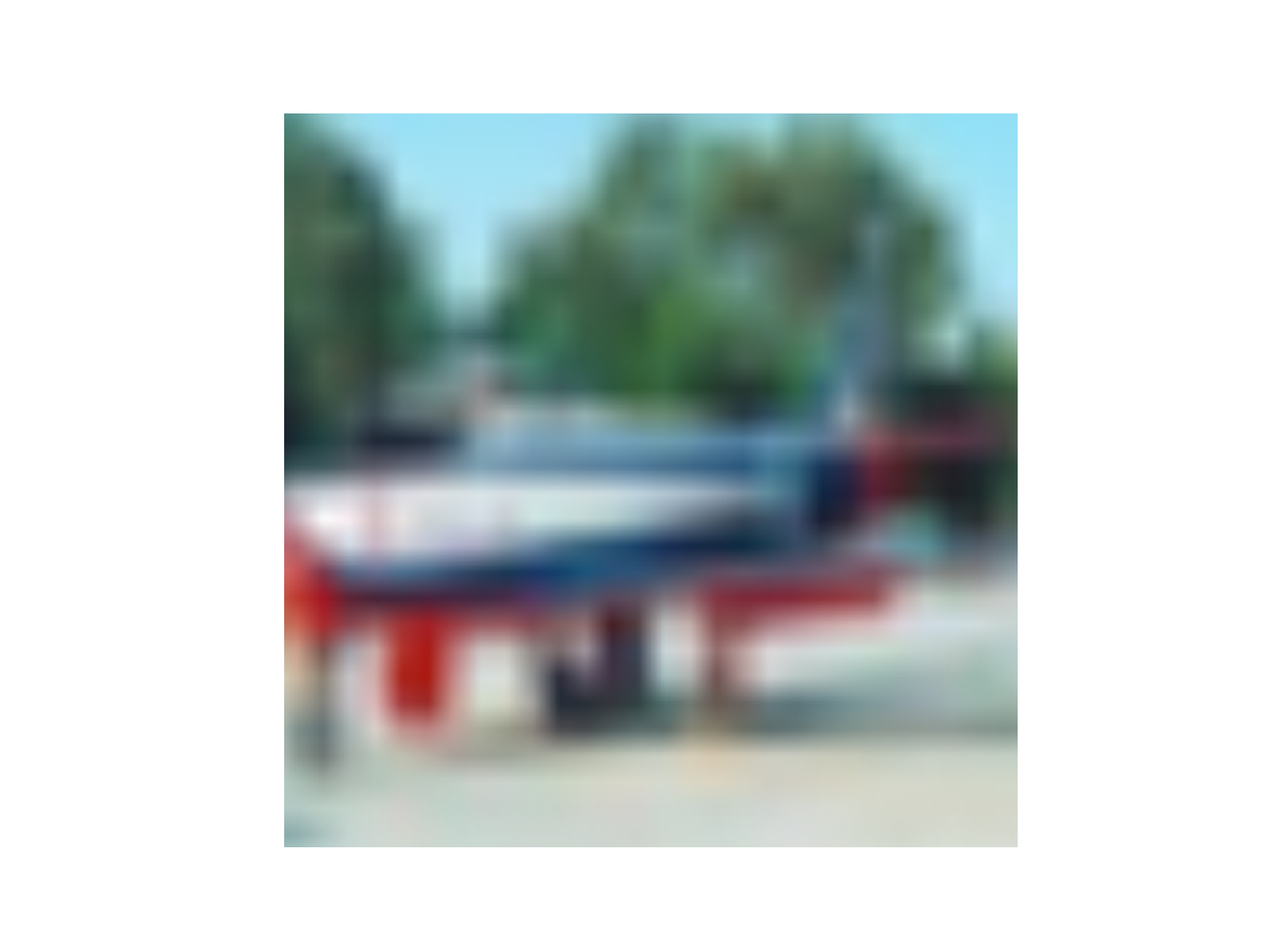}
      &
    \hspace{-2em}
    \includegraphics[width=0.13\textwidth]{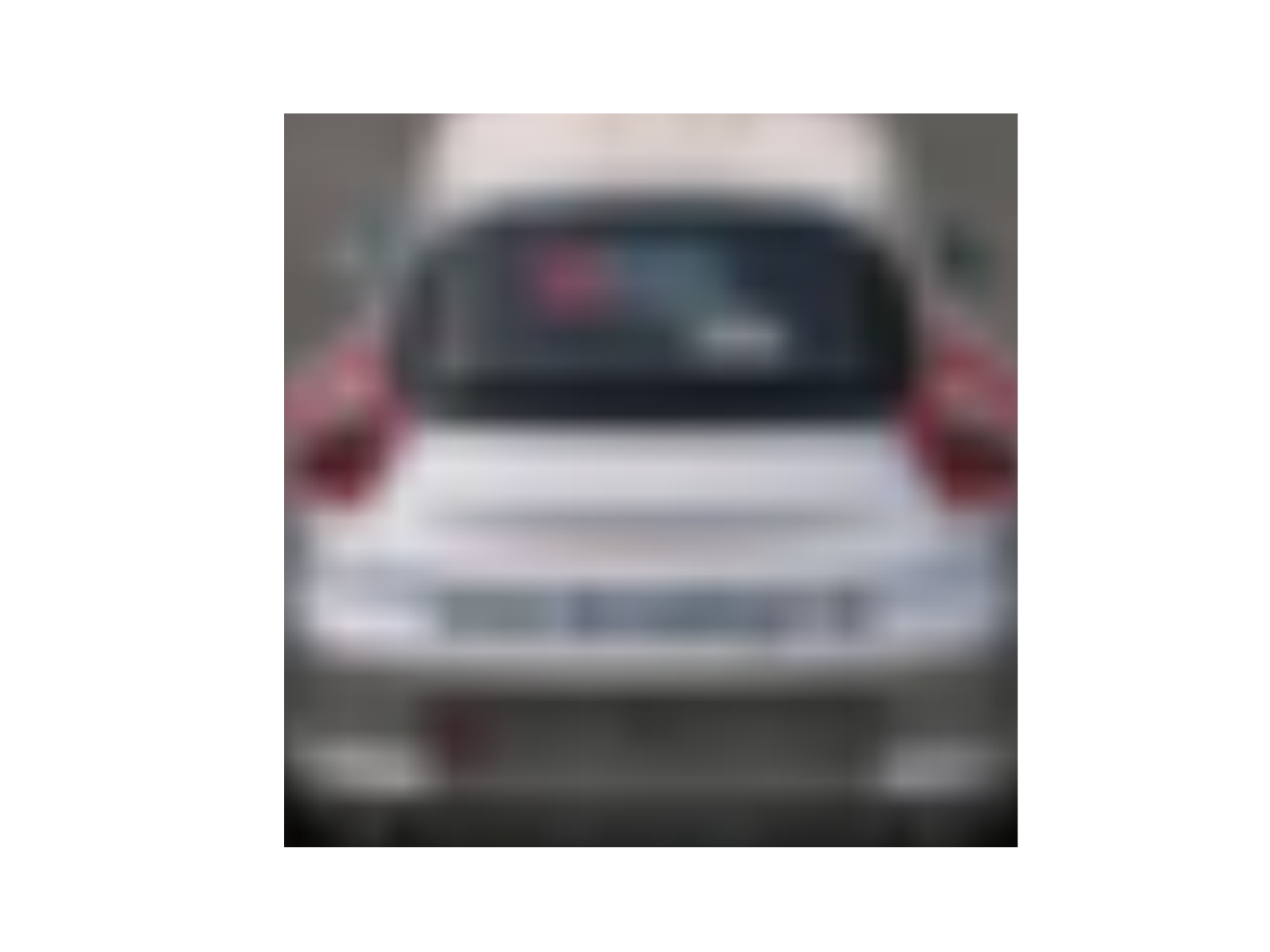}
    &
    \hspace{-2em}
    \includegraphics[width=0.13\textwidth]{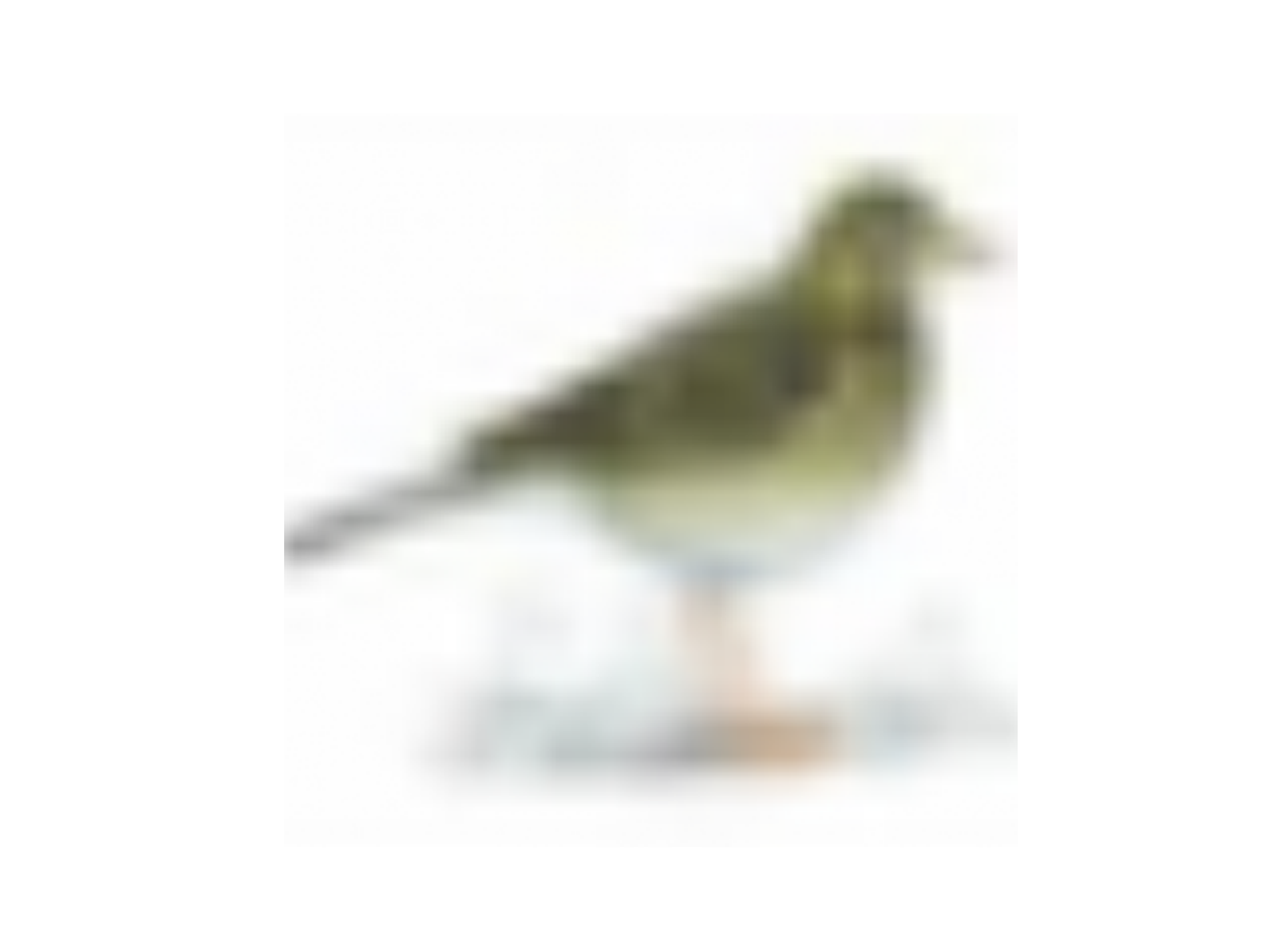}
    &
    \hspace{-2em}
    \includegraphics[width=0.13\textwidth]{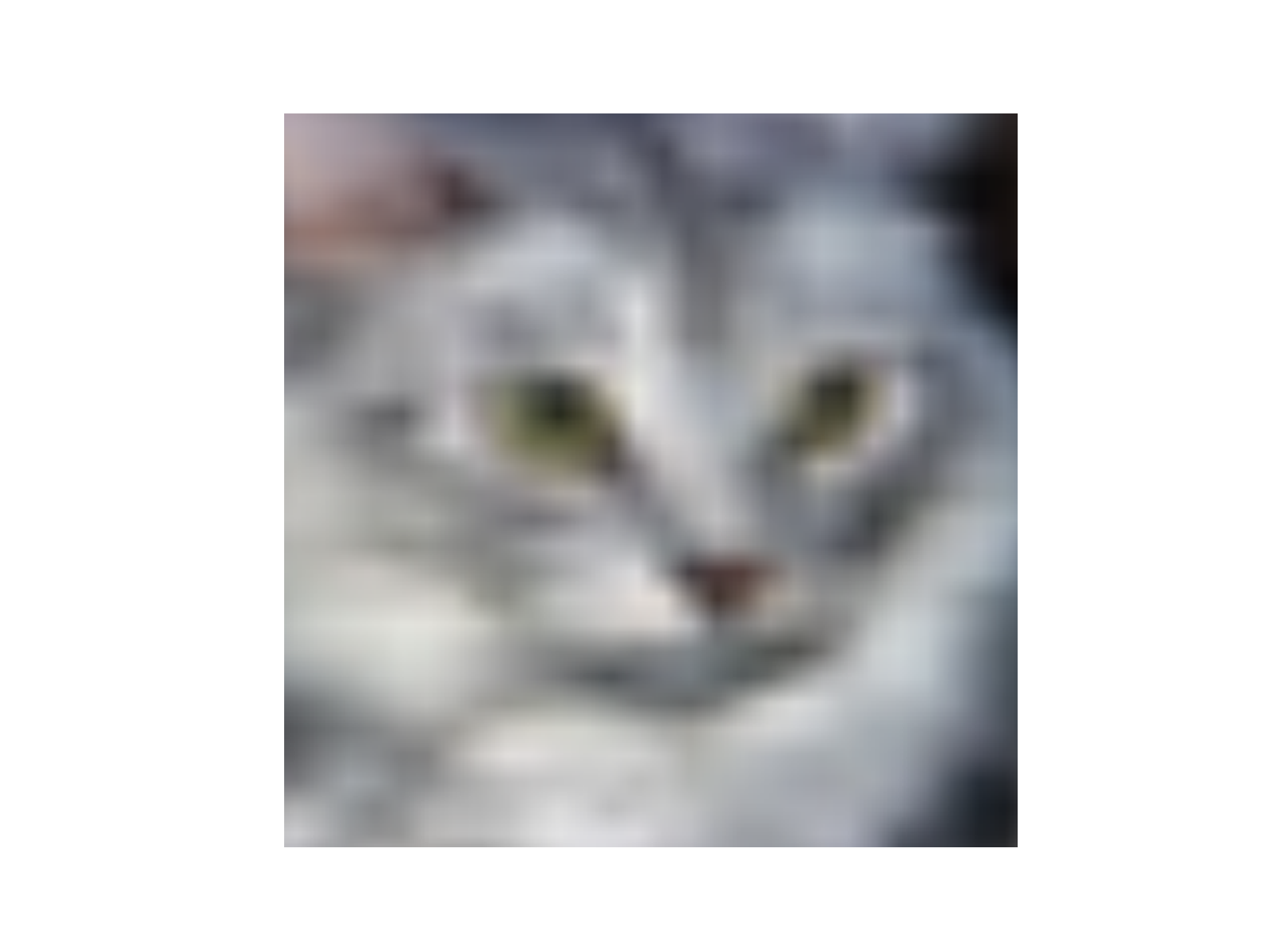}
    &
    \hspace{-2em}
    \includegraphics[width=0.13\textwidth]{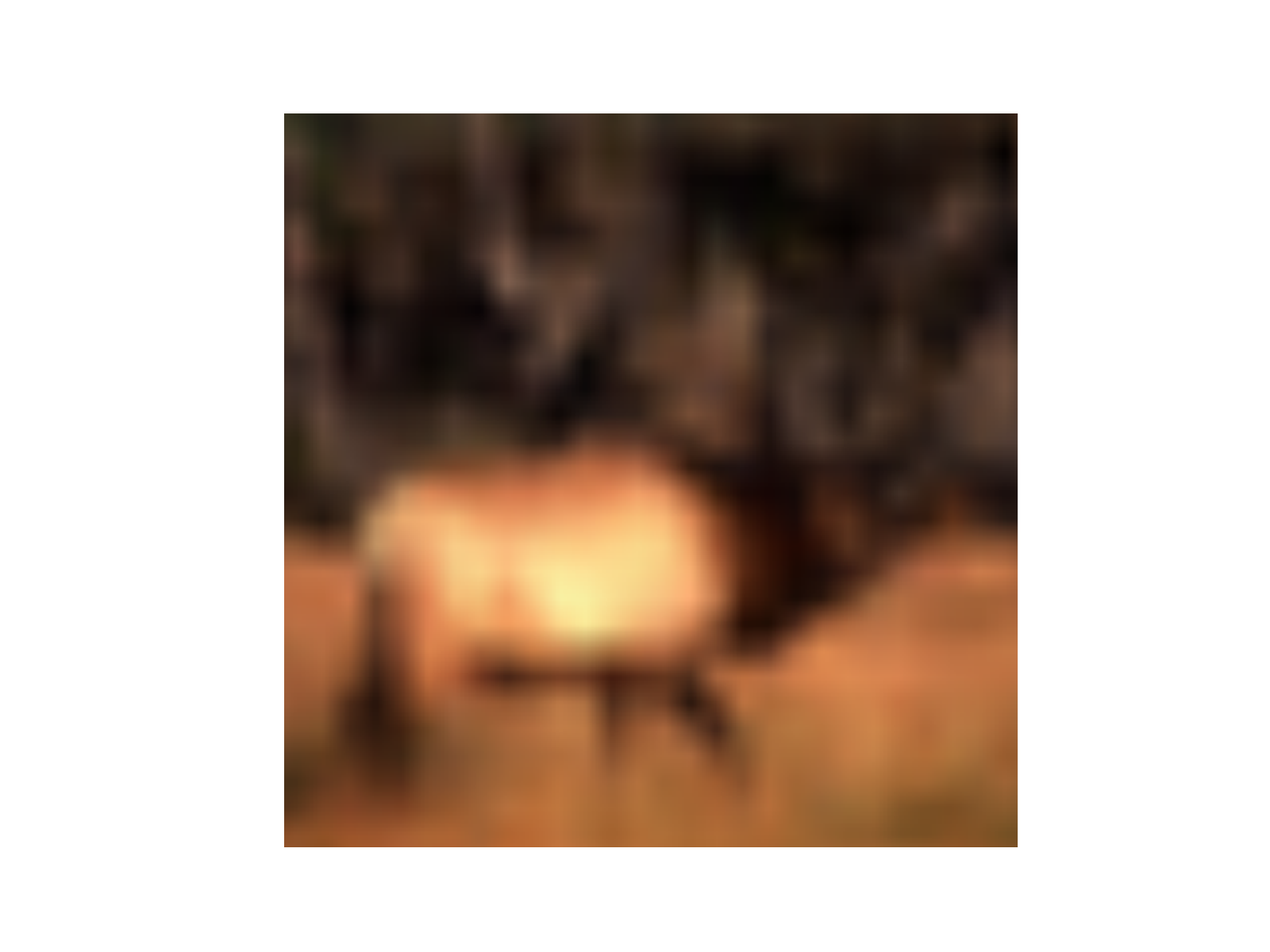}
    &
    \hspace{-2em}
    \includegraphics[width=0.13\textwidth]{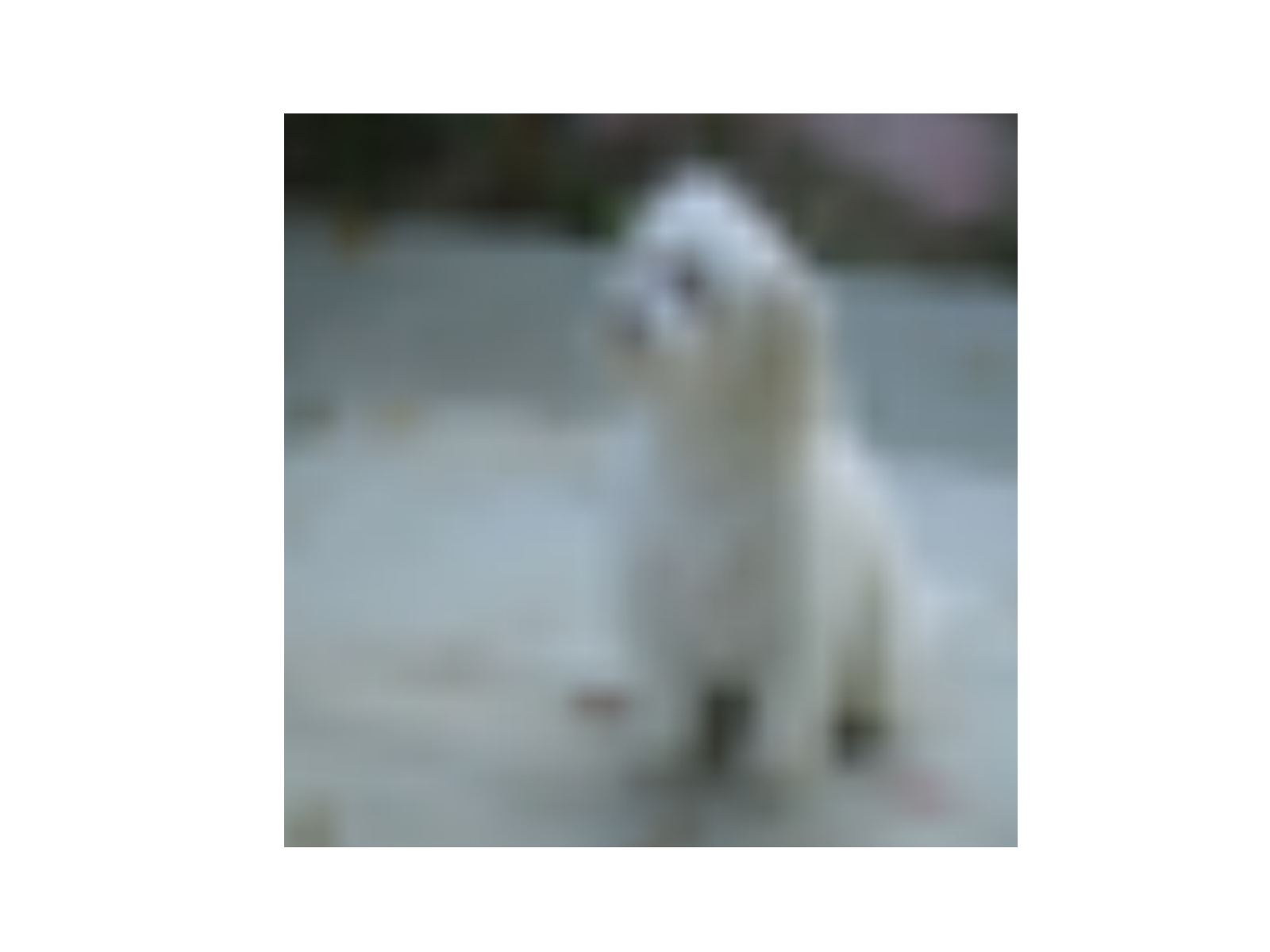}
    &
    \hspace{-2em}
    \includegraphics[width=0.13\textwidth]{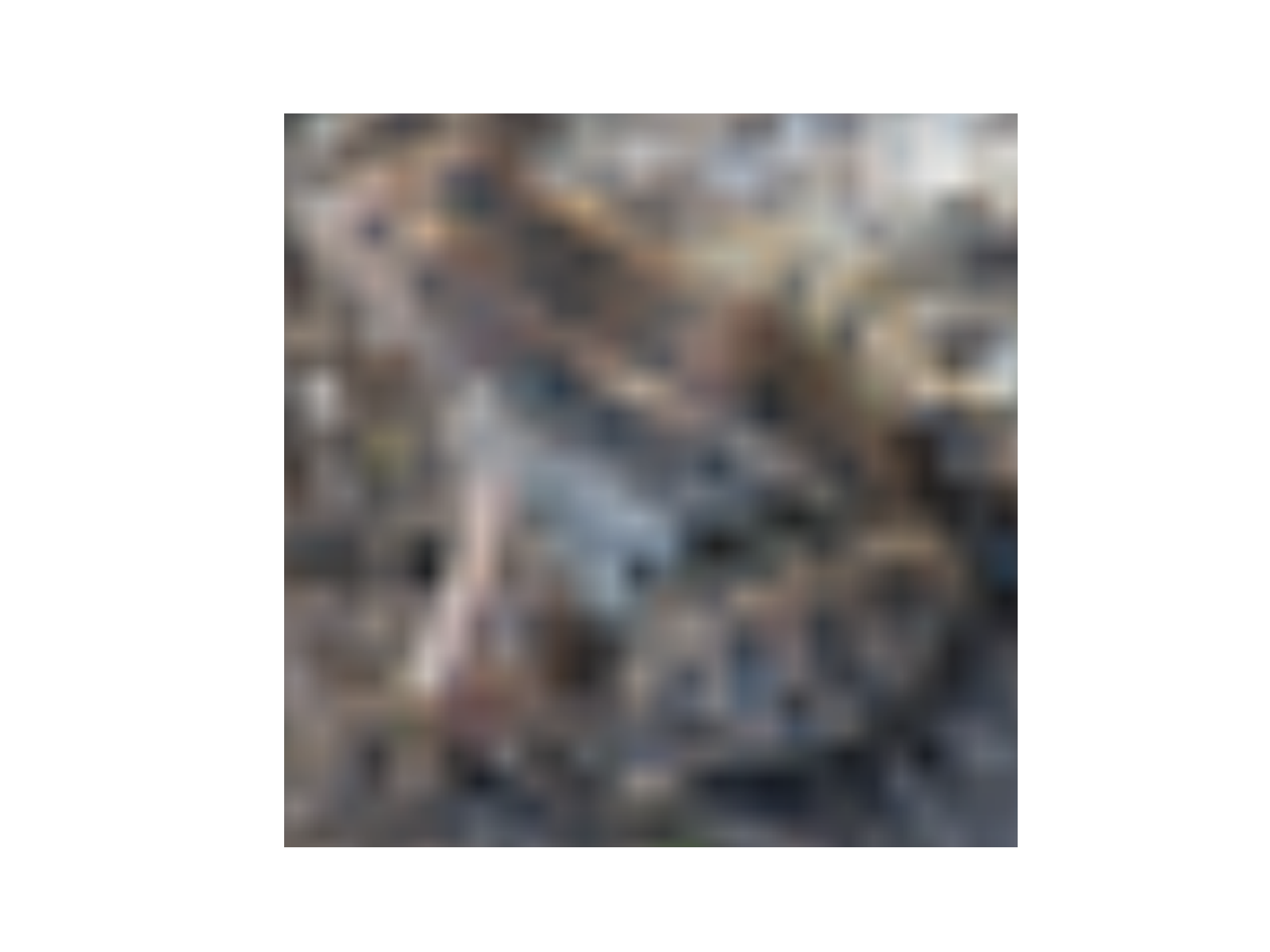}
    &
    \hspace{-2em}
    \includegraphics[width=0.13\textwidth]{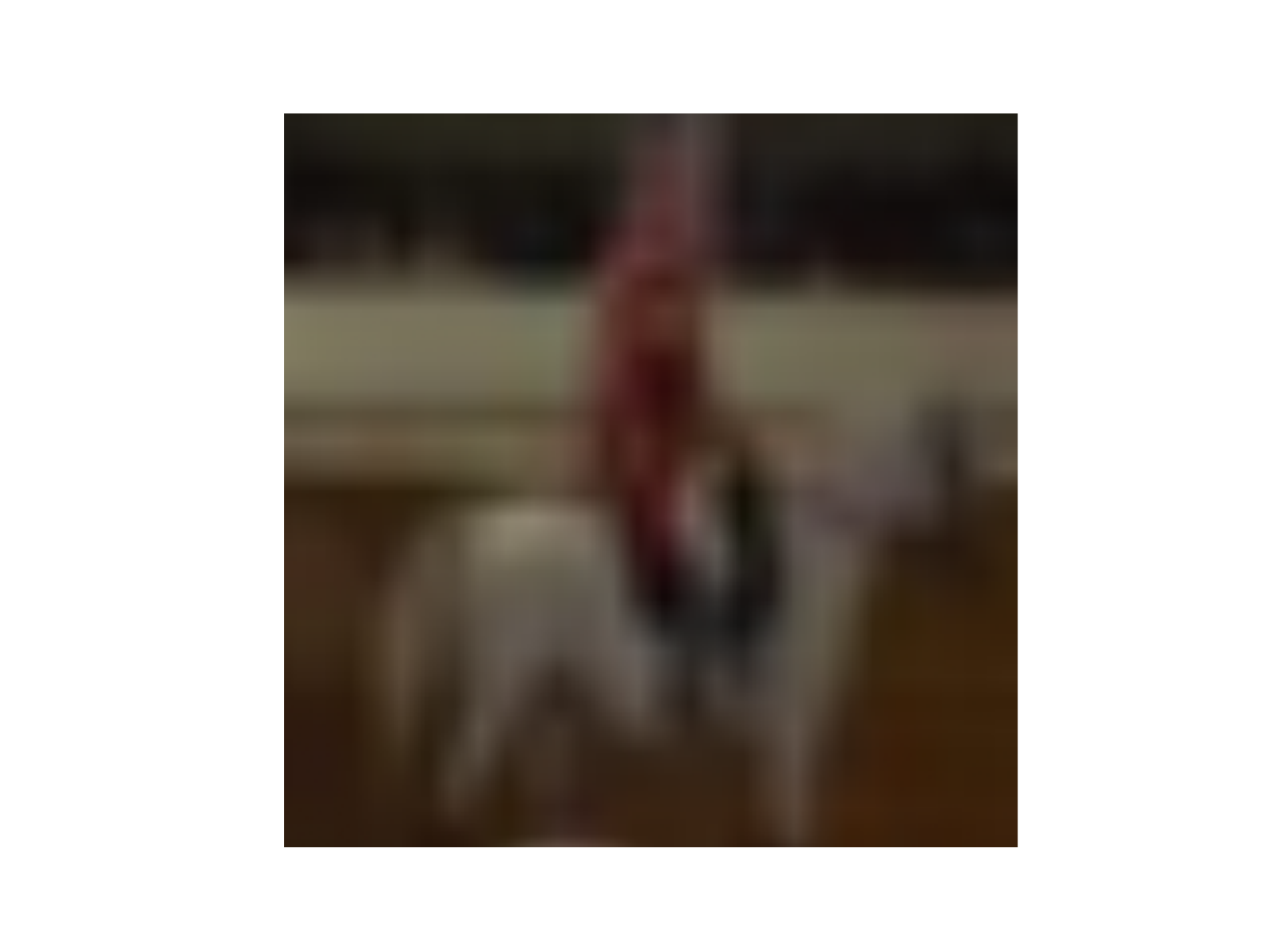}
    &
    \hspace{-2em}
    \includegraphics[width=0.13\textwidth]{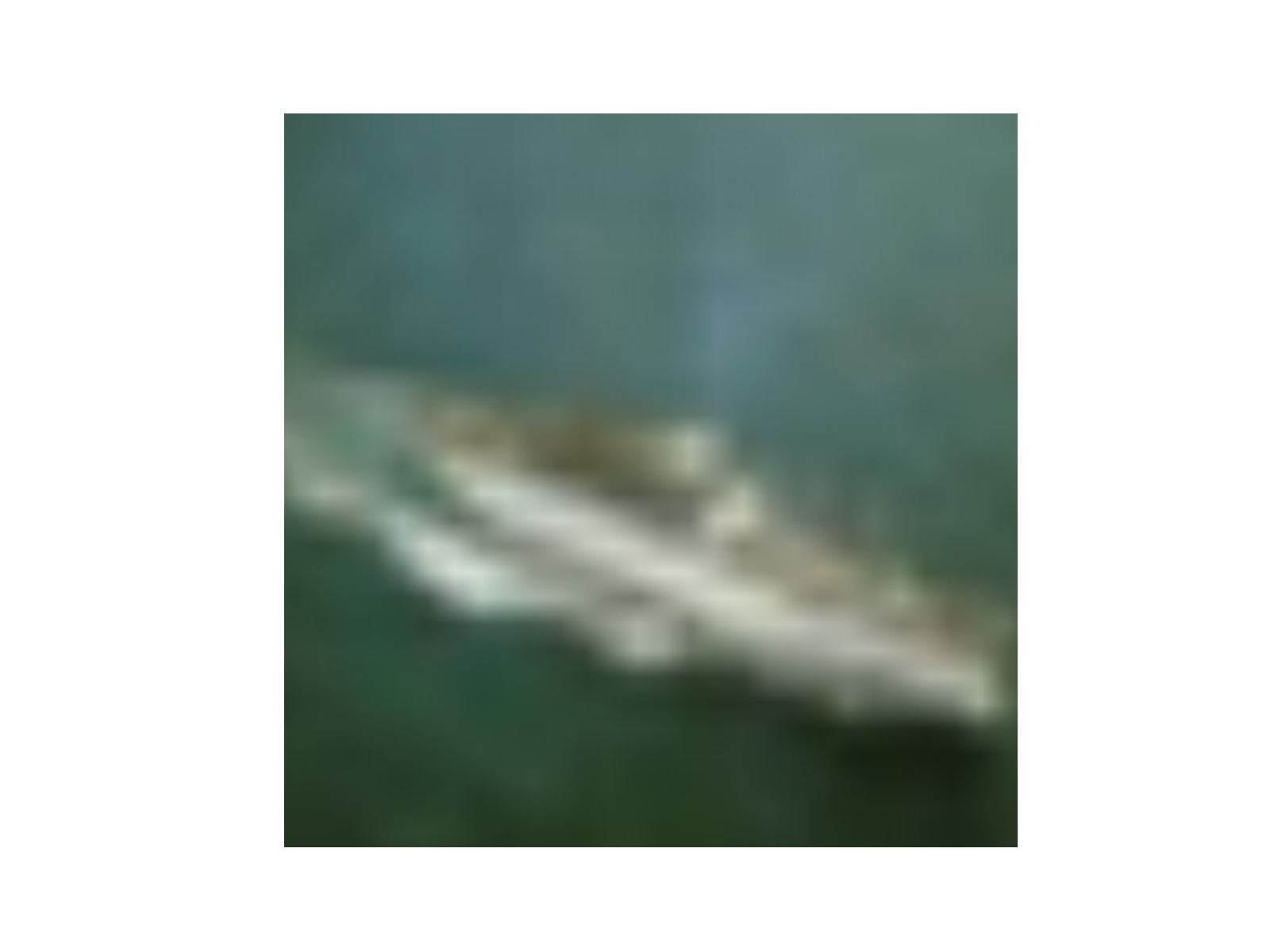}
    &
    \hspace{-2em}
    \includegraphics[width=0.13\textwidth]{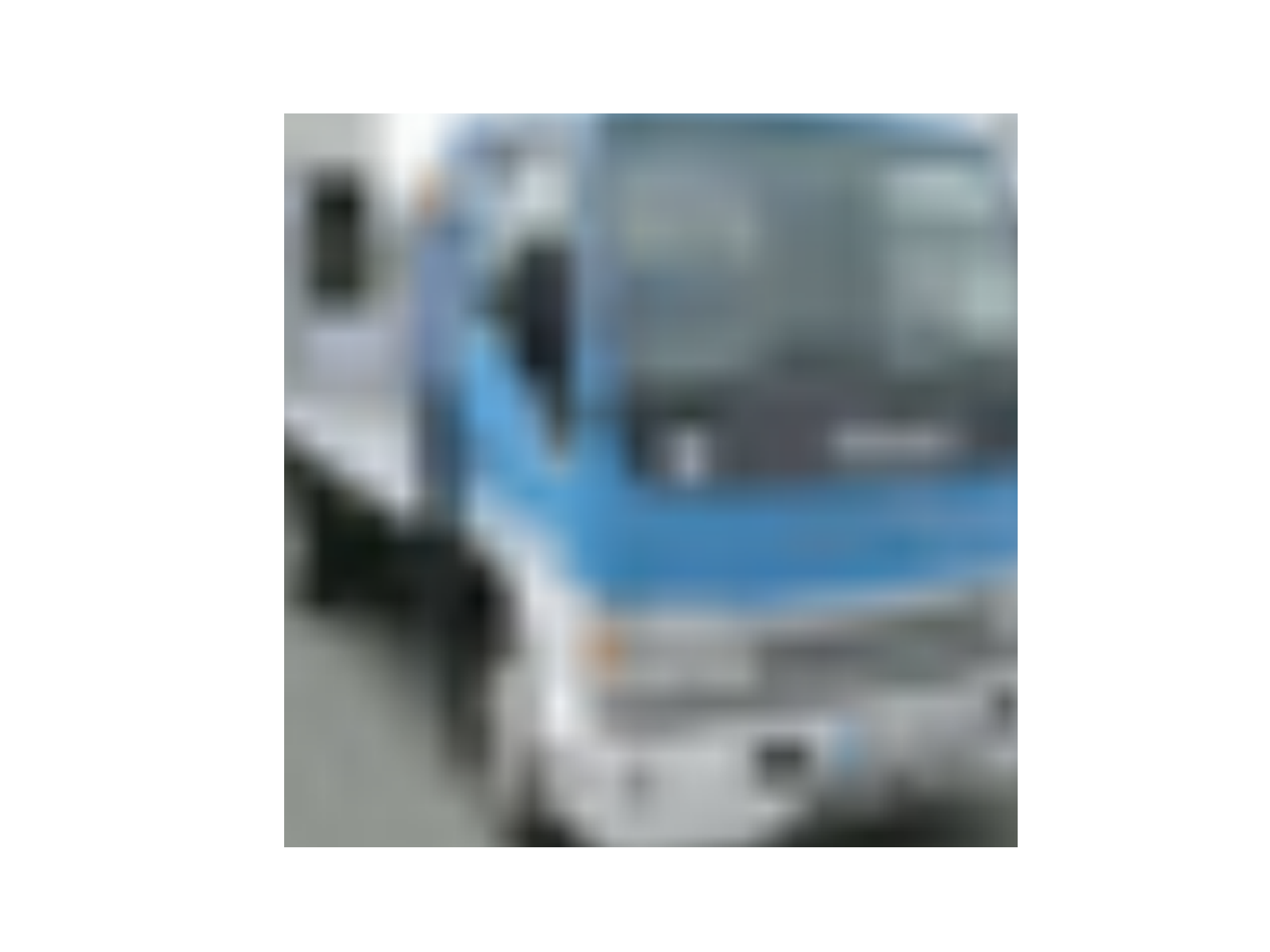}
    \vspace{-0.5em}
    \\
      \hspace{-2em}
      airplane &
      \hspace{-2em}
      car &
      \hspace{-2em}
      bird &
      \hspace{-2em}
      cat &
      \hspace{-2em}
      deer &
      \hspace{-2em}
      dog &
      \hspace{-2em}
      frog &
      \hspace{-2em}
      horse &
      \hspace{-2em}
      ship &
      \hspace{-2em}
      truck
    \end{tabular}
    \caption{
      \textbf{Example images on CIFAR-10.}
      CIFAR-10 consist of tiny images of 10 different categories: airplane ($y=0$),
      automobile ($y=1$), bird ($y=2$), cat ($y=3$),
      deer ($y=4$), dog ($y=5$), frog ($y=6$),
      horse ($y=7$), ship ($y=8$) and truck ($y=9$).
      The train size if 50K and each class
      consists of 8.3K images.
    }
    \label{fig:examples_cifar}
\end{figure*}

\begin{figure}
\resizebox{\textwidth}{!}{
\begin{tabular}{cccccc}
\hline\\[-3mm]
\multicolumn{5}{c}{\textbf{MultiNLI}} \\
\hline\\[-3mm]

& \textbf{Text examples} &
\begin{tabular}{c}\textbf{Class} \\ \textbf{label} \end{tabular}&
\textbf{Description} &
\begin{tabular}{c}\textbf{\# Train} \\ \textbf{data} \end{tabular} &
\\

$\mathcal{G}_1$ &
\begin{tabular}{c}
\small{\textit{``if residents are unhappy, they can put wheels}}\\
\small{\textit{on their homes and go someplace else, she said.}}\\
\small{\textit{[SEP] residents are stuck here but they can't }}\\
\small{\textit{ go anywhere else.''}} \\[2mm]
\end{tabular} &
0 &
\begin{tabular}{c}
contradiction  \\ no negations
\end{tabular} &
57498 (28\%)
\\

$\mathcal{G}_2$ & \begin{tabular}{c}
\small{\textit{``within this conflict of values is a clash }} \\
\small{\textit{about art. [SEP] there is \underline{no} clash about art.'' }} \\[2mm]
\end{tabular} &
0 &
\begin{tabular}{c}
contradiction  \\ has negations
\end{tabular}
& 11158 (5\%)
\\

$\mathcal{G}_3$ & \begin{tabular}{c}
\small{\textit{``there was something like amusement in  }} \\
\small{\textit{the old man's voice. [SEP]   }} \\
\small{\textit{the old man showed amusement.''}} \\[2mm]
\end{tabular}&
1 &
\begin{tabular}{c}
entailment  \\ no negations
\end{tabular}
& 67376 (32\%)
\\

$\mathcal{G}_4$ & \begin{tabular}{c}
\small{\textit{``in 1988, the total cost for the postal service}} \\
\small{\textit{ was about \$36. [SEP] the postal service cost us }} \\
\small{\textit{ citizens almost \underline{nothing} in the late 80's. ''}} \\[2mm]
\end{tabular} &
1 &
\begin{tabular}{c}
entailment  \\ has negations
\end{tabular}
& 1521 (1\%)
\\
$\mathcal{G}_5$ & \begin{tabular}{c}
\small{\textit{``yeah but even even cooking over an open fire }} \\
\small{\textit{ is a little more fun isn't it [SEP] }} \\
\small{\textit{i like the flavour of the food.''}} \\[2mm]
\end{tabular}&
2 &
\begin{tabular}{c}
neutral  \\ no negations
\end{tabular}
& 66630 (32\%)
\\
$\mathcal{G}_6$ & \begin{tabular}{c}
\small{\textit{``that's not too bad [SEP]  }} \\
\small{\textit{it's better than \underline{nothing}''}}
\end{tabular} &
2 &
\begin{tabular}{c}
neutral  \\ has negations
\end{tabular}
& 1992 (1\%)
\\[2mm]
\multicolumn{5}{l}{
\textbf{Target}: contradiction / entailment / neutral;\quad\quad \textbf{Spurious feature}: has negation words.\quad\quad \textbf{Minority}: $\mathcal{G}_4$, $\mathcal{G}_6$} \\
\\[2mm]
\hline
\\[-3mm]

\multicolumn{5}{c}{\textbf{CivilComments}}
\\[0mm]\hline\\[-3mm]
& \textbf{Text examples} &
\begin{tabular}{c}\textbf{Class} \\ \textbf{label} \end{tabular}&
\textbf{Description} &
\begin{tabular}{c}\textbf{\# Train} \\ \textbf{data} \end{tabular}
\\[4mm]

&
\begin{tabular}{c}
\small{\textit{``I'm quite surprised this worked for you. }} \\
\small{\textit{Infrared rays cannot penetrate tinfoil.''}} \\[2mm]
\end{tabular}&
0 &
\begin{tabular}{c}
non-toxic  \\ no identities
\end{tabular}&
148186 (55\%)
\\

& \begin{tabular}{c}
\small{\textit{``I think you may have misunderstood what  }} \\
\small{\textit{'straw \underline{men}' are. But I'm glad that your gravy is good.'' }} \\[2mm]
\end{tabular}&
0 &
\begin{tabular}{c}
non-toxic  \\ has identities
\end{tabular}
& 90337 (33\%)
\\

& \begin{tabular}{c}
\small{\textit{``Hahahaha putting his faith in Snopes. Pathetic.''}}\\[2mm]
\end{tabular}&
1 &
\begin{tabular}{c}
toxic  \\ no identities
\end{tabular}
& 12731 (5\%)
\\

& \begin{tabular}{c}
\small{\textit{``That sounds like something a \underline{white person} would say.'''}} \\[2mm]
\end{tabular} &
1 &
\begin{tabular}{c}
toxic  \\ has identities
\end{tabular}
& 17784 (7\%)
\\

\multicolumn{5}{l}{
\textbf{Target}: Toxic / not toxic comment;\quad\quad \textbf{Spurious feature}: mentions protected categories.} \\
\\
[0.3mm]
\hline

\end{tabular}
}
\caption{
Dataset examples for MultiNLI and CivilComments.
We underline the words corresponding to the spurious feature.
CivilComments contains $16$ overlapping groups corresponding to toxic / non-toxic comments and mentions of one of the protected identities:
male, female, LGBT, black, white, Christian, Muslim, other religion.
We only show examples with mentions of the male and white identities.
}
\label{fig:nlp_desc}
\end{figure}

\section{Additional Ablations}
\label{sec:app_ablations}

\subsection{Datasets without known spurious features}
\label{sec:cifar}

We evaluate \method on the CIFAR-10 \citep{krizhevsky2009learning} and Camelyon17 \citep{zech2018variable,koh2021wilds} datasets where there are no clearly identified spurious features
that we can label.
For CIFAR-10 we compute the test worst class accuracy and test mean accuracy.
For CIFAR-10 we use ResNet-18 \citep{he2016deep} pre-trained on ImageNet1k and for Camelyon17 we use ResNet-50 pretrained on ImageNet1k.
As Camelyon17 is a domain generalization benchmark, we compute the test mean accuracy on hospital 2 and hospital 4, which are not represented in the training data (hospitals 1, 3, and 5).
Reassuringly, we find that \method only marginally affects the performance on all evaluations compared to ERM, suggesting that it can be safely applied even in situations when we are unsure if the data contains spurious features.

\begin{table}[t]
\begin{center}
  \footnotesize{
\begin{tabular}{c cc c cc}
\hline\noalign{\smallskip}
\multirow{2}{*}{\textbf{Method}}
  & \multicolumn{2}{c}{\textbf{CIFAR-10}}
  && \multicolumn{2}{c}{\textbf{Camelyon17}}
\\
\cline{2-3}
\cline{5-6}
\\[-3mm]
  & Worst(\%) & Mean(\%)
  && H123 Mean(\%) & H5 Mean(\%)
\\[1mm]\hline\\[-3mm]
ERM &
    $89.6$ & $95.1$
    &&
    $85.0$ & $90.1$
\\
\method &
    $89.3_{\pm 0.5}$ & $94.9_{\pm 0.1}$
    &&
    $87.1_{\pm 0.1}$ & $88.9_{\pm 0.4}$
\\[-0.1mm]\hline
\end{tabular}
}
\end{center}
\caption{
\textbf{Performance on data with unidentified spurious features.}
For CIFAR-10 we report test worst class accuracy and mean accuracy.
For Camelyon17 we report  mean accuracy for two different hospital's data not seen during training.
H123 relates to data from hospitals 1, 2, and 3
used for training and H5 to the data from
hospital 5 which is out of distribution.
When spurious correlations are not identified, \method recovers the standard ERM performance.
}
\label{tab:unidentified}
\end{table}

\subsection{Sensitivity to $\gamma$ on CelebA}
\label{sec:other_gamma}
\begin{figure}[!ht]
    \centering
    \subfloat[Test WGA]{
    \includegraphics[width=0.4\columnwidth]{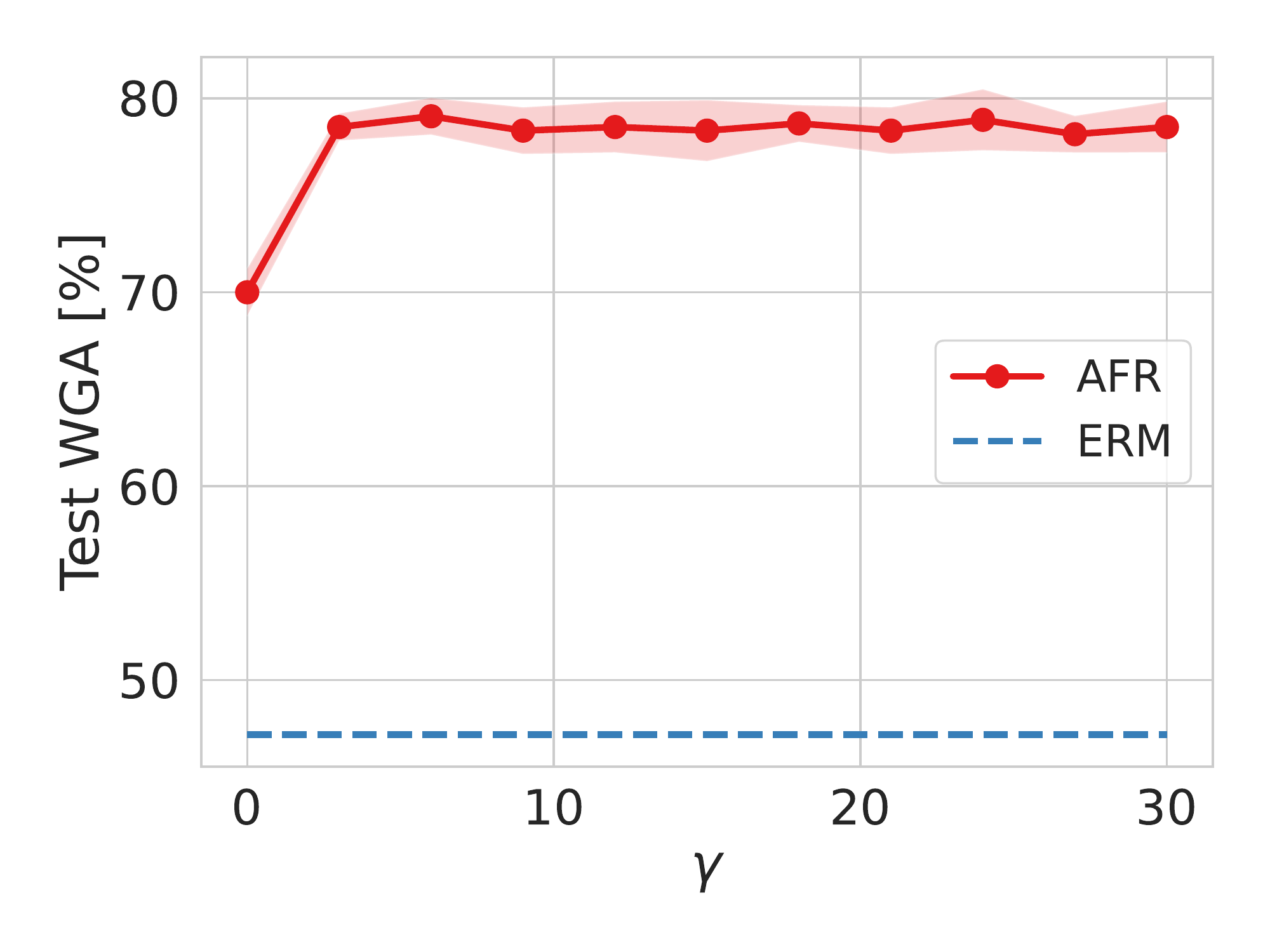}
    \label{fig:celeba_wga_vs_gamma}
    }
    \subfloat[Effective sample size]{\includegraphics[width=0.4\columnwidth]{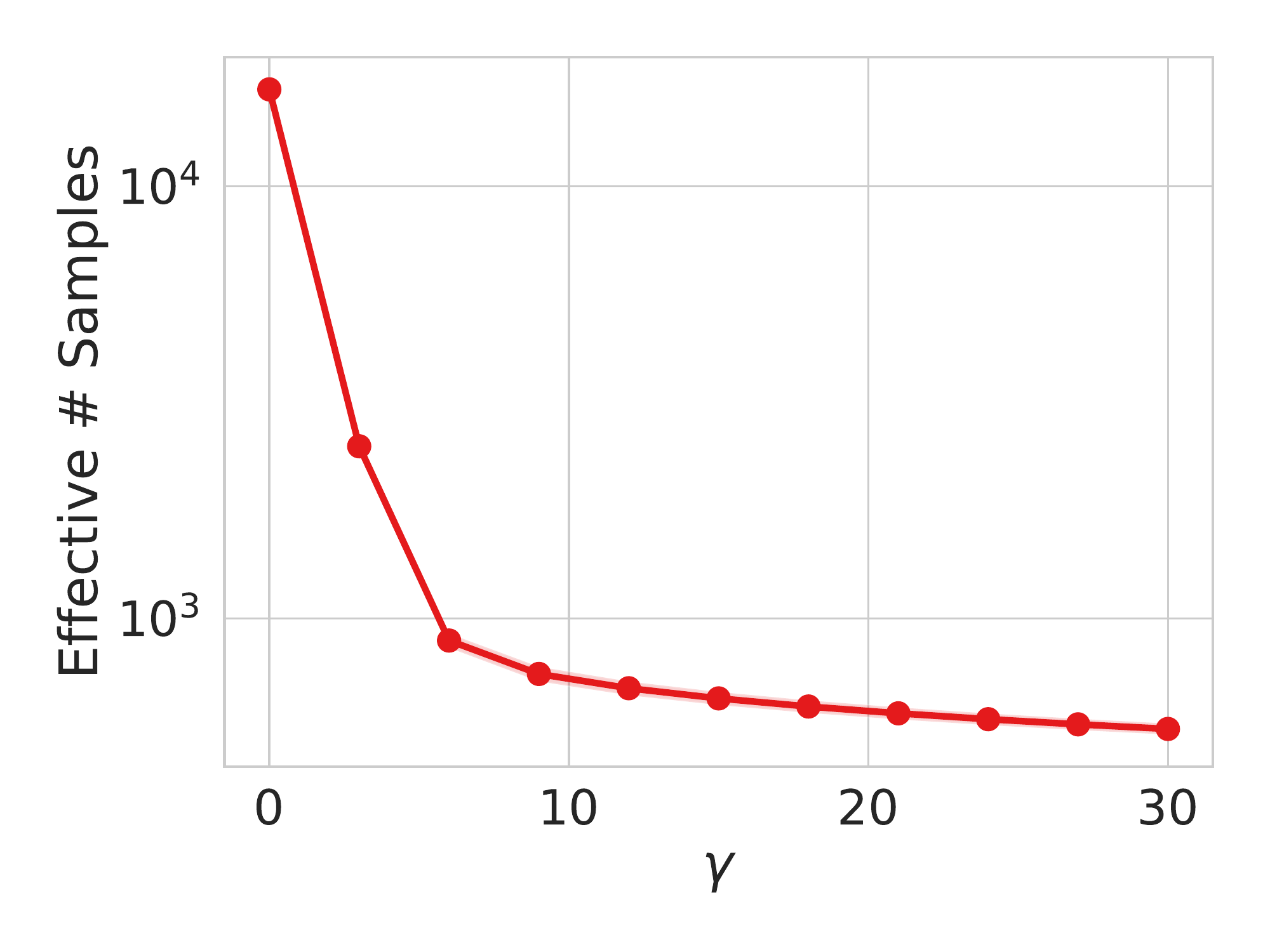}
    \label{fig:celeba_effsize_vs_gamma}
    }
    \caption{
    \textbf{\method's $\gamma$ robustness on CelebA.}
    \textbf{(a)} \method\ achieves high test worst group accuracy on CelebA across a wide range of $\gamma.$ For simplicity, we do not tune the regularization parameter $\lambda$ and set it to 0. We show mean and standard deviation across 3 runs.
    \textbf{(b)} The effective sample size decreases and then stabilizes as $\gamma$ increase.
    }
    \label{fig:celeba_gamma}
\end{figure}

Similar to what we observed on Waterbirds, Figure~\ref{fig:celeba_gamma} shows \method\ performs well across a wide range of $\gamma$. We set $\lambda = 0$ for simplicity. On both CelebA and Waterbirds, \method\ can be applied to improve worst group accuracy without tuning either $\gamma$ or $\lambda.$

\subsection{Ablating early stopping and $\ell_2$ regularization}
Similar to \jtt and \cnc, AFR employs early stopping during the 2nd stage. We re-ran AFR without early stopping while allowing the other hyperparameters ($\gamma$ and $\lambda$) to be tuned as usual. As expected, AFR’s performance decreases, but we still observe substantial improvement over ERM, showing that 2nd stage early stopping helps but is not essential to AFR’s performance gain. Similarly, we found ablating  $\ell_2$ regularization (setting $\lambda = 0$ and only tuning $\gamma$) only slightly degrades AFR's performance. These findings show AFR's performance gain does not rely on carefully tuned regularization. Table \ref{tab:ablate_regularization} summarizes these results.

\begin{table}[!h]
\centering
\begin{tabular}{lccc}
\hline
\textbf{Method} & \textbf{Waterbirds} & \textbf{MultiNLI} & \textbf{Chest X-Ray} \\
\hline
AFR & $0.022$ & $0.07$ & $0.10$ \\
\hline
\end{tabular}
\caption{\textbf{P-values} testing
the null hypothesis that AFR’s mean test WGA is lower than 2nd best performing method.
When the 2nd best performing method is CnC, we perform a two-sample t-test; when it’s JTT, we perform a one-sample t-test since only its mean was reported}
\label{tab:pvals}
\end{table}

\begin{table}[!h]
\centering
\begin{tabular}{lccc}
\hline
\textbf{Method} & \textbf{Waterbirds} & \textbf{CelebA} & \textbf{MultiNLI} \\
\hline
ERM & $72.6$ & $47.2$ & $67.9$ \\
AFR & $90.4_{\pm0.1}$ & $82.0_{\pm0.5}$ & $73.4_{\pm0.6}$ \\
AFR w/o $\ell_2$ reg. & $90.1_{\pm0.7}$ & $81.3_{\pm1.8}$ & $71.1_{\pm1.4}$ \\
AFR w/o early stopping & $89.9_{\pm0.7}$ & $80.0_{\pm1.6}$ & $71.1_{\pm1.4}$ \\
\hline
\end{tabular}
\caption{\textbf{AFR's performance without early stopping or $\ell_2$ regularization}. AFR's performance gain does not rely on carefully tuned regularization. Reported mean and standard deviation are computed over three independent runs.}
\label{tab:ablate_regularization}
\end{table}

\subsection{Improving group robustness with no hyperparameter tuning}
\label{sec:no-tuning}
To show that AFR improves group robustness without carefully tuning $\gamma$, $\ell_2$ regularization, or early stopping, we run AFR on Waterbirds and CelebA where we set $\gamma$ to four positive values $\{1, 2, 4, 8\}$ and run it for $100$ steps without $\ell_2$ regularization. We found in most cases AFR still significantly improves test WGA compared to ERM. While the optimal value for $\gamma$ depends on the dataset, most values between $1$ and $8$ lead to substantial improvement over ERM with the exception of $\gamma=1$ on Waterbirds where the value appears to be too low for AFR upweight the minority group examples.

\begin{table}[!h]
\centering
\begin{tabular}{lcc}
\hline\noalign{\smallskip}
 & \textbf{Waterbirds} & \textbf{CelebA} \\
\hline\noalign{\smallskip}
$\gamma=1, \lambda=0$, 100 steps & $72.3_{\pm1.8}$ & $75.9_{\pm1.7}$ \\
$\gamma=2, \lambda=0$, 100 steps & $76.2_{\pm3.5}$ & $81.1_{\pm1.4}$ \\
$\gamma=4, \lambda=0$, 100 steps & $84.4_{\pm3.3}$ & $72.1_{\pm1.1}$ \\
$\gamma=8, \lambda=0$, 100 steps & $84.1_{\pm4.7}$ & $61.0_{\pm0.9}$ \\
\hline\noalign{\smallskip}
ERM & $72.6$ & $47.2$ \\
\hline\noalign{\smallskip}
\end{tabular}
\caption{
\textbf{\method's outperforms ERM without hyperparameter tuning.}
AFR improves WGA on Waterbirds and CelebA without tuning hyperparameters at all. Reported mean and standard deviation are computed over three independent runs.}
\label{tab:hyperparams}
\end{table}

\subsection{Robustness to specification of the weights}
\label{sec:ablate_weights}
 While simple and intuitive, the choice to define the weights as $\mu_i \propto \beta_{y_i} \exp(-\gamma \, \hat{p}_i)$ is somewhat arbitrary. However, as the goal is to simply upweight the poorly predicted examples, we again do not expect the performance of \method\ to be sensitive to the exact functional form used for $\mu_i,$ as long as it is large for poorly predicted examples. We verify this by replacing $\exp(-\gamma \, \hat{p}_i)$ with two alternatives $\hat{p}_i^{-\gamma} = \exp(\gamma(-\log \hat{p}_i))$ and $(1 - \hat{p}_i)^\gamma,$ as used by focal loss\footnote{However, unlike in focal loss, our weights are fixed throughout training and do not depend on the parameters being optimized.} \citep{lin2017focal}, and compare the resulting test WGA after tuning $\gamma$ on validation WGA in Table~\ref{tab:ablate_weight_fn}. Indeed, both variants of \method\ are able to drastically improve test WGA compared to ERM and achieve similar performance as the original one.

\begin{table}[ht!]
\begin{center}
  \footnotesize{
\begin{tabular}{cc}
\hline\noalign{\smallskip}
\textbf{Method} & \textbf{Test WGA}\\
\hline\noalign{\smallskip}
ERM
    & $72.6$ \\
\method\ $\exp(-\gamma \, \hat{p}_i)$ &
        $\mathbf{90.4_{\pm1.1}}$\\
\method\ $(1 - \hat{p}_i)^\gamma$ & $89.3_{\pm1.9}$ \\
\method\ $\hat{p}_i^{-\gamma}$ &
        $89.1_{\pm1.3}$
\\\hline\noalign{\smallskip}
\end{tabular}
}
\end{center}
\caption{
\textbf{Performance on alternative functional choices.}
Test WGA achieved on the Waterbirds by ERM and variants of \method\ dataset using alternative definitions for the weights $\mu_i.$ $\gamma$ is always chosen to maximize validation worst group accuracy. The performance of \method\ is not sensitive to the choice of the exact functional form for the weights as long as it upweights poorly predicted examples. The mean and standard deviation are computed over $3$ independent runs.
}
\label{tab:ablate_weight_fn}
\end{table}

\subsection{Robustness to splitting ratio}
\label{sec:ablate_ratio}
The $80\%:20\%$ splitting ratio between $\mathcal{D}_\mathrm{ERM}$ and $\mathcal{D}_\mathrm{RW}$ is intended to keep most of training data for learning a sufficiently good feature extractor and only a small portion for last layer retraining since the latter tends to be much more sample-efficient. To study the sensitivity of \method's performance to this ratio, we compare the default choice with three alternatives. Table~\ref{tab:ablate_ratio} shows the default choice of $80\%:20\%$ performs well and the performance is not very sensitive to the value of this ratio. For all four choices, \method\ significantly improves over ERM on both datasets.

A clear case where \method would fail to improve group robustness is when $\mathcal{D}_\mathrm{RW}$ does not contain any minority group examples to upweight.
Luckily, this scenario is extremely unlikely when $\mathcal{D}_\mathrm{RW}$ is drawn from the train distribution.
For instance, suppose that we are considering a typical deep learning task where the training set has $\geq 10k$ examples.
Moreover, lets suppose the minority group constitutes only $5\%$ of the training data.
Then there are on average 100 minority group examples in the 2nd split, with a standard deviation of around 30.
The probability of having fewer than 10 minority group examples (a $3\sigma$ event) is less than 0.2\%.
Therefore, in practice there will be minority group examples, so long as the user follows our recommendations for the split.

\begin{table}[!h]
\begin{center}
  \footnotesize{
\begin{tabular}{ccc}
\hline\noalign{\smallskip}
$|\mathcal{D}_\mathrm{ERM}| : |\mathcal{D}_\mathrm{RW}|$ & \textbf{Waterbirds} & \textbf{CelebA} \\
\hline\noalign{\smallskip}
50:50 & $88.8_{\pm1.0}$ & $80.9_{\pm0.3}$ \\
70:30 & $89.8_{\pm1.1}$ & $84.2_{\pm1.6}$ \\
80:20 & $90.4_{\pm1.1}$ & $82.0_{\pm0.5}$ \\
90:10 & $89.0_{\pm2.7}$ & $82.7_{\pm0.2}$ \\
\hline\noalign{\smallskip}
ERM & $72.6$ & $47.2$ \\
\hline\noalign{\smallskip}
\end{tabular}
}
\end{center}
\caption{
\textbf{\method is robust to the splitting ratio.}
Test WGA achieved on CelebA by \method\ with different splitting ratios. The default value of $80\%:20\%$ performs well, but the performance is not very sensitive to this value. For all four choices, \method\ significantly improves over ERM. The mean and standard deviation are computed over $3$ independent runs
}
\label{tab:ablate_ratio}
\end{table}

\subsection{Impossibility for group-balanced weights on CelebA}
\label{sec:impossible}
\begin{figure}
    \centering
    \includegraphics[width=0.5\columnwidth]{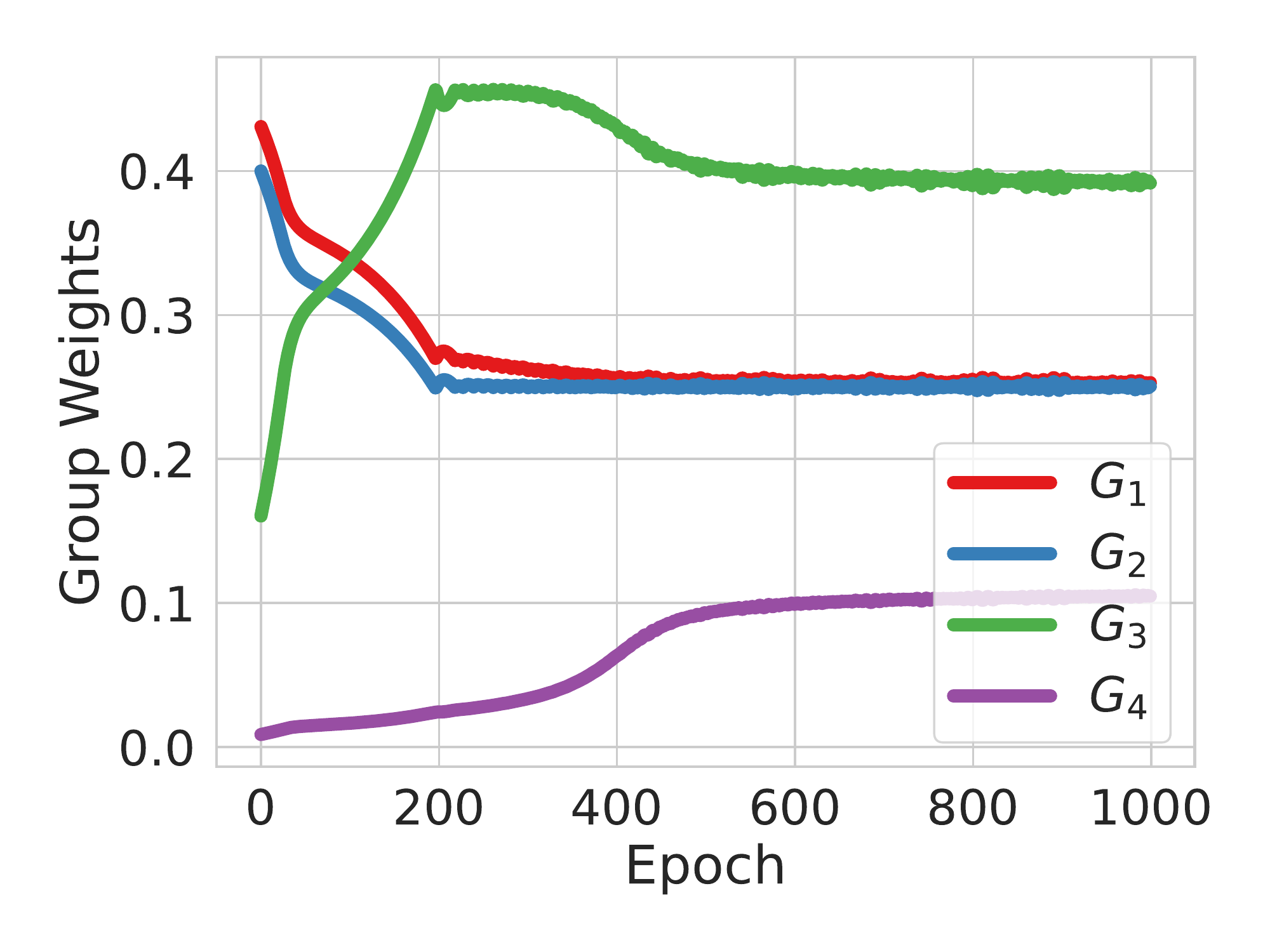}
    \label{fig:celeba_group_fit_weights.pdf}
    \caption{
    \textbf{A trained neural network fails to completely balance group weights on CelebA.}
    Group aggregated weights on CelebA produced by a neural network trained to minimize group-imbalance.
    }
    \label{fig:celea_nn_weights}
\end{figure}

In general, it is not possible to define weights $\mu_i$ purely as a function of the output of an ERM model trained on $\mathcal{D}_\mathrm{ERM}$ such that they are balanced over the groups on the held-out dataset $\mathcal{D}_\mathrm{RW}$. To verify this claim, we train a neural network $f$ to minimize the following loss
\begin{equation*}
    \mathcal{L} = \frac{1}{4} \sum_{g=1}^{4} \left| \left( \sum_{i\in\mathcal{D}_\mathrm{RW}:g_i=g} f(p^\mathrm{ERM}_i, y_i) \right) - \frac{1}{4} \right|,
\end{equation*}
which measures average deviation from $1/4$ of the group aggregated weights produced by the network over 4 groups. The input to $f$ is $(p^\mathrm{ERM}_i, y_i),$ the predicted probabilities for each class by the first-stage ERM model, and the true class label $y_i$.  We use an MLP with two hidden layers and 128 units each, whose output is constrained to be positive using a softplus function and normalized to sum to unity over all examples. Figure~\ref{fig:celea_nn_weights} shows the group aggregated weights on CelebA produced by the network during training, using the Adam optimizer. The group aggregated weights don't converge to $1/4,$ showing that any upweighting strategy based only on ERM prediction and class label is unlikely to produce group-balanced weights on CelebA.

\section{Additional Related Work} \label{Related Work}

\textbf{On spurious correlations.} \quad
Numerous studies expose how neural networks rely on spurious correlations for a diverse set of real word problems.
In image classification, neural networks rely on the background of the image and not on the actual objects as
seen in \citet{xiao2020noise}, \citet{sagawa2020GDRO} and \citet{moayeri2022comprehensive}.
Additionally, neural network classifiers might
rely on object textures \citep{geirhos2018imagenet} or on small image features (not actual spatial relationships) as argued in
\citet{brendel2019approximating}.
Critically, for chest X-ray classification tasks, neural networks have been shown to rely on hospital specific tokens,
chest drains or other features that irrelevant to the diagnosis of pneumonia as reported in \citet{zech2018variable} or \citet{oakden2020hidden}.
For a comprehensive survey of the area, see \citet{geirhos2020shortcut}, and \citet{yang2023change} provide a comprehensive evaluation of the existing methods.

\textbf{Leveraging group annotations.} \quad
When group annotations are present there are several methods that can provide high WGA.
We can leverage the group annotation through:
(i) class or group balancing or weighting \citep{cui2019class, menon2020overparameterisation, idrissi2021simple, kirichenko2022dfr,izmailov2022feature}, or via (ii) distributionally robust optimization \citep{sagawa2020GDRO}, or finally via (iii) contrastive methods \citep{taghanaki2021robust}.
\citet{taghanaki2021robust} propose
\textsc{CIM}: a method that leverages a contrastive loss and pixel-level image statistics to learn input-space transformations
that improve performance on downstream tasks.
However, requiring annotations for a large dataset can be expensive. Worse, as we showed in Section \ref{sec:results}, there are problems where it is not evident what the spurious feature is. Additionally there could be several spurious features present at the same time creating more difficulties.

\textbf{Annotation-free methods.} \quad
There has been plenty of efforts on developing methods that achieve high WGA without requiring group annotations.
A common theme amongst these methods is the presence
of two stages: first, train a checkpoint using
ERM and then, modify this model to improve WGA. We now focus on the
methods that were not discussed in Section \ref{sec:preliminaries}.
In \textsc{GEORGE} \citep{sohoni2020george} the authors infer group annotations based on the clusters formed in the feature space learnt by the ERM feature extractor $f_{\hat{\phi}}$.
With these discovered groups, the authors then run GDRO.
In \citet{nam2022spread} and \citet{sohoni2021barack}, the authors use semi-supervised learning to propagate the limited available group labels to the entire dataset.
In \textsc{LfF} \citep{nam2020lff} the authors use a generalization of the cross-entropy loss as to identify samples that have a strong agreement between the output of the neural network and the label and then to score higher the samples that do not.
Then, having identified the ``difficult" samples, the authors upweight the
cross-entropy loss using the score.
Finally, as shown in Section \ref{sec:preliminaries}, \textsc{JTT} \citep{liu2021jtt} and C\textsc{N}C \citep{zhang2022cnc} are also methods that improve WGA
without requiring group annotations.

In \citet{lin2022zin}, the authors argue how in general it is impossible to perfectly separate the spurious and invariant features without auxiliary information. We note that the good performance of \method\ does not contradict their finding. First, in our experiments we use the auxiliary information about the nature of the spurious features, as we tune the hyperparameters of the method for worst group accuracy on the validation set. More importantly, AFR does not aim to learn invariant features, rather it uses the same features learned by a standard ERM model and only retrains the last layer on a weighted heldout dataset that upweights minority group examples. The discovery that such a procedure is sufficient to achieve SOTA worst group performance on various vision and text benchmarks is an interesting and important one, which does not contradict findings in \citet{lin2022zin}.

\end{document}